%% file: main.tex
\documentclass[10pt]{article}

\usepackage{fullpage}
\usepackage{setspace}
\usepackage{parskip}
\usepackage{titlesec}
\usepackage[section]{placeins}
\usepackage{xcolor}
\usepackage{breakcites}
\usepackage{lineno}
\usepackage{hyphenat}

\usepackage[colorlinks = true,
            linkcolor = blue,
            urlcolor  = blue,
            citecolor = blue,
            anchorcolor = blue]{hyperref}
\usepackage{apacite}
\usepackage{caption}
\usepackage{subcaption}

\usepackage{adjustbox}
\usepackage{todonotes}

\usepackage{multirow}
\usepackage{longtable}
\usepackage{multicol}
\usepackage{array}
\usepackage{graphicx}
\usepackage{amsmath}

\usepackage{rotating}

\usepackage{color, colortbl}

\PassOptionsToPackage{hyphens}{url}
\usepackage{etoolbox}

\renewenvironment{abstract}
  {{\bfseries\noindent{\abstractname}\par\nobreak}\footnotesize}
  {\bigskip}

\titlespacing{\section}{0pt}{*3}{*1}
\titlespacing{\subsection}{0pt}{*2}{*0.5}
\titlespacing{\subsubsection}{0pt}{*1.5}{0pt}

\usepackage{authblk}

\usepackage{graphicx}
\usepackage[space]{grffile}
\usepackage{latexsym}
\usepackage{textcomp}
\usepackage{longtable}
\usepackage{tabulary}
\usepackage{booktabs,array,multirow}
\usepackage{amsfonts,amsmath,amssymb}
\providecommand\citet{\cite}
\providecommand\citep{\cite}

\newif\iflatexml\latexmlfalse

\AtBeginDocument{\DeclareGraphicsExtensions{.pdf,.PDF,.eps,.EPS,.png,.PNG,.tif,.TIF,.jpg,.JPG,.jpeg,.JPEG}}

\usepackage[utf8]{inputenc}
\usepackage[english]{babel}

\usepackage{float}

\usepackage[margin=1.5in]{geometry}

\begin{document}

\title{A survey on datasets for fairness-aware machine learning}

\author[1]{Tai Le Quy\thanks{Corresponding author. tai@l3s.de \includegraphics[scale=0.5]{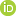}~\href{https://orcid.org/0000-0001-8512-5854}{0000-0001-8512-5854}}}%
\affil[1]{L3S Research Center, Leibniz University Hannover, Germany}%

\author[12]{Arjun Roy\thanks{arjun.roy@fu-berlin.de  \includegraphics[scale=0.5]{orcid.png}~\href{https://orcid.org/0000-0002-4279-9442}{0000-0002-4279-9442}}}%
\affil[2]{Institute of Computer Science, Free University Berlin, Germany} 

\author[1]{Vasileios Iosifidis\thanks{ iosifidis@l3s.de   \includegraphics[scale=0.5]{orcid.png}~\href{https://orcid.org/0000-0002-3005-4507}{0000-0002-3005-4507}}}%

\author[3]{Wenbin Zhang\thanks{wenbinzhang@cmu.edu  \includegraphics[scale=0.5]{orcid.png}~\href{https://orcid.org/0000-0003-3024-5415}{0000-0003-3024-5415}}}
\affil[3]{Carnegie Mellon University, United States} 

\author[2]{Eirini Ntoutsi\thanks{ eirini.ntoutsi@fu-berlin.de   \includegraphics[scale=0.5]{orcid.png}~\href{https://orcid.org/0000-0001-5729-1003}{0000-0001-5729-1003}}}%

\vspace{-1em}
\date{}

\begingroup
\let\center\flushleft
\let\endcenter\endflushleft
\maketitle
\endgroup

\textbf{Article category:}
Overview

\textbf{Conflict of interest:}
The authors have declared no conflicts of interest for this article.

\selectlanguage{english}
\begin{abstract}
As decision-making increasingly relies on Machine Learning (ML) and (big) data, the issue of fairness in data-driven Artificial Intelligence (AI) systems is receiving increasing attention from both research and industry. A large variety of fairness-aware machine learning solutions have been proposed which involve fairness-related interventions in the data, learning algorithms and/or model outputs. However, a vital part of proposing new approaches is evaluating them empirically on benchmark datasets that represent realistic and diverse settings. Therefore, in this paper, we overview real-world datasets used for fairness-aware machine learning. We focus on tabular data as the most common data representation for fairness-aware machine learning. We start our analysis by identifying relationships between the different attributes, particularly w.r.t. protected attributes and class attribute, using a Bayesian network. For a deeper understanding of bias in the datasets, we investigate the interesting relationships using exploratory analysis.
\end{abstract}%
\sloppy

\begin{figure}[h!]
  \centering
  \includegraphics[width=0.5\linewidth]{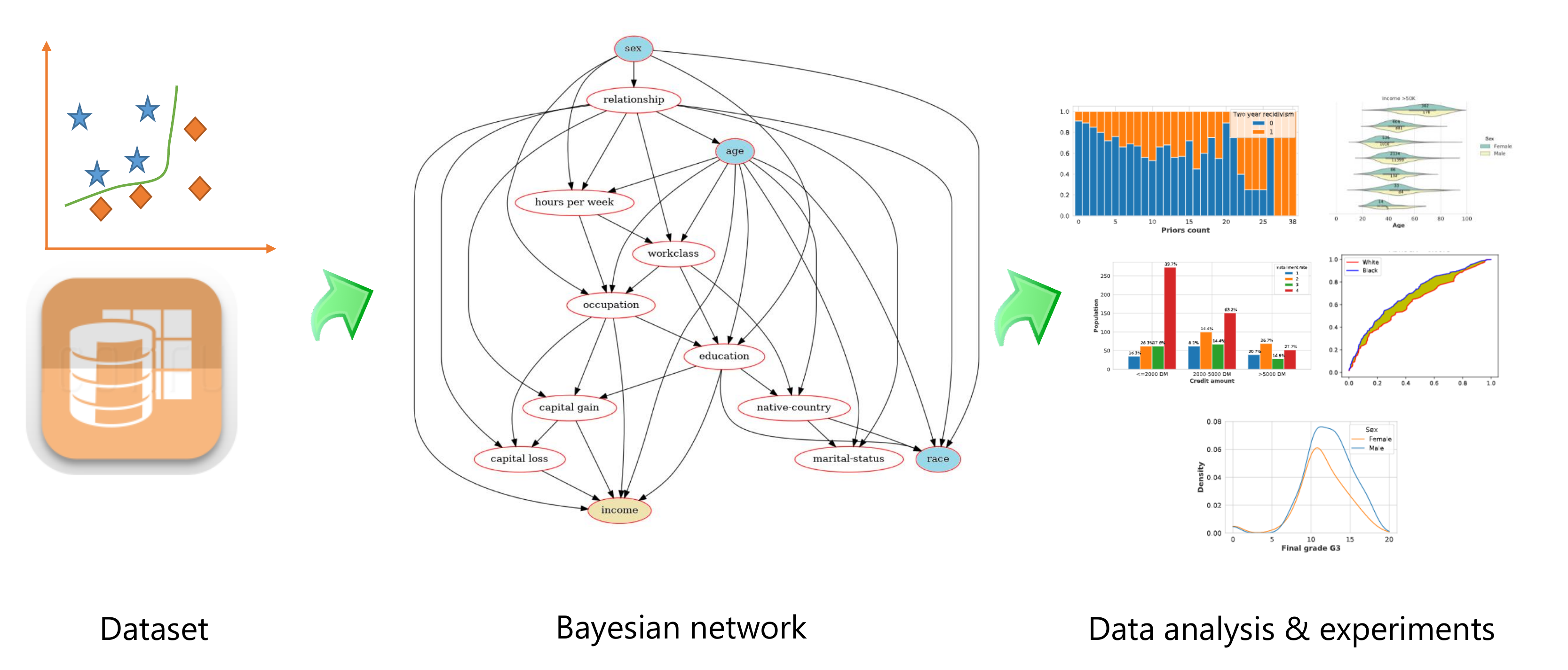}
  \captionsetup{labelformat=empty}
  \caption{A workflow of the survey on datasets for fairness-aware machine learning}
  \label{fig:abstract}
\end{figure}


\input{introduction}
\input{surveymethod}

\input{realdata}

\input{experiment}
\input{other-issues}
\input{conclusion}

\section*{Funding Information}
Ministry  of  Science  and  Education  of  LowerSaxony, German, project ID: 51410078

\section*{Acknowledgements}
The work of the first author is supported by the Ministry of Science and  Education of Lower Saxony, Germany, within the PhD programme  ``LernMINT: Data-assisted teaching in the MINT subjects”. The work of the second author is supported by the Volkswagen Foundation under the call ``Artificial Intelligence and the Society of the Future" (the BIAS project).


\bibliography{bibliography}
\input{appendix}

\end{document}

%% file: introduction.tex
\section{Introduction}
\label{sec:introduction}
Artificial Intelligence and Machine Learning are widely employed nowadays by businesses, governments and other organizations to improve their operational quality and assist in decision-making in areas such as loan approval~\cite{mukerjee2002multi}, recruiting~\cite{faliagka2012application}, school admission~\cite{moore1998expert}, risk prediction~\cite{yeh2009comparisons}. There are many advantages of using algorithmic decision-making as computers can efficiently analyze large amounts of data with high accuracy. Along with the advantages, unfortunately, there is plenty of evidence regarding the discriminative impact of ML-based decision-making on individuals and groups of people on the basis of \emph{protected attributes} such as gender or race. As an example, \textit{racial-bias} was observed in COMPAS~\cite{angwin2016machine}, a software used by the U.S. courts to assess the risk of recidivism; in particular, it has been found that black defendants were predicted with a higher risk of recidivism than their actual risk compared to white defendants. Another example refers to search algorithms in job search websites; it has been found that such algorithms exhibit \textit{gender-bias} as they display higher-paying jobs to male applicants compared to female ones~\cite{simonite2015probing,datta2015automated}.

Data are an essential part of machine learning. Usage of sensitive information during the learning process is undesirable but hard to guarantee even if known protected attributes are omitted from the analysis. The reason is the causal effects~\cite{madras2019fairness} 
of such attributes, including observable ``proxy'' attributes. As an example, the non-protected attribute ``zip-code'' was found to be a proxy for the protected attribute ``race''~\cite{datta2017use} or the ``credit rating'' can be used as a proxy for ``safe driving''~\cite{warner2021making}.
Hence, even if the protected attributes like race or gender are not used, the resulting ML models can still be biased ~\cite{angwin2016machine} due to the causal effects of such attributes. Although methods for detecting proxy attributes exist, e.g., \cite{yeom2018hunting} detects proxies in linear regression models by using a convex optimization procedure, eliminating all the correlated features might drastically reduce the utility of the data for the learning problem.

The domain of bias and fairness in machine learning has attracted much interest in recent years, and as a  result, several surveys exist that provide a broad overview of the area, its technical challenges and solutions~\cite{ntoutsi2020bias,mehrabi2019survey,chhabra2021overview,pitoura2021fairness,xivuri2021systematic}. However, an overview of the datasets used for fairness-aware machine learning evaluation is still missing. As data are a vital part of ML and benchmark datasets a decisive factor for the success of AI research\footnote{https://qz.com/1034972/the-data-that-changed-the-direction-of-ai-research-and-possibly-the-world/}, we believe our survey is serving to fill a gap in the extant research.

In this survey, we overview the different datasets used in the domain of fairness-aware machine learning, and we characterize them according to their application domain, protected attributes and other learning characteristics like cardinality, dimensionality and class (im)balance. For each dataset, we provide an exploratory analysis by first using a Bayesian network to identify the relationships among attributes. Based on the Bayesian network, we provide a graphical analysis of the attributes for a deeper understanding of bias in the dataset. The Bayesian network illustrates the conditional (in)dependence between the protected attribute(s) and the class attribute; thus, it reduces the space and complexity of data analysis that needs to be performed to discover and clarify the fairness-related problems in the dataset. We then focus our exploratory analysis on features having a direct or indirect relationship with the protected attributes. We accompany our exploratory analysis with a quantitative evaluation of measures related to predictive and fairness performance. 

We believe that our survey is useful as it gathers many fairness-related datasets scattered around the web and organizes them in terms of different principles (application domain, learning challenges like dimensionality and class imbalance, fairness-aware related challenges like the number of protected attributes, etc.). As such, we except that it will help researchers to easily select the most appropriate datasets for their application domain (e.g., learning analytics vs recidivism), learning challenges (e.g., balanced vs imbalanced classification), classification task (e.g., binary classification vs multiclass learning), fairness-related challenges (e.g., single protected vs multiple protected attributes etc.).

As datasets have played a foundational role in the advancement of machine learning research~\cite{paullada2021data}, our survey also indicates the need for more open benchmark datasets that would reflect different application domains (from education and healthcare to recruitment and logistics), different contexts (e.g., spatial, temporal, etc.), various (machine) learning challenges (dimensionality, imbalance, number of classes, etc.) as well as different notions of fairness (multi-discrimination, temporal fairness, distributional fairness, etc.). We advocate that the community should also pay attention to benchmark datasets in parallel to new methods and algorithms. The area of fairness-aware machine learning will undoubtedly benefit from having benchmark datasets for various tasks.

The rest of the paper is structured as follows:   In Section~\ref{sec:method},  we describe our methodology for dataset collection and evaluation. The most commonly used datasets for fairness are presented in Section~\ref{sec:realdatasets} together with the results of their exploratory analysis. Section~\ref{sec:experiments} demonstrates a quantitative evaluation of a classification model on the different datasets w.r.t. predictive performance and fairness. We summarize several open issues on datasets for fairness-aware machine learning in Section~\ref{sec:other_issues}. Finally, the conclusion and outlook are summarized in Section~\ref{sec:conclusion}.

%% file: surveymethod.tex
\section{Methodology of the survey process}
\label{sec:method}
In this section, we describe our dataset collection strategy and introduce Bayesian networks as a tool for learning the structure from the data. In addition, we provide a summary of fairness measures we will use for the quantitative evaluation.

\subsection{Strategy for collecting datasets}
\label{subsec:strategy}
To identify the relevant datasets, we use Google Scholar\footnote{https://scholar.google.com/} with ``fairness datasets'' as the primary query term along with other terms like ``bias'', ``discrimination'', ``public'' to narrow down the search. 
After identifying the related datasets, we use Google Scholar to find the related papers which satisfy the following conditions: 1) The public dataset is used in the experiments, and 2) The learning tasks, i.e., classification, clustering,  are related to fairness problems. 
To restrict the investigation of the related work, we consider only important works as assessed by the number of citations, quality of publication venue, i.e., published in ranked conferences, journals. We consider datasets that have been used in at least three fairness-related papers. Datasets that are not publicly available via some known repository like 
the UCI machine learning repository~\footnote{https://archive.ics.uci.edu}, Kaggle~\footnote{https://www.kaggle.com}, 
etc., are not taken into consideration.

\subsection{Bayesian network}
\label{subsec:bayesian}

A Bayesian network (BN)~\cite{holmes2008introduction} is a directed and acyclic probabilistic graphical model which provides a graphical representation to understand the complex relationships between a set of random variables. In the case of a dataset, random variables corresponding to the attributes of the feature space in which the data are represented. The graphical structure $\mathcal{M}:\{\mathcal{V},\mathcal{E}\}$ of a BN contains a set of nodes $\mathcal{V}$ (random variables/attributes) and a set of directed edges $\mathcal{E}$. Let $X_1,X_2,\cdots,X_d$ be the attributes defining the feature space $\mathcal{X}$ of a dataset $\mathcal{D}$, such that $\mathcal{X}\in \mathbb{R}^d$.
For two attributes $X_i,X_j\in \mathcal{X}$, if there is a directed edge from $X_i$ to $X_j$, then $X_i$ is called the parent of $X_j$. The edges indicate conditional dependence relations, i.e., if we denote $X_{{pa}_i}$ as the parents of $X_i$,
the probability of $X_i$ is conditionally dependent on the probability of $X_{{pa}_i}$. If we know the outcome (value) of $X_{{pa}_i}$, then the probability of $X_i$ is conditionally independent of any other ancestor node. The structure of a BN describes the relationships between given attributes, i.e., the joint probability distribution of the attributes in the form of conditional independence relations. Formally:
\begin{equation}
\label{eq:cond_prob}
    \begin{aligned}
        P(X_1,X_2,\cdots, X_d)=\prod\limits_{i=1}^d P(X_{i}\mid X_{{pa}_{i}})
    \end{aligned}
\end{equation}

Learning the structure of a BN from the dataset $\mathcal{D}$ is an optimization problem \cite{husmeier2006probabilistic}, namely to learn an optimal BN model $\mathcal{M}^\star$ which maximizes the likelihood of generating $\mathcal{D}$. A set of parameters of any BN model $\mathcal{M}$, denoted by $\widehat{\mathcal{M}}$, is the set of edges $\mathcal{E}$ which represents the conditional independence relationship between the attribute set $\mathcal{V}$. 
Moreover, between the possible models $M$, the less complex one, i.e., the one with the least $\widehat{\mathcal{M}}$, should be selected. 

Note that in a learned BN model $\mathcal{M}$, the position of the class attribute $y$ can be in any position (root-, internal- or leaf-node), since the objective is to maximize $P(\mathcal{D}\mid \mathcal{M})$. However, we aim to investigate the factors (protected/non-protected attributes) that determine the class attribute's prediction probability. Therefore, we also employ a constraint on the class attribute to be a leaf node in our learning objective. Formally the problem is defined as:
\begin{equation}\label{eq:model_learn}
  \begin{aligned}
        \max\limits_{\mathcal{M}^\star} \{ P(\mathcal{D}\mid\mathcal{M}) - \gamma \widehat{\mathcal{M}}\}\\
        \text{subject to } y\in \mathcal{L}
  \end{aligned}
\end{equation}
where $y\in \mathcal{X}$ is the class attribute, $\mathcal{L}$ is the set of leaf nodes and $\gamma$ is a penalty hyperparameter controlling the effect of the model's complexity in the final model selection. 
The aim of the learned model is to maximize $P(X_i \mid X_{{pa}_i})$ for each $X_i\in \mathcal{X}$ (Eq.~\ref{eq:cond_prob} and Eq.~\ref{eq:model_learn}). 

A high conditional probability often refers to a strong correlation \cite{daniel2017thinking}. Attribute $X_i$ is strongly correlated with $X_j$ if there exists a \emph{direct edge} between $X_i$ and $X_j$, for any pair of attributes $X_i,X_j\in \mathcal{X}$.
Intuitively, the correlation is comparatively weaker with ancestors that are not immediate parents, i.e., \emph{indirect edges}. In addition, the attributes which do not have any incoming or outgoing edge (direct/indirect connection) with $X_i$, the correlation between them will be negligible.
As a consequence, if we find any direct/indirect edge from any protected attribute to the class attribute in our learned BN structure $\mathcal{M}^\star$ then we may infer that the dataset is biased w.r.t. the specific protected attribute. 

When learning a BN, the continuous variables are often discretized because many BN learning algorithms cannot efficiently handle continuous variables~\cite{chen2017learning}. Therefore, we need to discretize the continuous numeric data attributes into meaningful categorical attributes to keep the complexity of learning the BN model in a polynomial time. We describe the discretization procedure for each dataset in Section \ref{sec:realdatasets}.

\subsection{Fairness metrics}
\label{subsec:fairness_metrics}
Measuring bias in ML models comprises the first step to bias elimination. Fairness depends on context; thus, a large variety of fairness measures exists. Only in the computer science research area, more than 20 measures of fairness have been introduced thus far~\cite{vzliobaite2017measuring,verma2018fairness}. 
Nevertheless, there is no fairness measure that is universally suitable \cite{foster2016big,verma2018fairness}. 
Therefore, to make the experimental results more diverse, we report on three prevalent fairness measures: \textit{statistical parity}, \textit{equalized odds} and \textit{Absolute Between-ROC Area (ABROCA)}. In which, \textit{statistical parity} \cite{dwork2012fairness} is one of the earliest and most popular discrimination measures in the fairness-aware ML literature. 
Statistical parity is also considered as the statistical counterpart of the legal doctrine of disparate impact \cite{krop1981age}. However, one main disadvantage of statistical parity is that it does not require compliance to the ground truth labels; hence, in many ML scenarios might not be ideal \cite{hardt2016equality}. \textit{Equalized odds} introduced by \cite{hardt2016equality} countered this problem by considering the ground truth of both positive and negative class instances and grew to be one of the most promising fairness notion, being used in the leading edge methods \cite{zafar2017fairness,krasanakis2018adaptive,iosifidis2019adafair}. Later, \cite{gardner2019evaluating} argued that \textit{equalized odds} does not consider any formal strategy such as slicing analysis to identify the prevalent biases, which might be a necessity in particular domains such as education. ABROCA measure introduced by \cite{gardner2019evaluating} tackles such an analysis issue and is argued to be an illustratively efficient method of representing the divergence of values of a protected attribute. Although, as mentioned earlier, there is a rich literature of fairness notions to follow, in this work, we limit our study to the above-mentioned notions, as these notions together cover a diverse area of the fairness concepts currently followed in the state-of-the-art fairness-aware ML practices.

The measures are presented hereafter assuming the following problem formulation:
Let $\mathcal{D}$ be a binary classification dataset with class attribute
$ y = \{+, -\}$. 
Let $S$ be a binary protected attribute with $S \in \{s,\overline{s}\}$, in which
\textit{s} is the discriminated group (referred to as \emph{protected group}), and $\overline{s}$ is the non-discriminated group (referred to as \emph{non-protected group}). For example, let $S$ = ``Sex'' $\in$ \{Female, Male\} be the protected attribute; $s$ = ``Female'' could be the protected group and $\overline{s}$ = ``Male'' could be the non-protected group.
We use the notation $s_{+}$ ($s_{-}$), $\overline{s}_{+}$ ($\overline{s}_{-}$) to denote the protected and non-protected groups for the positive (negative, respectively) class.

\subsubsection{Statistical parity}

Statistical parity (SP) introduced by \cite{dwork2012fairness} states that the output of any classifier satisfies SP if the difference (bias) in predicted outcome ($\hat{y}$) between any two groups under study (i.e., $s$ and $\overline{s}$) is up to a predefined tolerance threshold $\epsilon$. Formally:

\begin{equation}
\label{eqn:statistical_parity_def}
 P(\hat{y}|S=s) - P(\hat{y}|S=\overline{s}) \leq \epsilon
\end{equation}
Using the definition in Eq.~\ref{eqn:statistical_parity_def} to measure the bias of a classifier, various measuring notions~\cite{simoiu2017sp_benchmarking,zliobaite2015relation} have been proposed. The violation of statistical parity can be measured as:
\begin{equation}
\label{eqn:statistical_parity}
SP =  P(\hat{y}=+|S=\overline{s}) - P(\hat{y}=+|S=s)
\end{equation}
The value domain is: $SP \in [-1, 1]$,
with $SP=0$ standing for no discrimination, $SP\in (0,1]$ indicating that the protected group is discriminated, and  $SP\in [-1,0)$ meaning that the non-protected group is discriminated (\emph{reverse discrimination}).

\subsubsection{Equalized odds}  
Equalized odds (shortly \textit{Eq.Odds})~\cite{hardt2016equality} is preserved when the predictions $\hat{y}$ conditional on the ground truth $y$ is equal for both the groups $s$ and $\overline{s}$ defined by $S$. Formally: 
\begin{equation}\label{eq: eq_odds_def}
   Eq. Odds: P(\hat{y}=+|S=s,Y=y)= P(\hat{y}=+|S=\overline{s},Y=y)
\end{equation}
where $y$ is the ground truth class label, $\hat{y}$ is the predicted label.

Using Eq.~\ref{eq: eq_odds_def} we can measure the prevalent bias as: 
\begin{equation}
\label{eq:equalized_odds}
Eq.Odds_{viol} = \sum_{y\in \{+,-\}}|P(\hat{y}=+|S=s,Y=y) - P(\hat{y}=+|S=\overline{s},Y=y)| 
\end{equation}

The value domain is: $Eq.Odds_{viol} \in [0, 2]$, with 0 standing for no discrimination and 2 indicating the maximum discrimination.

\subsubsection{Absolute Between-ROC Area (ABROCA)} This is a fairness measure introduced by the research of~\cite{gardner2019evaluating}. It is based on the Receiver Operating Characteristics (ROC) curve. ABROCA measures the divergence between the protected ($ROC_{s}$) and non-protected group ($ROC_{\overline{s}}$) curves across all possible thresholds $t \in [0,1]$ of \emph{false positive rates} and \emph{true positive rates}. 
In particular, it measures the absolute difference between the two curves in order to capture the case that the curves may cross each other and is defined as:
\begin{equation}
\label{eq:abroca}
    \int_{0}^{1}\mid ROC_{s}(t) - ROC_{\overline{s}}(t)\mid \,dt
\end{equation}

ABROCA takes values in the $[0, 1]$ range. The higher value indicates a higher difference in the predictions between the two groups and therefore, a more unfair model.

%% file: realdata.tex
\section{Datasets for fairness}
\label{sec:realdatasets}
In this section, we provide a detailed overview of real-world datasets used frequently in fairness-aware learning.
We organize the datasets in terms of the application domain, namely: financial datasets (Section~\ref{subsec:financial}), criminological datasets (Section~\ref{subsec:criminology}), healthcare and social datasets (Section~\ref{subsec:social}) and educational datasets (Section~\ref{subsec:educational}).
\input{realdata_overview}

For each dataset, we discuss the basic characteristics like cardinality, dimensionality and class imbalance as well as typically used protected attributes in the literature. When available, we also provide temporal information regarding the data collection and the timespan of the datasets.

We start our analysis with the BN structure learned from the data (see Section \ref{subsec:bayesian}), which can help us to understand the relationships between attributes of the dataset. In addition, the BN visualization already provides interesting insights on the dependencies between non-protected and protected attributes and their conditional dependencies in predicting the class attribute. 
We further provide an exploratory analysis of interesting correlations from the Bayesian graph (for both direct- and indirect- edges), particularly those related to the fairness problem (paths to and from protected attributes).

\input{realdata_finance}
\input{realdata_criminology}

\input{realdata_society}

\input{realdata_education}

%% file: realdata_overview.tex
A summary of the statistics of the different datasets\footnote{We use the names of the protected attributes given in the original datasets, i.e. \emph{sex, gender} are used with the same meaning. We do not present the \emph{class ratio} (denoted by `-') in several datasets because their \emph{class label} is `multiple'.} is provided in Table \ref{tbl:real_world_datasets}. 

\begin{table*}[h!]
\centering
\caption{Overview of real-world datasets for fairness.}
\label{tbl:real_world_datasets}
\begin{adjustbox}{width=1\textwidth}
    \begin{tabular}{lrrcccccccc}
        \hline
        \multicolumn{1}{c}{\textbf{Dataset}} & \multicolumn{1}{c}{\textbf{\#Instances}} & 
        \multicolumn{1}{c}{\begin{tabular}[c]{@{}c@{}}\textbf{\#Instances}\\\textbf{(cleaned)}\end{tabular}} &
        \multicolumn{1}{c}{\begin{tabular}[c]{@{}c@{}}\textbf{\#Attributes}\\\textbf{(cat./bin./num.)}\end{tabular}} &  
        \textbf{Class} & 
        \multicolumn{1}{c}{\textbf{Domain}}  & \multicolumn{1}{c}{\begin{tabular}[c]{@{}c@{}}\textbf{Class}\\\textbf{ratio (+:-)}\end{tabular}} & 
        \multicolumn{1}{c}{\begin{tabular}[c]{@{}c@{}}\textbf{Protected}\\\textbf{attributes} \end{tabular}} &
        \multicolumn{1}{c}{\textbf{Target class}} &
        \multicolumn{1}{c}{\begin{tabular}[c]{@{}c@{}}\textbf{Collection}\\\textbf{period} \end{tabular}} &
        \multicolumn{1}{c}{\begin{tabular}[c]{@{}c@{}}\textbf{Collection}\\\textbf{location} \end{tabular}} \\ \hline
        \rowcolor{gray!0}
        Adult& 48,842 & 45,222 &7/2/6  & Binary   & Finance & 1:3.03 & Sex, race, age & Income & 1994 & US\\
        \rowcolor{gray!0}
        KDD Census-Income & 299,285 & 284,556 & 32/2/7 & Binary  & Finance & 1:15.30 & Sex, race & Income & 1994-1995 & US\\
        \rowcolor{gray!0}
        German credit & 1,000 & 1,000 & 13/1/7 & Binary   & Finance & 2.33:1 & Sex, age & Credit score & 1973-1975 & Germany\\
        \rowcolor{gray!0}
        Dutch census & 60,420 & 60,420 & 10/2/0 & Binary   & Finance & 1:1.10 & Sex & Occupation & 2001 & Netherlands\\
        \rowcolor{gray!0}
        Bank marketing & 45,211 & 45,211 & 6/4/7 & Binary  & Finance & 1:7.55 & Age, marital & Deposit subscription & 2008-2013 & Portugal\\
        \rowcolor{gray!0}
        Credit card clients & 30,000 & 30,000 & 8/2/14 & Binary & Finance & 1:3.52 & Sex, marriage, education & Default payment & 2005 & Taiwan\\ 
        \rowcolor{gray!10}
        COMPAS recid. & 7,214 & 6,172 & 31/6/14 & Binary  & Criminology & 1:1.20 & Race, sex & Two year recidivism & 2013-2014 & US\\
        \rowcolor{gray!10}
        COMPAS viol. recid. & 4,743 & 4,020 &31/6/14 & Binary   & Criminology & 1:5.17 & Race, sex & Two year violent recid. & 2013-2014 & US\\
        \rowcolor{gray!10}
        Communities\&Crime & 1,994 & 1,994 &4/0/123 & Multi  & Criminology & - & Black &  Violent crimes rate & 1995 & US\\ 
        \rowcolor{gray!20}
        Diabetes & 101,766 & 45,715& 33/7/10 & Binary & Healthcare & 1:3.13 & Gender & Readmit in 30 days & 1999–2008 & US\\ 
        \rowcolor{gray!20}
        Ricci & 118 & 118 & 0/3/3 & Binary  &  Society & 1:1.11 & Race  & Promotion & 2003 & US\\ \rowcolor{gray!40}
        Student-Mathematics & 649 & 649 & 4/13/16 &Binary& Education &1:2.04&Sex, age & Final grade & 2005-2006 & Portugal\\
        \rowcolor{gray!40}
        Student-Portuguese & 649 & 649 & 4/13/16 &Binary& Education &1:5.49&Sex, age & Final grade & 2005-2006 & Portugal\\
        \rowcolor{gray!40}
        OULAD & 32,593 & 21,562 & 7/2/3 & Multi &  Education & - & Gender & Outcome & 2013-2014 & England\\
        \rowcolor{gray!40}
        Law School & 20,798 & 20,798 & 3/3/6 & Binary & Education & 8.07:1 & Male, Race & Pass the bar exam & 1991 & US\\
        \hline
    \end{tabular}
\end{adjustbox}
\end{table*}

%% file: realdata_finance.tex
\subsection{Financial datasets}
\label{subsec:financial}

\subsubsection{Adult dataset}
\label{subsubsec:adult}
\setcounter{figure}{0}
The adult dataset~\cite{kohavi1996scaling} (also known as ``Census Income'' dataset\footnote{https://archive.ics.uci.edu/ml/datasets/adult}) is one of the most popular datasets for fairness-aware classification studies
(Appendix~\ref{sec:citation}). 
The classification task is to decide whether the annual income of a person exceeds 50,000 US dollars based on demographic characteristics. 

\noindent\textbf{Dataset characteristics:}
The dataset consists of 48,842 instances, each  described via 15 attributes, of which 6 are numerical, 7 are categorical and 2 are binary attributes.  An overview of  attribute characteristics is shown in Table~\ref{tbl:adult_attribute}. We discard the attribute \textit{fnlwgt} (final weight) as the suggestions of related work~\cite{zhang2018mitigating,kamiran2012data,calders2009building,calders2010classification}. 

\begin{table*}[h!]
\caption{Adult: attributes characteristics}
\label{tbl:adult_attribute}

\begin{adjustbox}{width=1\linewidth}
    \begin{tabular}{llccl}
        \hline
        \multicolumn{1}{c}{\textbf{Attributes}} & \multicolumn{1}{c}{\textbf{Type}} & 
        \multicolumn{1}{c}{\textbf{Values}} & 
        \multicolumn{1}{c}{\textbf{\#Missing values}} & 
        \multicolumn{1}{c}{\textbf{Description}}
        \\ \hline
        age & Numerical &   [17 - 90] & 0 & The age of an individual\\
        workclass & Categorical & 7 & 2,799 & The employment status (Private, State-gov, etc.)\\
        fnlwgt  & Numerical &   [13,492 - 1,490,400] & 0 & The final weight\\ 
        education & Categorical & 16 & 0 & The highest level of education\\
        educational-num & Numerical &   1 - 16 & 0 & The highest level of education achieved in numerical form\\
        marital-status & Categorical & 7  & 0 & The marital status\\
        occupation & Categorical & 14 & 2,809 & The general type of occupation\\
        relationship & Categorical & 6 & 0 & Represents what this individual is relative to others\\
        race & Categorical & 5 & 0 & Race\\
        sex & Binary &   \{Male, Female\} & 9 & The biological sex of the individual\\
        capital-gain & Numerical &   [0 - 99,999] & 0 & The capital gains for an individual \\
        capital-loss & Numerical &   [0 - 4,356] & 0 & The capital loss for an individual\\
        hours-per-week & Numerical &   [1 - 99] & 0 & The hours an individual has reported to work per week\\
        native-country & Categorical & 41 & 857 & The country of origin for an individual\\
        income & Binary  & \{$\leq$50K, $>$50K\} & 0 & Whether or not an individual makes more than \$50,000 annually\\
        \hline
    \end{tabular}
\end{adjustbox}

\end{table*}

Missing values exist in 3,620 (7.41\%) records.
Many researchers remove the instances containing missing values~\cite{zafar2015fairness,iosifidisdealing,iosifidis2019adafair,choi2020learning} in their experiments; other researches consider the whole dataset or do not clarify how the missing values are handled. To avoid the effect of missing values on the analysis, we remove the missing data and obtain a clean dataset with 45,222 instances.

\noindent\textbf{Protected attributes:} Typically the following attributes have been used as bias triggers in the literature\footnote{Please note that the majority of fairness-aware ML methods can handle only single protected attribures. The problem of multi-fairness has only recently been addressed~\cite{hebert2018multicalibration,martinez2020minimax,abraham2019fairness}}:
\begin{itemize}
    \item \emph{sex} = \{\textit{male, female}\}:
    the dataset is dominated by male instances. The ratio of \textit{male:female}  is 32,650:16,192 (66.9\%:33.1\%).
    \item  \emph{race} = \{\textit{white, black, asian-pac-islander, amer-indian-eskimo, other}\}. Typically, \textit{race} is used as a binary attribute in the related work~\cite{luong2011k,chakraborty2020making,zafar2015fairness}: \emph{race} = \{\textit{white, non-white}\}. The dataset is dominated by \textit{white} people, the  \textit{white:non-white} ratio is 38,903:6,319 (86\%:14\%).
    In our analysis we also encode \textit{race} as a binary attribute.  
    \item \emph{age} = [17-90]. Typically, \textit{age} is used as a categorical attribute in the related work. In our analysis, we also discretize \textit{age} as~\cite{zafar2015fairness}: \emph{age} = \{25-60, $<$25 or $>$60\}. The dataset is dominated by the $[25-60]$ years old group, the ratio is 35,066:10,156 (77.5\%:22.5\%). 
\end{itemize}
In the research of \cite{deepak2020fair}, \emph{marital-status} and \emph{native-country} are considered as the protected attributes. However, due to missing information on their pre-processing method on these attributes, we will not consider those as the protected attributes in our survey.

\textbf{Class attribute:} The class attribute is \emph{income} $\in \{\leq50K, >50K\}$ indicating whether an individual makes less or more than 50K. The positive class is ``$>50K$''. The dataset is imbalanced with an imbalance ratio (IR) $1:3.03$ (positive:negative).

\begin{figure}[h!]
  \centering
  \includegraphics[width=0.7\linewidth]{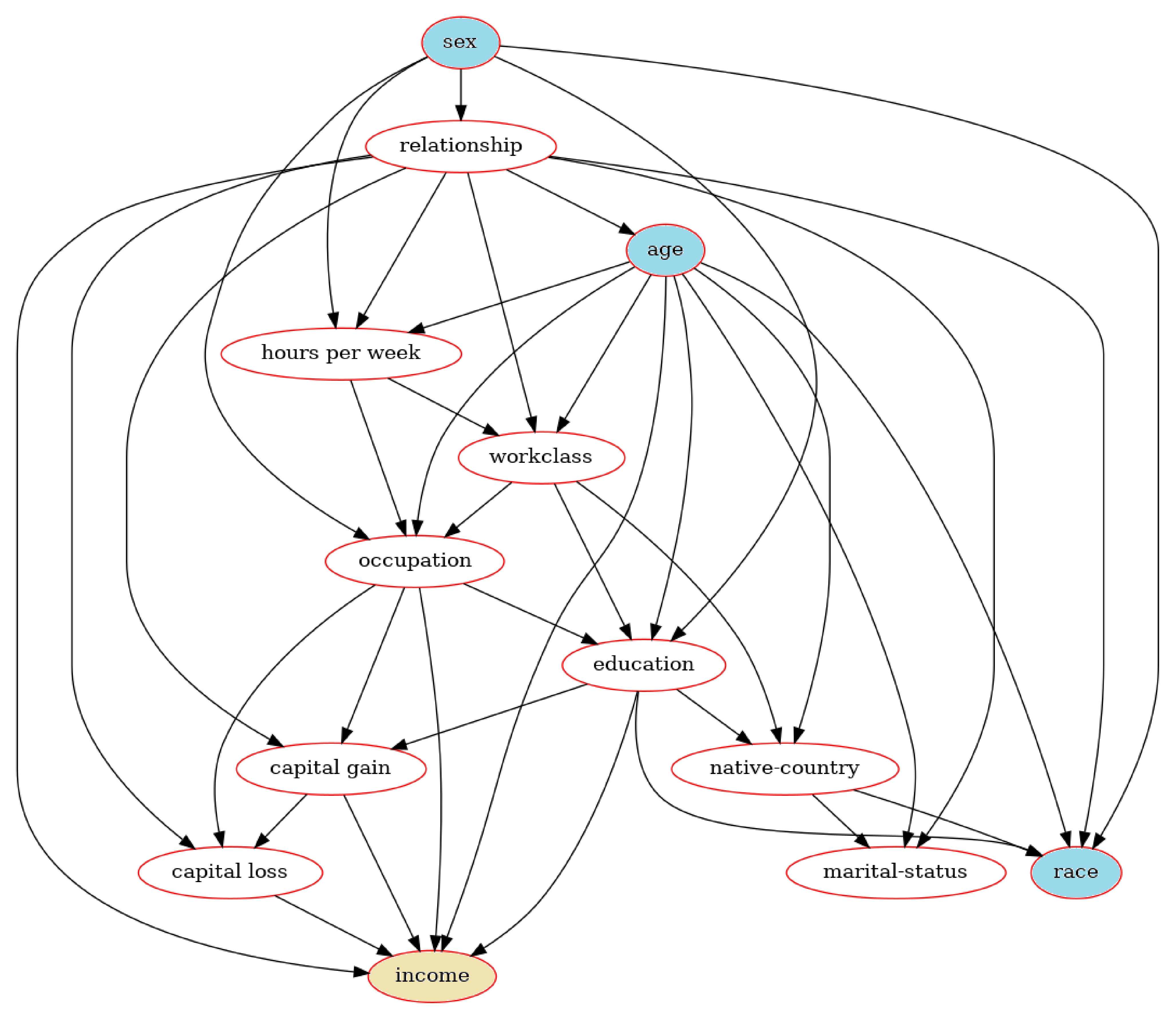}
  \caption{Adult: Bayesian network (class label: \textit{income}, protected attributes: \textit{sex, race, age})}
  \label{fig:adult_NB}
\end{figure}

\noindent\textbf{Bayesian network}: Figure~\ref{fig:adult_NB} illustrates the Bayesian network learned from the dataset. The class label \textit{income} is the leaf node, i.e., there are no outgoing edges. To generate the Bayesian network, we discretize four numerical attributes (\textit{age, capital gain, capital loss, hours per week}) as follows: \textit{age} = \{25-60, $<$25 or $>$60\};
\textit{capital gain} = \{$\leq$5000, $>$5000\}, 
\textit{capital loss} = \{$\leq$40, $>$40\}; \textit{hours per week} = \{$<$40, 40-60, $>$60\}. 
To reduce the computation space of the BN generator, we also transform seven categorical attributes as follows: \textit{workclass} = \{\textit{private, non-private}\}; \textit{education} = \{\textit{high, low}\}; \textit{marital-status} = \{\textit{married, other}\};  \textit{relationship} = \{\textit{married, other}\};  \textit{native-country} = \{\textit{US, non-US}\};  \textit{race} = \{\textit{white, non-white}\};  \textit{occupation} = \{\textit{office, heavy-work, other}\}.

\begin{figure} [h!]
     \centering
     \begin{subfigure}[b]{0.48\linewidth}
         \centering
         \includegraphics[width=1\linewidth]{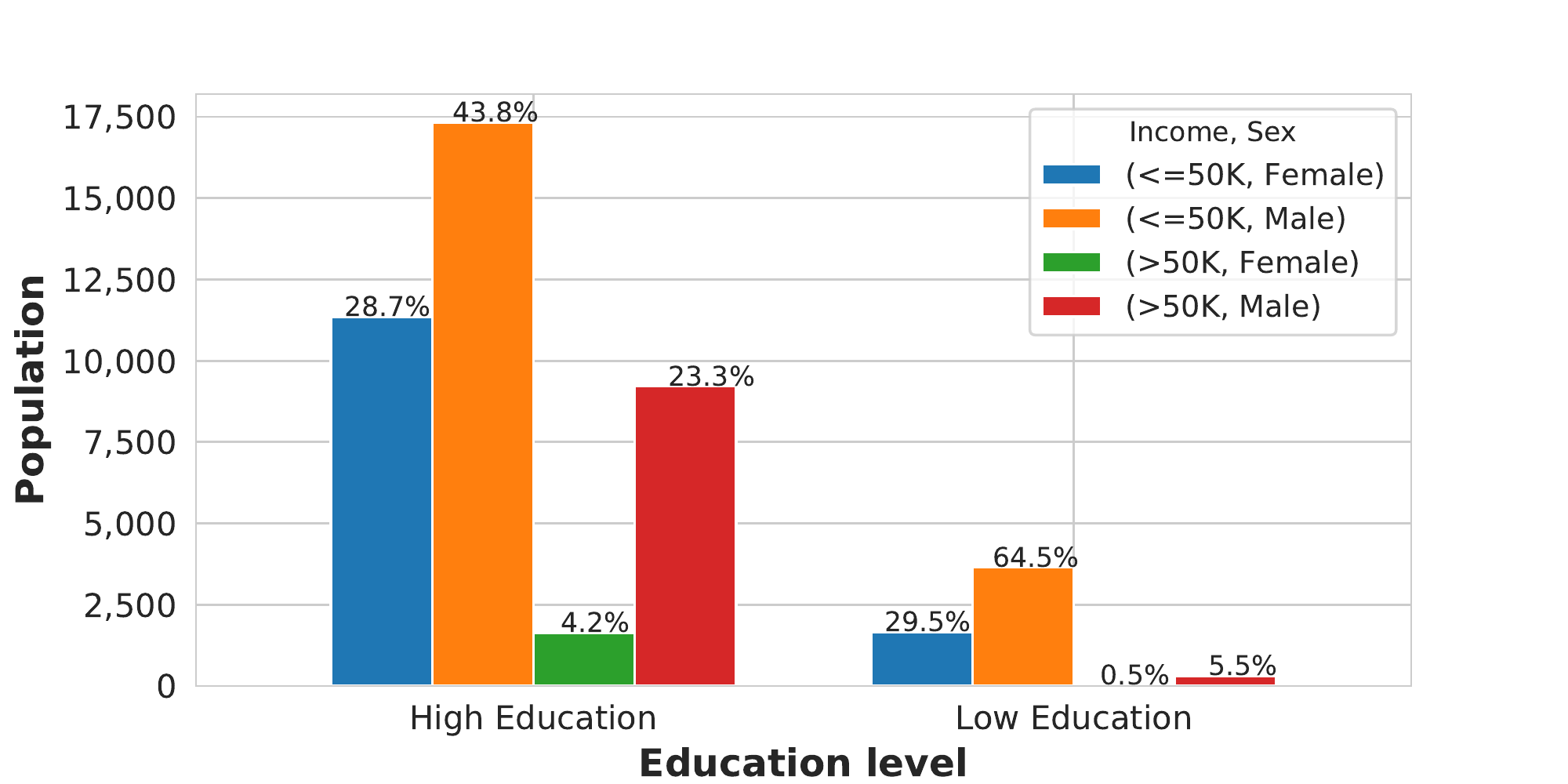}
         \caption{Distribution of \textit{education} and \textit{income} w.r.t \textit{sex}} 
         \label{fig:adult_edu_income_sex}
     \end{subfigure}
     \quad
     \begin{subfigure}[b]{0.48\linewidth}
         \centering
         \includegraphics[width=1\linewidth]{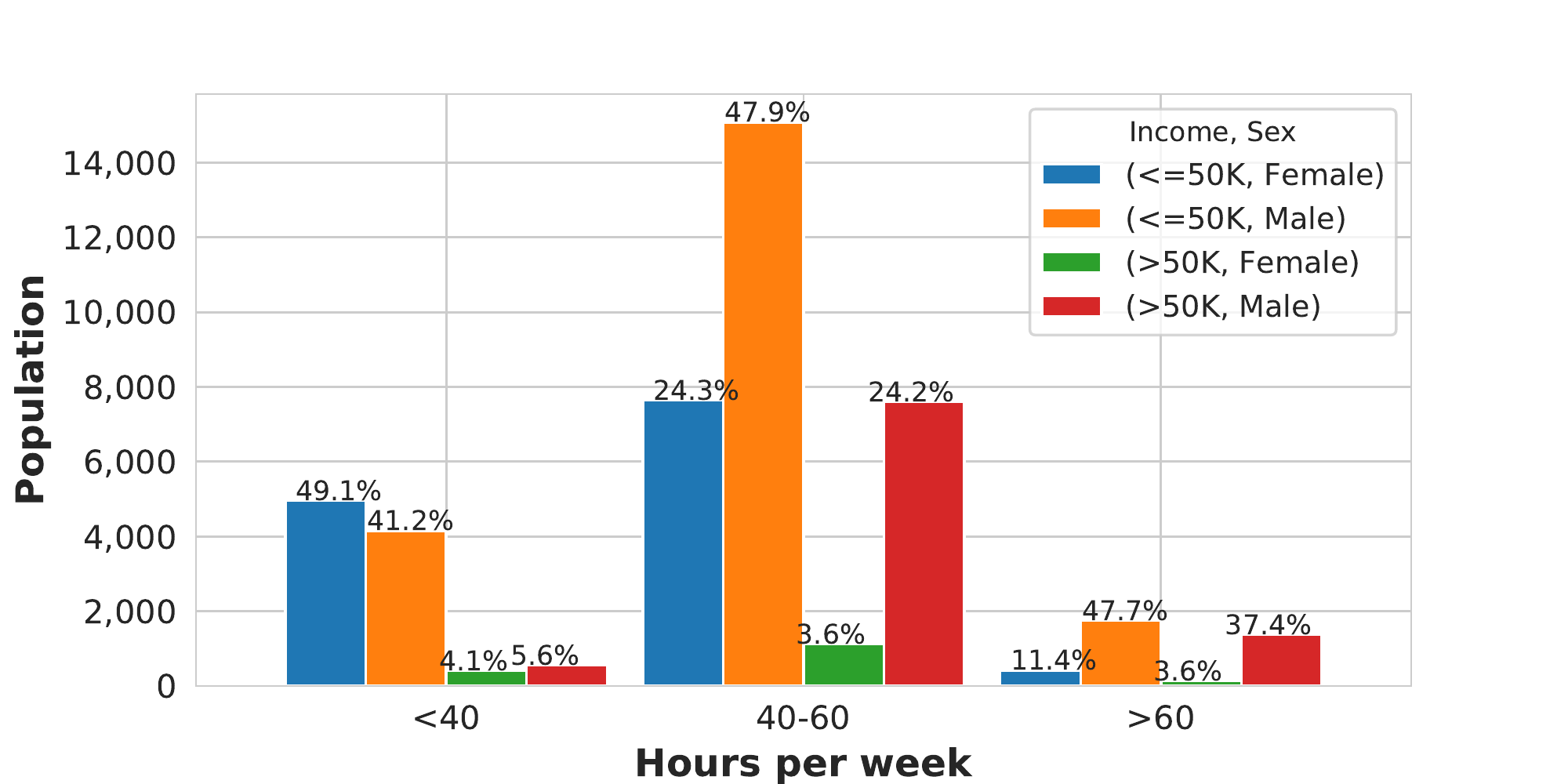}
         \caption{Distribution of \textit{weekly working hours} and \textit{income} w.r.t \textit{sex}}
         \label{fig:adult_hours_income_sex}
     \end{subfigure}
     \caption{Adult: relationships of \textit{education, weekly working hours, education} and \textit{income} attributes}
     \label{fig:adult_attributes}
\end{figure}

As demonstrated in Figure~\ref{fig:adult_NB}, there is a direct dependency between
\textit{income} and \textit{education} as well as between  \textit{sex} and \textit{education}. Therefore, we explore in more detail the distribution of the population w.r.t. \textit{education, income} and \textit{sex} in 
Figure~\ref{fig:adult_edu_income_sex}. As expected, highly educated people have a high income. 
 However, in the high education segment and for the high-income class, the number of males is at least 5 times higher than that of females showing an under-representation of high education women in the high-income class. Based on the dependence of \textit{hours per week} attribute on \textit{sex}, we plot the weekly working hours w.r.t \textit{income} and \textit{sex} (Figure~\ref{fig:adult_hours_income_sex}). The number of males who work more than 40 hours per week is approximately 7 times higher than the number of females.

\begin{figure}[!h]
  \centering
  \includegraphics[width=\linewidth]{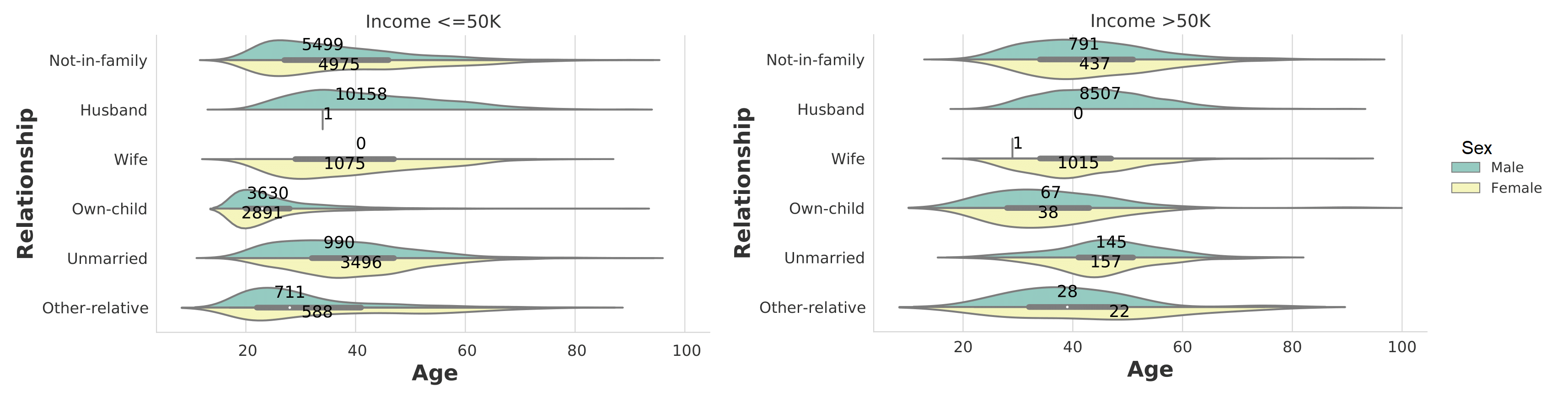}
  \caption{Adult: distribution of \textit{age}, \textit{relationship} and \textit{income} w.r.t \textit{sex}}
  \label{fig:adult_violin_relationship_age_income_sex}
\end{figure}

Interestingly, there are many outgoing edges from the \textit{relationship} and \textit{age} attributes in the Bayesian network. We show the distribution of \textit{sex} in each class based on the \textit{age} (x-axis) and the \textit{relationship} status (y-axis) in Figure~\ref{fig:adult_violin_relationship_age_income_sex}. 
A first observation is that a great amount of young (less than 25 years old) or old (more than 60 years old) people do not receive more than 50K.
\textit{``Unmarried''} people have an income higher than 50K when they are older than 45 years, while people in the \textit{``Own-child''} group can have a high income when they are young. In general, there are more males than females for almost all relationship statuses for the high-income group.

Another interesting observation is that there is a direct edge from protected attribute \emph{sex} to \emph{race}. This suggests that choosing \emph{sex} as the protected attribute would make the fairness-aware classifier attain fairness w.r.t \emph{race}. Evidence of such outcome is seen in the work of \cite{friedler2019comparative}.

\subsubsection{KDD Census-Income dataset}
\label{subsubsec:kdd-adult}
The KDD Census-Income\footnote{https://archive.ics.uci.edu/ml/datasets/Census-Income+(KDD)} dataset~\cite{dua2017} was collected from  Current Population Surveys implemented by the U.S. Census Bureau from 1994 to 1995. The dataset has been considered in numerous related works (Appendix~\ref{sec:citation}).
The prediction task is to decide if a person receives more than 50,000 US dollars annually or not. The prediction task is the same as the \textit{Adult} dataset. However, the differences between the two datasets described by the dataset's authors~\cite{dua2017} are: ``the goal field was drawn from the \textit{total person income} field rather than the \textit{adjusted gross income} and may, therefore, behave differently than the original adult goal field''. 
\begin{table*}[h!]
\caption{KDD Census-Income: attributes characteristics}
\label{tbl:kdd_attribute}
\begin{adjustbox}{width=1\linewidth}
    \begin{tabular}{llccl}
         \hline
        \multicolumn{1}{c}{\textbf{Attributes}} & \multicolumn{1}{c}{\textbf{Type}} & 
        \multicolumn{1}{c}{\textbf{Values}} & 
        \multicolumn{1}{c}{\textbf{\#Missing values}} &
        \multicolumn{1}{c}{\textbf{Description}}
        \\ \hline
age & Numerical &  [0 - 90] & 0 & The age of an individual\\
workclass  & Categorical &  9  & 0 & Represents class of the worker \\
industry & Categorical & 52   & 0 & The industry code\\
occupation & Categorical & 47  & 0 & The occupation code\\
education  & Categorical &  17  & 0 & The highest level of education\\
wage-per-hour & Numerical &   [0 - 9,999] & 0 & Wage per hour\\
marital-status  & Categorical &  7  & 0 & The marital status\\
race  & Categorical &  5  & 0 & Race\\
sex  & Binary &   \{Male, Female\}& 0 & The biological sex of the individual\\
employment-status  & Categorical &  8  & 0 & The employment status (full or part time)\\
capital-gain & Numerical &   [0 - 99,999] & 0 & The capital gains for an individual\\
capital-loss & Numerical &   [0 - 4,608] & 0 & The capital loss for an individual\\
dividends-from-stocks & Numerical &   [0 - 99,999] & 0 & The dividends from stocks\\
tax-filer-stat  & Categorical &  6  & 0 & The tax filer status {(joint under 65, joint 65+, etc.)}\\
detailed-household-and-family-stat  & Categorical &  38  & 0 & The detailed household and family {(child under 18, grandchild etc.)}\\
detailed-household-summary-in-household  & Categorical &  8  & 0 & The detailed household summary {(spouse, non-relative, etc.)}\\
num-persons-worked-for-employer & Numerical &   [0 - 6] & 0 & The number of persons worked for the employer\\
family-members-under-18  & Categorical &  5  & 0 & Family members under 18 {(both parent, mother only, etc.)}\\
citizenship  & Categorical &  5  & 0 & The citizenship\\
own-business & Categorical & 3   & 0 & Own business or self employed\\
veterans-benefits & Categorical & 3 &  0 & Veterans benefits \\
weeks-worked & Numerical &   [0 - 52] & 0 & The number of weeks worked in a year \\
year & Categorical & 2  & 0 & The year in which the interviewee answered\\
income (class)  & Binary &    \{$\leq$50K, $>$50K\} & 0 & Whether an individual makes more than \$50,000 annually \\
        \hline
    \end{tabular}
\end{adjustbox}
\end{table*}

\noindent\textbf{Dataset characteristics:} The dataset contains 299,285 instances with 41 attributes, 32 of which are categorical, 7 are numerical and 2 are binary attributes.  An overview of the dataset characteristics\footnote{Table~\ref{tbl:kdd_attribute} describes attributes used in the Bayesian network} is shown in Table~\ref{tbl:kdd_attribute} and Table~\ref{tbl:kdd_attribute_extra} (Appendix~\ref{sec:dataset_characteristics}). Attribute \textit{weight} is omitted as proposed by the authors of the dataset~\cite{dua2017}.

Missing values exist in 157,741 (52.71\%) instances. Because related studies only focus on a subset of data and features, we clean the dataset by eliminating all missing values. 
In particular, we remove four features \textit{migration-code-change-in-msa, migration-code-change-in-reg, migration-code-move-within-reg, migration-prev-res-in-sunbelt} due to their high proportion in the missing values, as illustrated in Table~\ref{tbl:kdd_attribute}.
The result is a cleaned dataset with 284,556 instances. 

\noindent\textbf{Protected attributes:} Previous researches consider \textit{sex} as a protected attribute~\cite{iosifidis2019adafair,ristanoski2013discrimination,iosifidis2020fabboo}. Attribute  \textit{race} = \{\textit{white, black, asian-pac-islander, amer-indian-eskimo, other}\} could be also employed as a protected attribute because it has the same role as in the original \textit{Adult} dataset. 
Similarly to the Adult dataset, the KDD Census-Income dataset is dominated by \textit{white} people;  there are 239,081 (84.01\%) \textit{white} people, hence,  we encode \textit{race} as a binary attribute for our analysis.
\begin{itemize}
    \item \emph{sex} = \{\textit{male, female}\}. The dataset is slightly imbalanced towards female instances, the \textit{male:female} ratio is 136,447:148,109 (48\%:52\%).
    \item  \emph{race} = \{\textit{white, non-white}\}. The dataset is dominated by white people, the \textit{white:non-white} ratio is 239,081:29,239 (86\%:14\%).
\end{itemize}

\textbf{Class attribute:} The class attribute is \emph{income} $\in \{\leq50K, >50K\}$ indicating whether an individual makes less or more than 50K. The positive class is ``$>50K$''. The dataset is very imbalanced with an IR $1:15.30$ (positive:negative). 

\begin{figure*}[h!]
  \centering
  \includegraphics[width=0.8\linewidth]{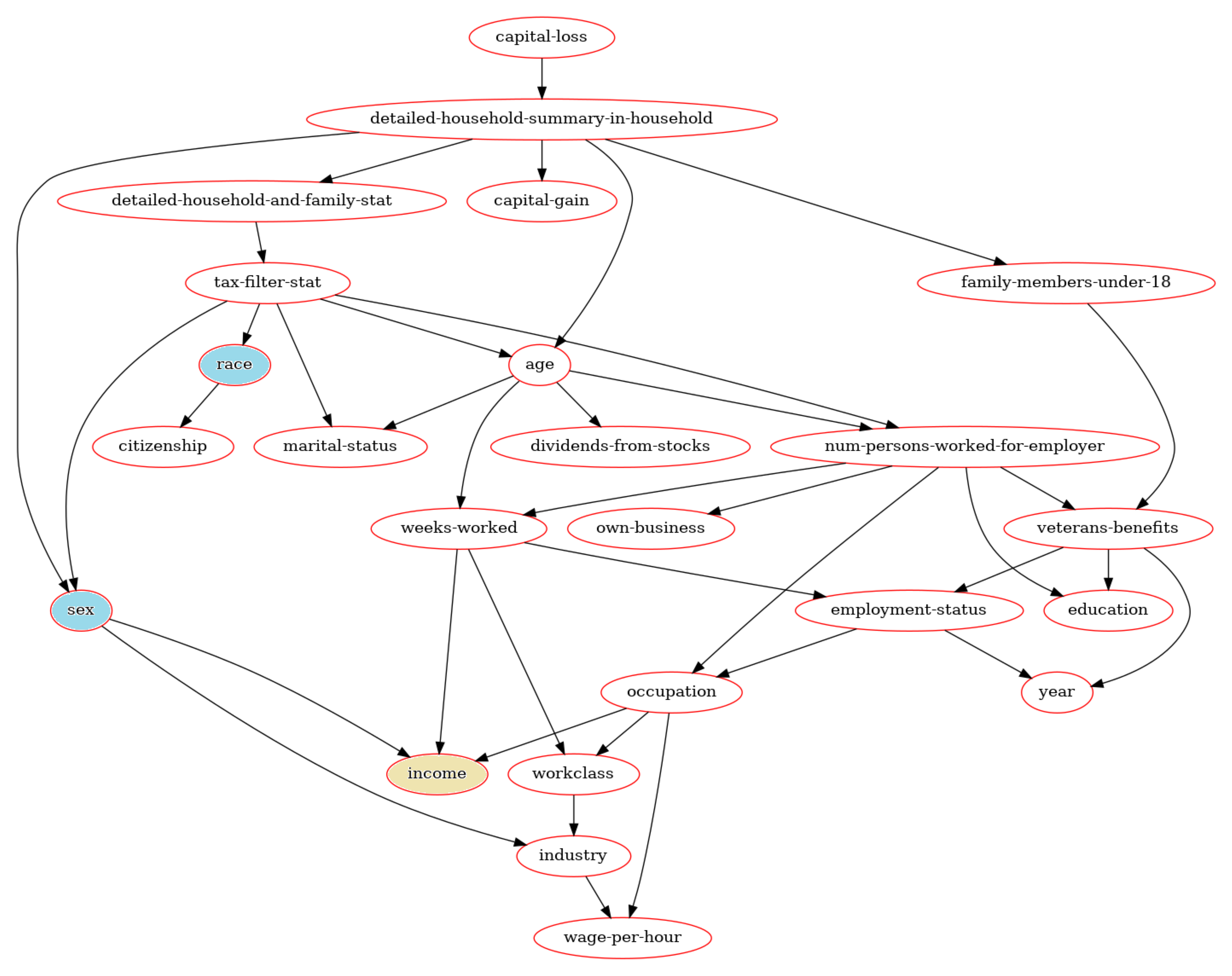}
  \caption{KDD Census-Income: Bayesian network (class label: \textit{income}, protected attributes: \textit{sex, race})}
  \label{fig:KDD-NB}
\end{figure*}
\noindent\textbf{Bayesian network:} To generate the Bayesian network, we encode the following attributes: \textit{age} = \{$\leq$25, 26-60, $>$60\}; \textit{wage-per-hour} = \{$\leq$500, 501-1000, $>$1000\}; \textit{industry} = \{$\leq$30, $>$30\}; \textit{occupation} = \{$\leq$10, $>$10\};  \textit{capital-gain} = \{$\leq$500, $>$500\}; \textit{capital-loss} = \{$\leq$500, $>$500\}; \textit{dividends-from-stocks} = \{$\leq$500, 501-2000, $>$2000\}; \textit{num-persons-worked-for-employer} = \{0, $>$0\}; \textit{weeks-worked-in-year} = \{$\leq$26, 27-51, 52\}. The ranges of encoded attributes are chosen to ensure each group has values. To reduce the complexity, we eliminate these attributes: \textit{enroll-in-edu-inst-last-wk, major-industry, major-occupation} since they have a very low correlation with other features. Also, for efficiency purposes, we generate the Bayesian network on a randomly selected 10\% sample of the dataset rather than on the complete dataset. The learned Bayesian network is shown in Figure~\ref{fig:KDD-NB}; the class label \textit{income} is set as a leaf node.

\begin{figure}[h!]
  \centering
  \includegraphics[width=0.45\linewidth]{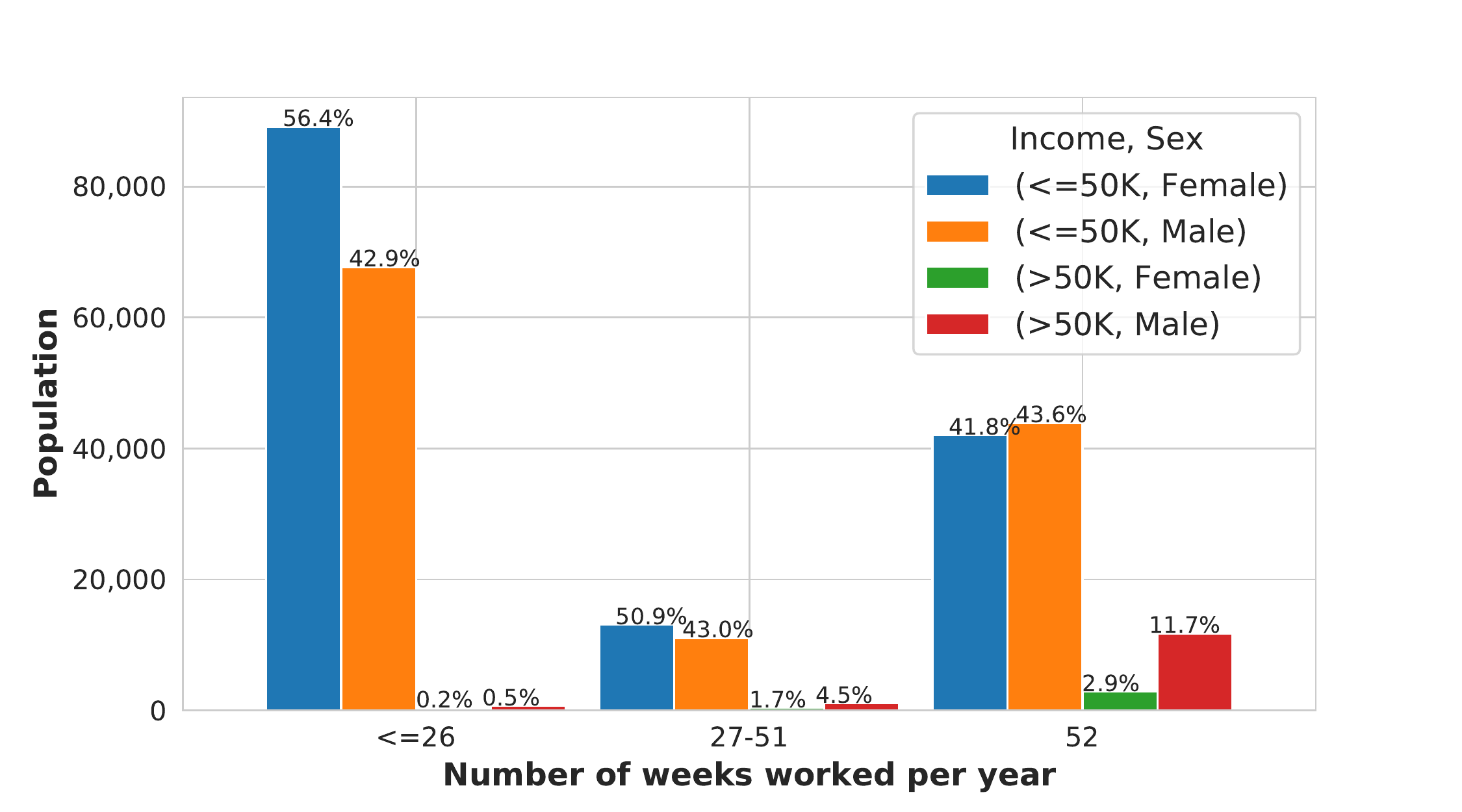}
  \caption{KDD Census-Income: relationship of \textit{the number of weeks worked per year} and \textit{income} w.r.t \textit{sex}}
  \label{fig:kdd_weeks_income}
\end{figure}

As shown in Figure~\ref{fig:KDD-NB}, \textit{income} is conditionally dependent on \textit{sex},  \emph{occupation} and the number of week worked in year (\textit{weeks-worked}) attributes. Regarding \textit{sex} attribute, females are largely underrepresented in the high income group, consisting of 13,691 males ($\sim$10.03\% of the male population) and only 3,711 females ($\sim$2.51\% of the female population).
Regarding the number of weeks worked per year and \textit{income}, as shown in Figure~\ref{fig:kdd_weeks_income}, women tend to do part-time jobs, i.e., the number of weeks worked per year is less than 26. In addition, women earn less money than men even though they all work 52 weeks per year. That is shown by the number of men with high income is approximately five times more than the number of women.

\begin{figure}[h!]
  \centering
  \includegraphics[width=\linewidth]{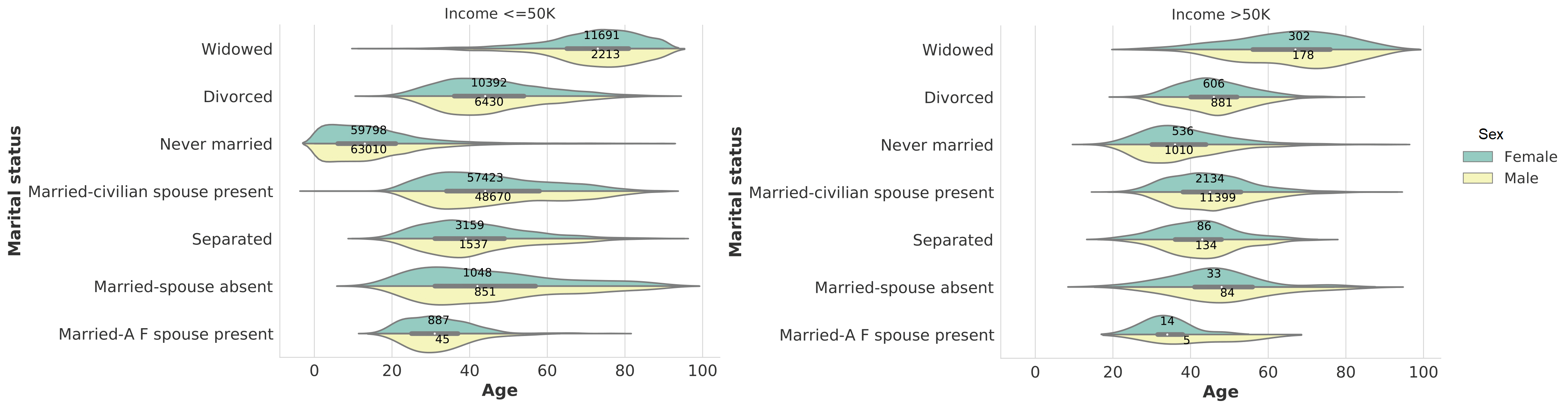}
  \caption{KDD Census-Income: relationship of \textit{marital status}, \textit{age}, \textit{sex} and \textit{income}}
  \label{fig:kdd_violin}
\end{figure}

As mentioned, \textit{race} could also be considered as the protected attribute.
Based on the data, the income of \textit{non-white} people is significantly different from the income of the \textit{white} group. Only 3.2\% of the \textit{non-white} group have an income above 50K, compared to 6.7\% for the \textit{white} group. Furthermore, since \textit{age} has a conditional dependence on \textit{marital-status} attribute, we investigate the relationship between these attributes, the protected attribute \textit{sex} and the class label \textit{income} in Figure~\ref{fig:kdd_violin}. 
As shown in this figure, males comprise the majority of the high-income group, especially for certain population segments like the \textit{Married-civilian spouse present} segment where the number of males is 5 times higher than that of females. Interestingly, the number of widows is 1.7 times higher than the number of widowers in terms of high income. Regarding the \textit{age} effect, most people have a high income when they are over 40 years old. With respect to the protected attributes, there is no edge between \textit{race} and \textit{sex}, which suggests the researchers should perform their fairness-aware models on both these protected attributes.

\subsubsection{German credit dataset}
\label{subsubsec:german-credit}
The German credit\footnote{https://archive.ics.uci.edu/ml/datasets/statlog+(german+credit+data)} dataset~\cite{dua2017} consists of samples of bank account holders. The dataset is used for risk assessment prediction, i.e., to determine whether it is risky to grant credit to a person or not. 
The dataset is frequently employed in fairness-aware learning researches (Appendix~\ref{sec:citation}).

\noindent\textbf{Dataset characteristics:} The dataset contains only 1,000 instances without any missing values. Each sample is described by 13 categorical, 7 numerical and 1 binary attributes. An overview of all attributes is presented in Table~\ref{tbl:german_attribute}. Attribute \textit{personal-status-and-sex} contains information of marital status and the gender of people. We disentangle gender from personal status and create two separate attributes:  \textit{marital-status} and \textit{sex}. The original \textit{personal-status-and-sex} attribute is omitted from further analysis.
\begin{table*}[h!]
\caption{German credit: attributes characteristics}
\label{tbl:german_attribute}
\begin{adjustbox}{width=1\linewidth}
    \begin{tabular}{llccl}
        \hline
        \multicolumn{1}{c}{\textbf{Attributes}} & \multicolumn{1}{c}{\textbf{Type}} & 
        \multicolumn{1}{c}{\textbf{Values}} & 
        \multicolumn{1}{c}{\textbf{\#Missing values}} &
        \multicolumn{1}{c}{\textbf{Description}}
        \\ \hline
checking-account  & Categorical &  4 &  0 & The status of existing checking account \\
duration & Numerical &   [4 - 72] & 0 & The duration of the credit (month)\\
credit-history  & Categorical &  5  & 0 & The credit history\\
purpose  & Categorical &  10 &  0 & {Purpose (car, furniture, education, etc.)}\\
credit-amount & Numerical &   [250 - 18,424] & 0 & Credit amount \\
savings-account  & Categorical &  5  & 0 & Savings account/bonds\\
employment-since  & Categorical &  5  & 0 & Present employment since\\
installment-rate & Numerical &   [1 - 4] & 0 & The installment rate in percentage of disposable income\\
personal-status-and-sex  & Categorical &  4  & 0 & The personal status and sex\\
other-debtors  & Categorical &  3 &  0 & Other debtors/guarantors\\
residence-since & Numerical &   [1 - 4] & 0 & Present residence since\\
property  & Categorical &  4 &  0 & Property\\
age & Numerical &   [19 - 75] & 0 & The age of the individual\\
other-installment  & Categorical &  3  & 0 & Other installment plans\\
housing  & Categorical &  3 &  0 & Housing {(rent, own, for free)}\\
existing-credits & Numerical &   [1 - 4] & 0 & Number of existing credits at this bank \\
job  & Categorical &  4 &  0 & Job {(unemployed, (un)skilled, management)}\\
number-people-provide-maintenance-for & Numerical &   [1 - 2] & 0 & Number of people being liable to provide maintenance for\\
telephone  & Binary &  \{Yes, None\}  & 0 & Telephone number\\
foreign-worker  & Binary &  \{Yes, No\}  & 0 & Is the individual a foreign worker? \\
class-label  & Binary &    \{Good, Bad\} & 0 & Class\\
        \hline
    \end{tabular}
\end{adjustbox}

\end{table*}

\noindent\textbf{Protected attributes:} In all studies, \textit{sex} is considered as the protected attribute. \textit{Age} can also be considered as the protected attribute after binarization into  \{\textit{young}, \textit{old}\} by \emph{age} thresholding at 25~\cite{kamiran2009classifying,friedler2019comparative}.
\begin{itemize}
    \item \emph{sex} = \{\textit{male, female}\}. The dataset is dominated by male instances, the ratio of \textit{male:female} is 690:310 (69\%:31\%). The percentage of women identified as \textit{bad} customers is 35.2\% while that of men is only 27.7\%.
    \item \emph{age} = \{\textit{$\leq$25, $>$25}\}: The dataset is dominated by people older than 25 years, the ratio is 810:190 (81\%:19\%).
    We discover that there is a discrimination on the age of customers. There are 42.1\% of \textit{young} people are recognized as \textit{bad} customers while this proportion in \textit{old} people is 27.2\%.
    
\end{itemize}

\textbf{Class attribute:} The class attribute is \emph{class-label} $\in \{good, bad\}$ revealing the customer's level of risk. The positive class is ``good''. The dataset is imbalanced with an IR $2.33:1$ (positive:negative). 

\begin{figure*}[!htb]
  \centering
  \includegraphics[width=0.8\linewidth]{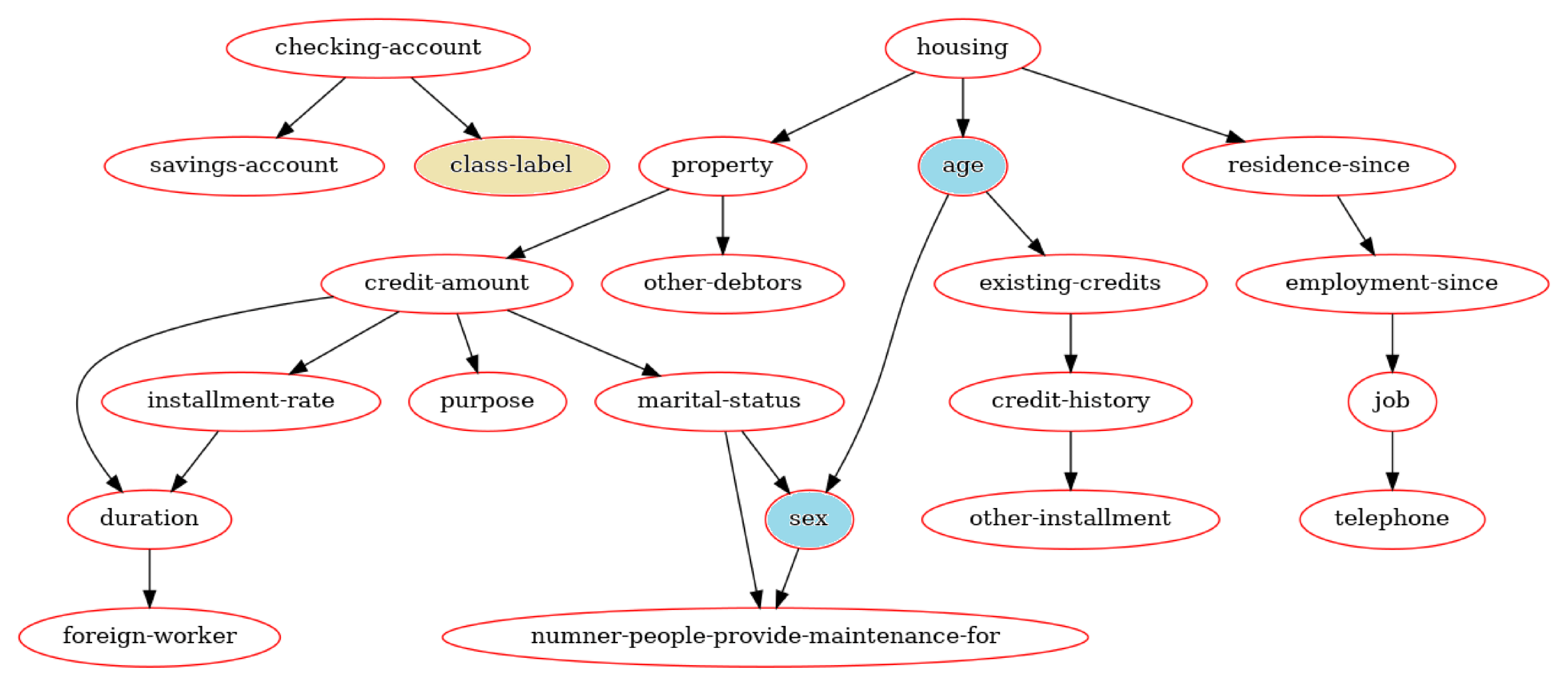}
  \caption{German credit: Bayesian network (class label: \textit{class-label}, protected attributes: \textit{sex, age})}
  \label{fig:German_NB}
\end{figure*}

\noindent\textbf{Bayesian network:} 
We transform the numerical attributes into categorical as follows: 
\textit{duration} = \{$\leq$6, 7-12, $>$12\} (short, medium and long-term); \textit{credit-amount} = \{$\leq$2000, 2000-5000, $>$5000\} (low, medium and high income); \textit{age} = \{$\leq$25, $>$25\}. 
The extracted Bayesian network is shown in Figure~\ref{fig:German_NB}; \textit{class-label} is set as a leaf node.

\begin{figure} [ht!]
     \centering
     \begin{subfigure}[b]{0.48\linewidth}
         \centering
         \includegraphics[width=1\linewidth]{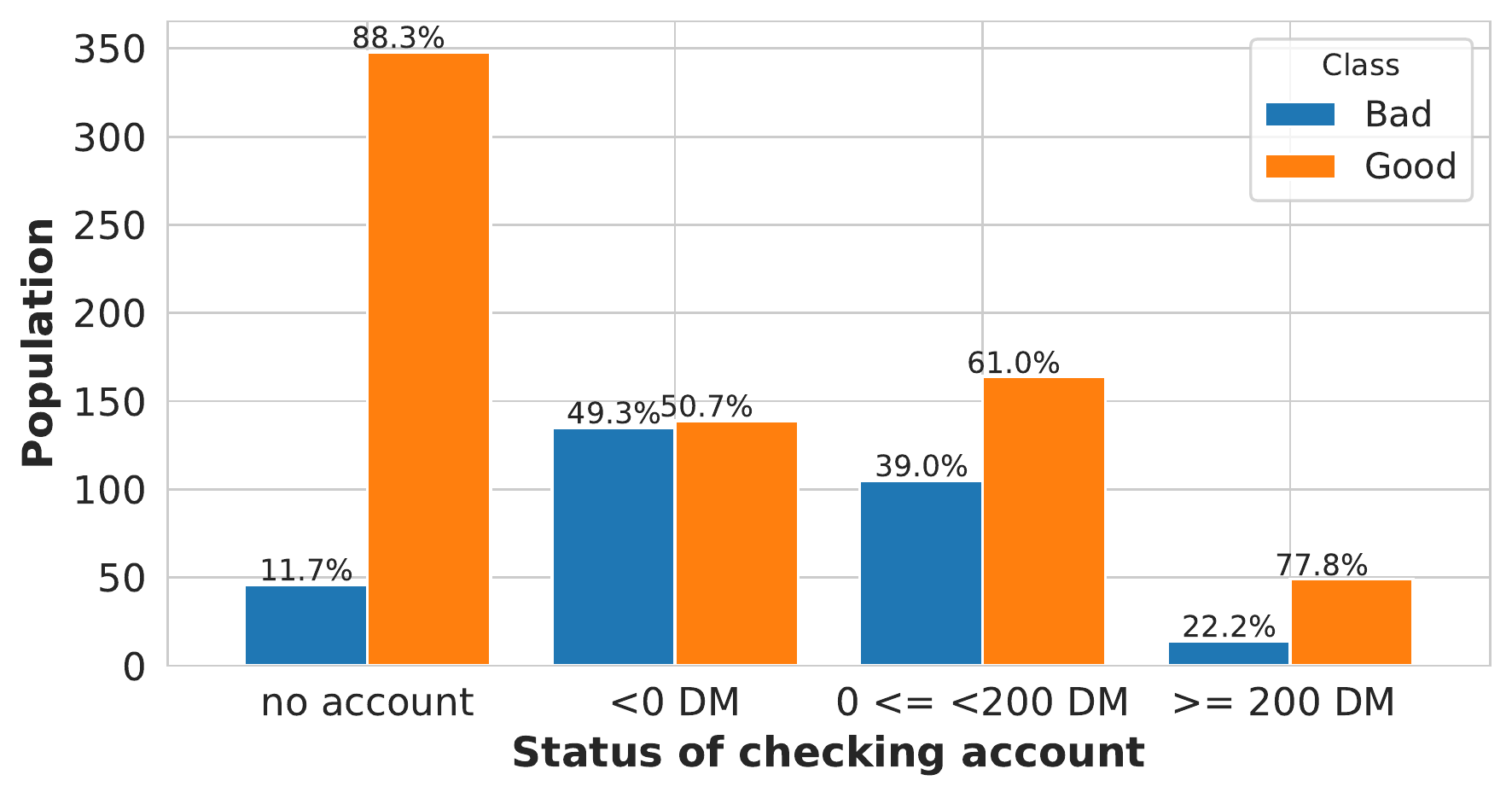}
         \caption{Distribution of \textit{class label} on \textit{status of checking account}}
         \label{fig:German_checking_account_class_label}
     \end{subfigure}
     \quad
     \begin{subfigure}[b]{0.48\linewidth}
         \centering
         \includegraphics[width=1\linewidth]{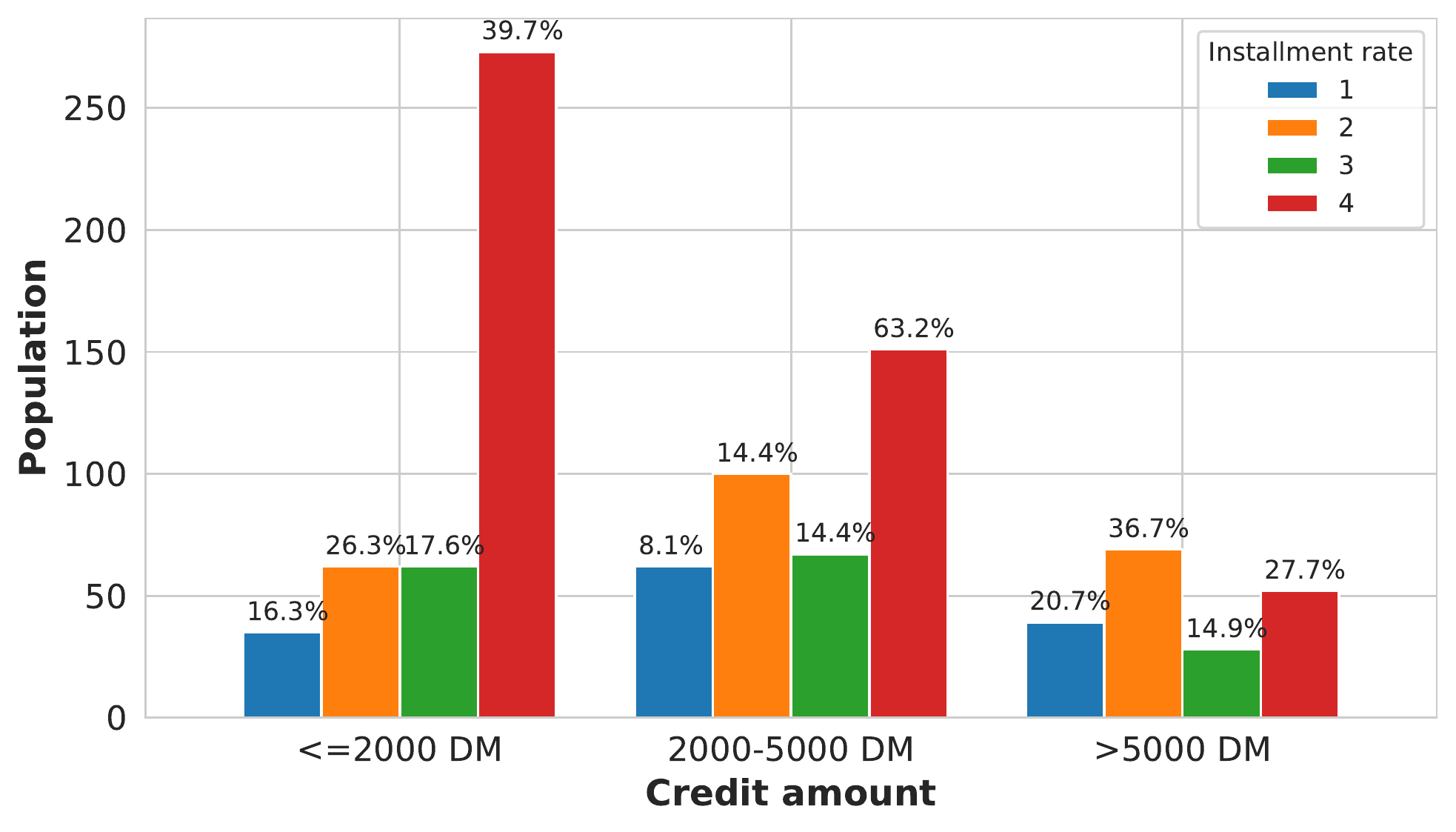}
         \caption{Relationship between \textit{credit amount} and \textit{installment rate}}
         \label{fig:German_installment_credit}
     \end{subfigure}
     \caption{German credit: relationships of class label and attributes}
     \label{fig:German_credit}
\end{figure}

The Bayesian network consists of two disconnected components.
First, \textit{class-label} is conditionally dependent on the \textit{checking-account} attribute. We investigate in more detail this relationship in Figure~\ref{fig:German_checking_account_class_label}. As we can see, a very high proportion of people, i.e., 88.3\%, having no checking account is identified as the \textit{good} customers while half of the customers having a balance less than 0 DM ( \textit{Deutsche Mark}) in their checking account are classified as the \textit{bad} customers.

Second, interestingly, \textit{credit-amount}
has a direct effect on many attributes such as \textit{installment-rate, duration}. We discover that people who borrow a great amount of money tend to borrow for a long period. For example, 93.6\% of interviewees make a loan of more than 5000 DM with a loan duration of more than 12 months.  
As illustrated in Figure~\ref{fig:German_installment_credit}, 
the number of customers who have to pay the highest installment rate (visualized as the ``red'' columns) is inversely proportional to the \textit{credit-amount}. Regarding the protected attributes, a direct edge between \emph{sex} and \emph{age} is observed. This is the starting point of the research question ``Does the fairness-aware model obtain fairness w.r.t. \emph{sex} if \emph{age} is chosen as the protected attributes?'' 

\subsubsection{Dutch census dataset}
\label{subsubsec:dutch-census}
The Dutch census dataset~\cite{van20002001} represented aggregated groups of people in the Netherlands for the year 2001. Researchers (Appendix~\ref{sec:citation})
have used Dutch dataset to formulate a binary classification task to predict a person's \textit{occupation} which can be categorized as \textit{high-level} (prestigious) or \textit{low-level} profession.

\noindent\textbf{Dataset characteristics:} The dataset includes {60,420} samples\footnote{https://github.com/tailequy/fairness\_dataset/tree/main/Dutch\_census} where each sample is described by 12 attributes.
An overview of attributes is presented in Table~\ref{tbl:dutch_attributes}.

\begin{table*}[h!]
\caption{Dutch census: attributes characteristics}
\label{tbl:dutch_attributes}
\begin{adjustbox}{width=1\linewidth}
    \begin{tabular}{llccl}
       \hline
        \multicolumn{1}{c}{\textbf{Attributes}} & \multicolumn{1}{c}{\textbf{Type}} & 
        \multicolumn{1}{c}{\textbf{Values}} & 
        \multicolumn{1}{c}{\textbf{\#Missing values}}&
        \multicolumn{1}{c}{\textbf{Description}}
        \\ \hline
sex  & Binary &    \{Male, Female\} & 0 & The biological sex of the person \\
age & Categorical & 12   & 0 & {The age group of the person (0-4, 5-9, etc.)} \\
household\_position & Categorical & 8  & 0 &  {The relationship to household head (spouse, child, etc.) }\\
household\_size & Categorical & 6 &   0 &  The size of the household the person belongs to\\
prev\_residence\_place & Binary & \{Netherlands, non-Netherlands\}   & 0  & The place of the person's residence prior to the Census\\
citizenship & Categorical & 3 &   0 & The person's citizenship status\\
country\_birth & Categorical & 3   & 0 & Whether the person was born in the Netherlands or elsewhere\\
edu\_level & Categorical & 6 &   0 & The person's level of educational attainment\\
economic\_status & Categorical & 3   & 0 & The person's economic status (class of worker)\\
cur\_eco\_activity & Categorical & 12   & 0 & The current economic activity\\
marital\_status & Categorical & 4 &   0 & The person's current marital status according to law or custom\\
occupation & Binary &   \{0, 1\} & 0 & The person's occupation {(0: low-level, 1: high-level)}\\
        \hline
    \end{tabular}
\end{adjustbox}
\end{table*}

\begin{figure}[h!]
    \centering
    \includegraphics[width=0.6\linewidth]{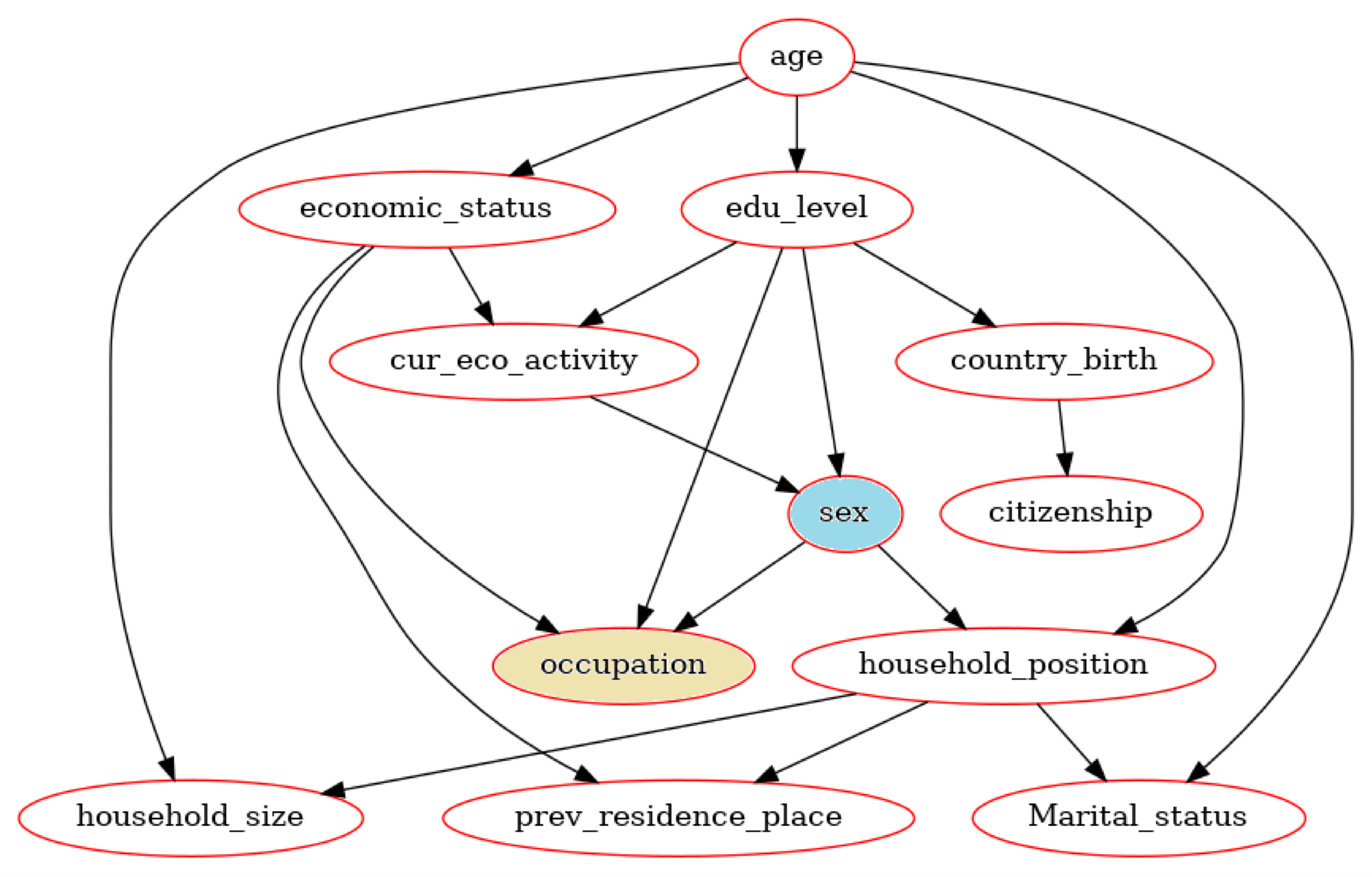}
    \caption{Dutch census: Bayesian network (class label: \textit{occupation}, protected attribute: \textit{sex})}
    \label{fig:dutch_NB}
\end{figure}

\noindent\textbf{Protected attributes:} In the related work, they consider attribute \textit{sex} = \{\textit{male, female}\} as the protected attribute, \textit{male:female} ratio is 30,147:30,273 (49.9\%:50.1\%).

\textbf{Class attribute:} The class attribute is \emph{occupation} $\in \{0, 1\}$ demonstrating if an individual has a prestigious profession or not. The positive class is \emph{1 (high-level)}. This is a fairly balanced dataset in our survey with an IR $1:1.10$ (positive:negative). 

\begin{figure}[h!]
  \centering
  \includegraphics[width=0.45\linewidth]{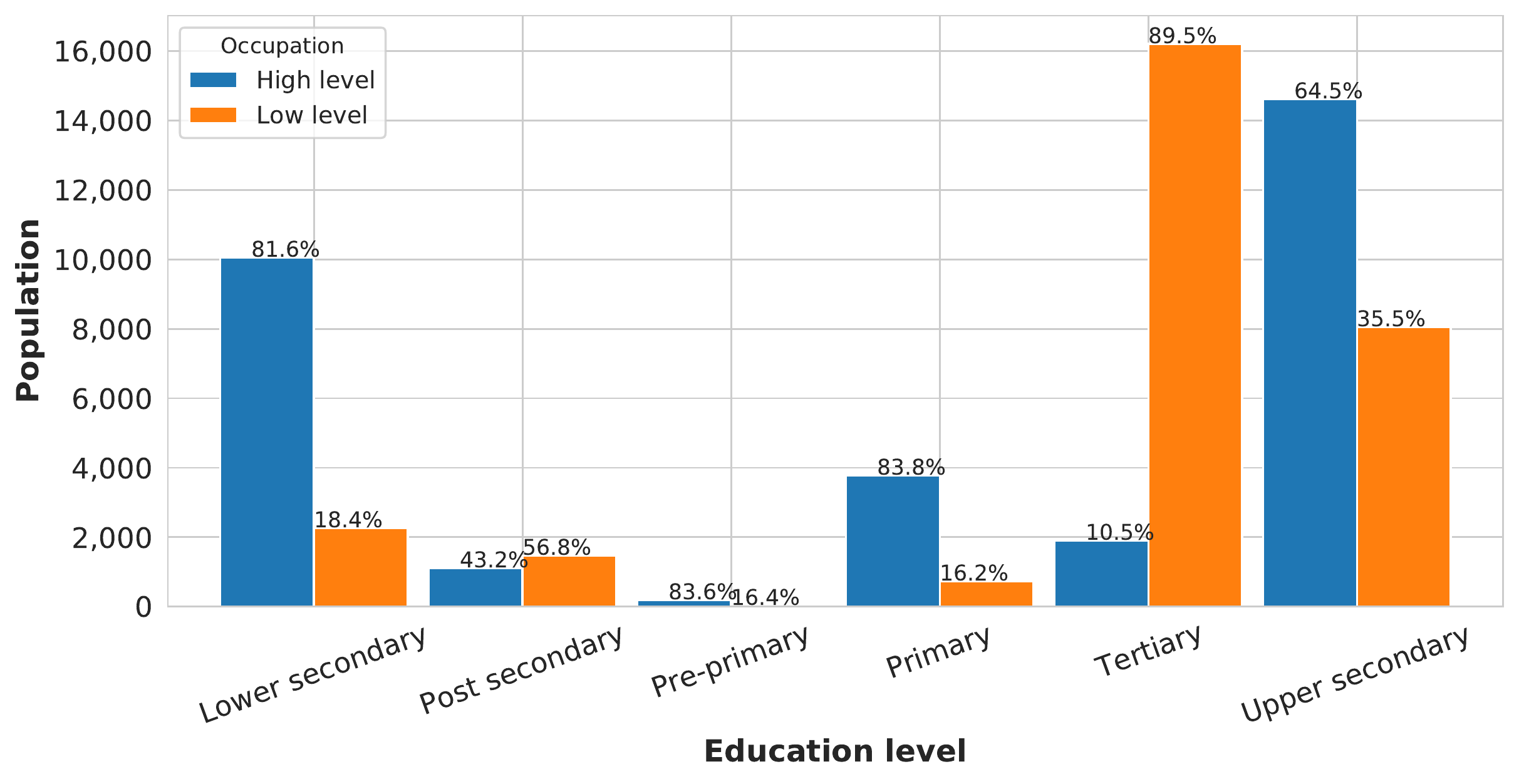}
  \caption{Dutch census: relationship between \textit{education level} and \textit{occupation}}
  \label{fig:dutch_edu_occupation}
\end{figure}

\noindent\textbf{Bayesian network:} We use all attributes in the dataset to generate the Bayesian network. As illustrated in Figure~\ref{fig:dutch_NB}, the leaf node \textit{occupation} is conditionally dependent on \textit{economic status, education level} and \textit{sex} attributes. In fact, 62.6\% of males (18,860 out of 30,147) have a high-level occupation, while this proportion on females group is only 32.7\%. In addition, people with high education are doing prestigious jobs and vice versa, as depicted in Figure~\ref{fig:dutch_edu_occupation}. For example, 89.5\% of people having \textit{tertiary} level are working in high-level jobs while around 80\% of people with \textit{lower secondary} degrees are doing low-level work. Interestingly, \textit{age} has a direct effect on many attributes. 

\subsubsection{Bank marketing dataset}
\label{subsubsec:bank-marketing}
The bank marketing\footnote{https://archive.ics.uci.edu/ml/datasets/Bank+Marketing} dataset~\cite{moro2014data} is related to the direct marketing campaigns of a Portuguese banking institution from 2008 to 2013.
There is a variety of researchers investigating this dataset in their studies (Appendix~\ref{sec:citation}).
The classification goal is to predict whether a client will make a deposit subscription or not.

\noindent\textbf{Dataset characteristics:}
The dataset comprises 45,211 samples, each with 6 categorical, 4 binary and 7 numerical attributes, as summarized in Table~\ref{tbl:bank_attributes}.

\begin{table*}[h!]
\caption{Bank marketing: attributes characteristics}
\label{tbl:bank_attributes}
\begin{adjustbox}{width=1\linewidth}
    \begin{tabular}{llccl}
        \hline
        \multicolumn{1}{c}{\textbf{Attributes}} & \multicolumn{1}{c}{\textbf{Type}} & 
        \multicolumn{1}{c}{\textbf{Values}} & 
        \multicolumn{1}{c}{\textbf{\#Missing values}} &
        \multicolumn{1}{c}{\textbf{\#Missing values}}
        \\ \hline
age & Numerical &   [18 - 95] & 0 & The age of the client \\
job  & Categorical &  12  & 0 & The type of job {(admin, self-employed, technician, ect.)}\\
marital  & Categorical &  3  & 0 & The marital status \\
education  & Categorical &  4  & 0 & The education level\\
default  & Binary &    \{Yes, No\} & 0  & Has the credit in default? \\
balance & Numerical &   [-8,019 - 102,127] & 0 & The balance of this client's account\\
housing  & Binary &    \{Yes, No\}& 0 & Has a housing loan?\\
loan  & Binary &    \{Yes, No\} & 0 & Has a personal loan?\\
contact  & Categorical &  3 &  0 & The contact communication type\\
day & Numerical &   [1 - 31] & 0 & The last contact day of the week\\
month  & Categorical &  12  & 0 & The last contact month of the year\\
duration & Numerical &   [0 - 4,918] & 0 & The last contact duration, in seconds \\
campaign & Numerical &   [1 - 63] & 0 &The number of contacts performed during this campaign and for this client\\
pdays & Numerical &   [-1 - 871] & 0 & The number of days that passed by after the client was last contacted\\
previous & Numerical &   [0 - 275] & 0 & The number of contacts performed before this campaign and for this client\\
poutcome  & Categorical &  4  & 0 & The outcome of the previous marketing campaign\\
y (class)  & Binary &   \{Yes, No\} & 0 & Has the client subscribed a term deposit?\\
        \hline
    \end{tabular}
\end{adjustbox}
\end{table*}

\noindent\textbf{Protected attributes:}
In the literature, \textit{marital-status} can be considered as the protected attribute~\cite{backurs2019scalable,hu2020fairnn,chierichetti2017fair,ziko2019variational,berafair2019}. Besides, in several studies \cite{krasanakis2018adaptive,zafar2015fairness,fish2016confidence}, they consider \textit{age} as the protected attribute which is binary separated into people who are between the age of 25 to 60 years old and less than 25 or more than 60 years old.

\begin{itemize}
    \item \emph{age} = \{\textit{25-60, $<$25 or $>$60}\}: the dataset is dominated by people from 25 to 60 years old, the ratio of \textit{``25-60'':``$<$25 or $>$60''} is 43,214:1,997 (95.6\%:4.4\%).
    \item  \emph{marital} = \{\textit{married, non-married}\}: \emph{married} group is the majority with the ratio of \textit{married:non-married} is 27,214:17,997 (60.2\%: 39.8\%).
\end{itemize}

\textbf{Class attribute:} The class attribute is \emph{y} $\in \{Yes, No\}$ presenting whether a customer will subscribe a term deposit or not. The positive class is ``Yes''. The dataset is imbalanced with an IR $1:7.55$ (positive:negative). 

\begin{figure}[!htb]
  \centering
  \includegraphics[width=0.65\linewidth]{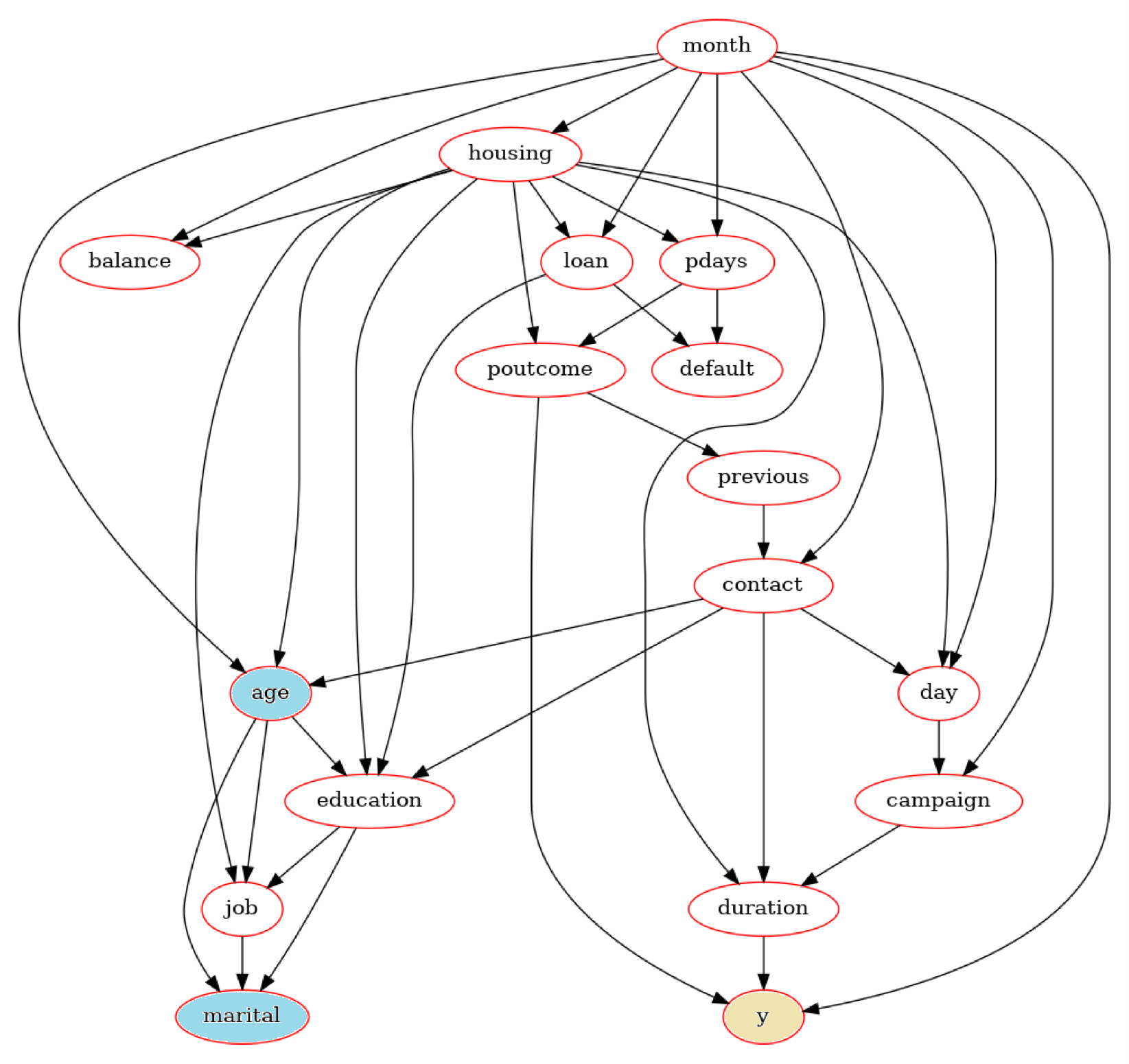}
  \caption{Bank marketing: Bayesian network (class label: \textit{y}, protected attributes: \textit{age, marital})}
  \label{fig:bank_NB}
\end{figure}

\begin{figure}[h!]
  \centering
  \includegraphics[width=0.45\linewidth]{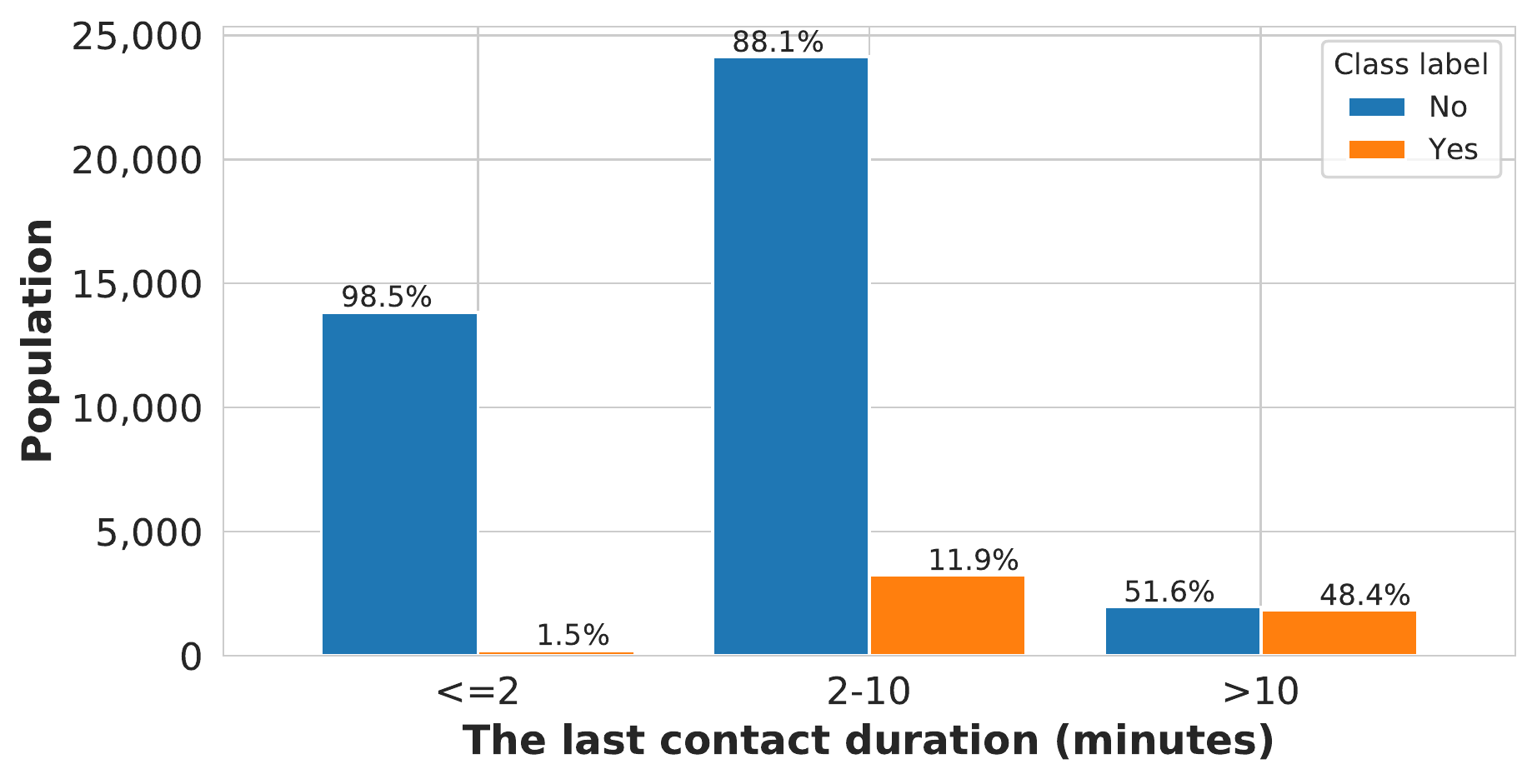}
  \caption{Bank marketing: Relationship between the last contact duration and class label}
  \label{fig:bank_duration_y}
\end{figure}

\noindent\textbf{Bayesian network:} We perform a pre-processing step to transfer the numerical attributes into categorical: \textit{job} = \{\textit{blue-collar, management-service, other}\}; \textit{balance} = \{0, $>$0\}; \textit{day} = \{$\leq$15, $>$15\}; \textit{duration} = \{$\leq$120, 121-600, $>$600\}; \textit{campaign} = \{$\leq$1, 2-5, $>$5\}; \textit{pdays} = \{$\leq$30, 31-180, $>$180\}; \textit{previous} = \{0, 1-5, $>$5\}. Figure~\ref{fig:bank_NB} visualizes the Bayesian network of the Bank marketing dataset. 
The class label \textit{y}, as illustrated in Figure~\ref{fig:bank_NB}, is conditionally dependent on \textit{poutcome, month} and \textit{duration} attributes. An insight about the relationship between the last contact \textit{duration} and class label \textit{y} is described in Figure~\ref{fig:bank_duration_y}. The ratio of clients who will make a deposit subscription is proportional to the duration of the last contact. When the talk is taken place in less than 2 minutes, 98.5\% of people will not make the deposit subscription. However, if a marketing staff can maintain the talk with customers over 10 minutes, 48.4\% of customers will say ``\textit{Yes}''. Interestingly, in the Bayesian network, both protected attributes \textit{age} and \textit{marital} have no effect on the class label \textit{y}. However, the protected attributes are connected together by an in-direct edge, which could be a reason for a similar accuracy of fairness-aware models of the related work \cite{hu2020fairnn} and \cite{krasanakis2018adaptive}.

\subsubsection{Credit card clients dataset}
\label{subsubsec:credit_card_clients}
The credit card clients\footnote{https://archive.ics.uci.edu/ml/datasets/default+of+credit +card+clients} dataset~\cite{yeh2009comparisons} investigated the customers' default payments in Taiwan in October 2005. The goal is to predict whether a customer will face the default situation in the next month or not. The data have been used for default payment prediction in several studies (Appendix~\ref{sec:citation}).

\noindent\textbf{Dataset characteristics:} The dataset includes 30,000 customers described by 8 categorical, 14 numerical and 2 binary attributes, as depicted in Table~\ref{tbl:credit_attributes}. There is no missing value in the dataset.
\begin{table*}[h!]
\caption{Credit card clients: attributes characteristics}
\label{tbl:credit_attributes}
\centering
\begin{adjustbox}{width=0.9\linewidth}
    \begin{tabular}{llccl}
        \hline
        \multicolumn{1}{c}{\textbf{Attributes}} & \multicolumn{1}{c}{\textbf{Type}} & 
        \multicolumn{1}{c}{\textbf{Values}} & 
        \multicolumn{1}{c}{\textbf{\#Missing values}} &
        \multicolumn{1}{c}{\textbf{Description}}
        \\ \hline
limit\_bal & Numerical &  [10,000 - 1,000,000] & 0 & The amount of the given credit (New Taiwan dollar)\\
sex & Binary &  \{Male, Female\} & 0 & The biological sex of the client\\
education & Categorical & 7   & 0 & {The education level} \\
marriage & Categorical & 3   & 0 & The marital status\\
age & Numerical &   [21 - 79] & 0 & The age of the client (year)\\
pay\_0 & Categorical   & 11 & 0 & The repayment status in September 2005 {(pay duly, delay 1 month, etc.)}\\
pay\_2 & Categorical   & 11 & 0 & The repayment status in August 2005\\
pay\_3 & Categorical   & 11 & 0 & The repayment status in July 2005\\
pay\_4 & Categorical   & 11 & 0 & The repayment status in June 2005\\
pay\_5 & Categorical   & 10 & 0 & The repayment status in May 2005\\
pay\_6 & Categorical   & 10 & 0 & The repayment status in April 2005\\
bill\_amt1 & Numerical   & [-165,580 - 964,511] & 0 & The amount of bill statement in September 2005 \\
bill\_amt2 & Numerical   & [-69,777 - 983,931] & 0 & The amount of bill statement in August 2005 \\
bill\_amt3 & Numerical   & [-157,264 - 1,664,089] & 0 & The amount of bill statement in July 2005\\
bill\_amt4 & Numerical   & [-170,000 - 891,586] & 0 & The amount of bill statement in June 2005\\
bill\_amt5 & Numerical   & [-81,334 - 927,171] & 0 & The amount of bill statement in May 2005 \\
bill\_amt6 & Numerical   & [-339,603 - 961,664] & 0 & The amount of bill statement in April 2005\\
pay\_amt1 & Numerical   & [0 - 873,552] & 0 & The amount paid in September 2005\\
pay\_amt2 & Numerical   & [0 - 1,684,259] & 0 & The amount paid in August 2005\\
pay\_amt3 & Numerical   & [0 - 896,040] & 0 & The amount paid in July 2005\\
pay\_amt4 & Numerical   & [0 - 621,000] & 0 & The amount paid in June 2005\\
pay\_amt5 & Numerical   & [0 - 426,529] & 0 & The amount paid in May 2005\\
pay\_amt6 & Numerical   & [0 - 528,666] & 0 & The amount paid in April 2005\\
default payment& Binary   & \{0, 1\} & 0 & Whether or not the client face the default situation\\
        \hline
    \end{tabular}
\end{adjustbox}

\end{table*}

\noindent\textbf{Protected attributes:}
In the literature, \textit{sex} \cite{deepak2020fair,bechavod2017penalizing,berk2017convex}, \textit{education, marriage}~\cite{deepak2020fair,berafair2019} are considered as the protected attributes.
\begin{itemize}
    \item \emph{sex} = \{\textit{male, female}\}: the dataset is dominated by females, the ratio of \textit{male:female} is 11,888:18,112 (39.6\%:60.4\%).
    \item  \emph{marriage} = \{\textit{married, single, others}\}: \emph{single} group is the majority with the ratio of \textit{married:single:others} is 13,659:15,964:377 (45.5\%:53.2\%:1.3\%).
    \item \emph{education} = \{\textit{graduate school, university, high school, others}\}: \emph{university} is the biggest group with 14,030 (46.8\%) clients.
\end{itemize}

\textbf{Class attribute:} The class attribute is \emph{default payment} $\in \{0, 1\}$ indicating whether a customer will suffer the default payment situation in the next month (1) or not (0). The positive class is \emph{1}. This is an imbalanced dataset with an IR $1:3.52$ (positive:negative). 

\begin{figure*}[!h]
  \centering
  \includegraphics[width=0.7\linewidth]{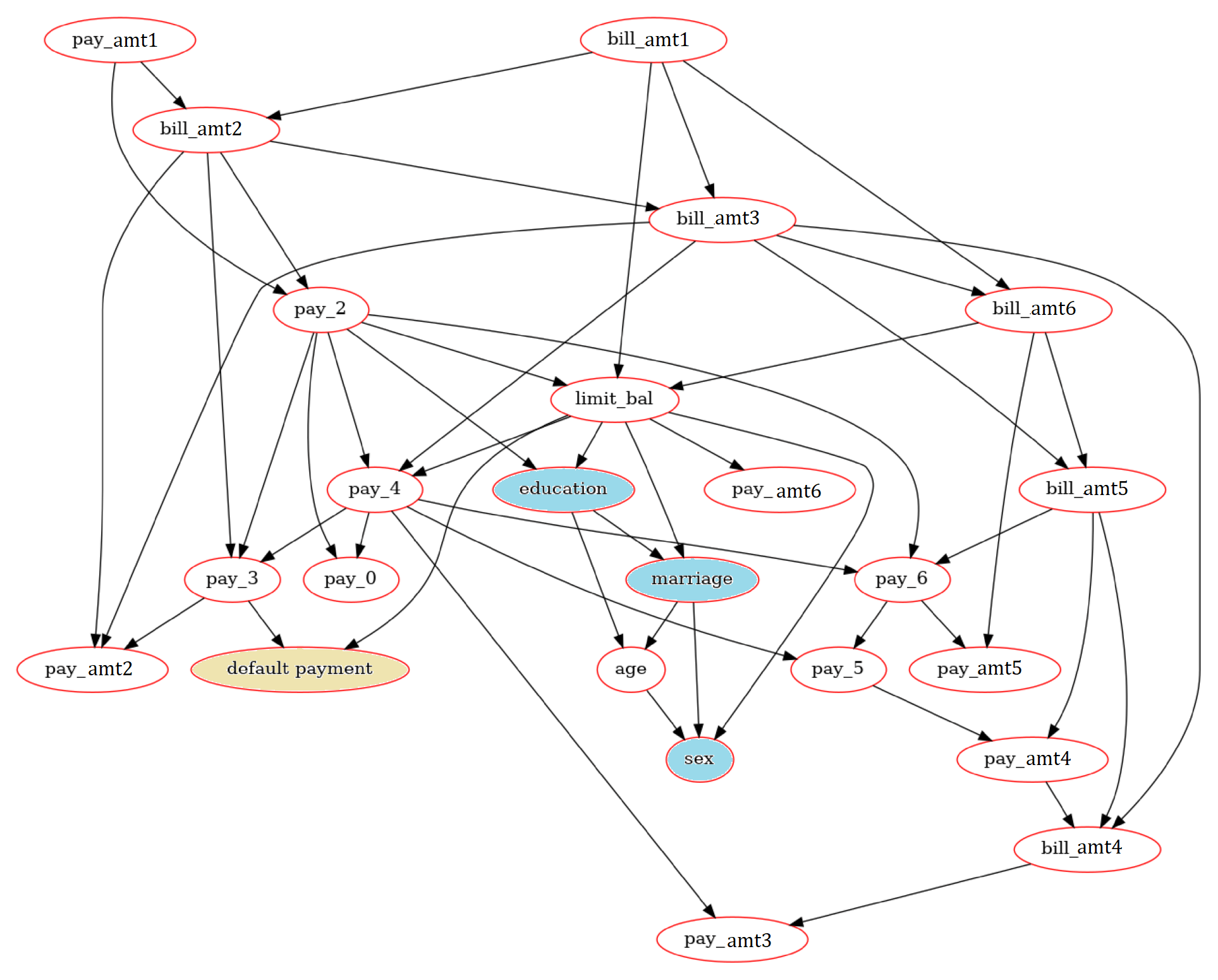}
  \caption{Credit card clients: Bayesian network (class label: \textit{default payment}, protected attributes: \textit{sex, marriage, education})}
  \label{fig:credit_NB}
\end{figure*}

\noindent\textbf{Bayesian network:}
To generate the Bayesian network, we convert the numerical attributes: \textit{age} = \{$\leq$35, 36-60, $>$60\}; the amount of the given credit (\textit{limit\_bal}), the amount of the bill statements \textit{(bill\_amt\_1,$\ldots$, bill\_amt\_6)}, and the amount of the previous payments \textit{(pay\_amt\_1, bill\_1,$\ldots$, pay\_amt\_6}) = \{$\leq$50,000, 50,001-200,000, $>$200,000\} (corresponding to the \textit{low, medium, high} levels); history of the past payments \textit{pay\_0,$\ldots$, pay\_6} = \{\textit{pay duly, 1-3 months, $>$3 months}\}. The Bayesian network is presented in Figure~\ref{fig:credit_NB}. 

\begin{figure}[!ht]
  \centering
  \includegraphics[width=0.45\linewidth]{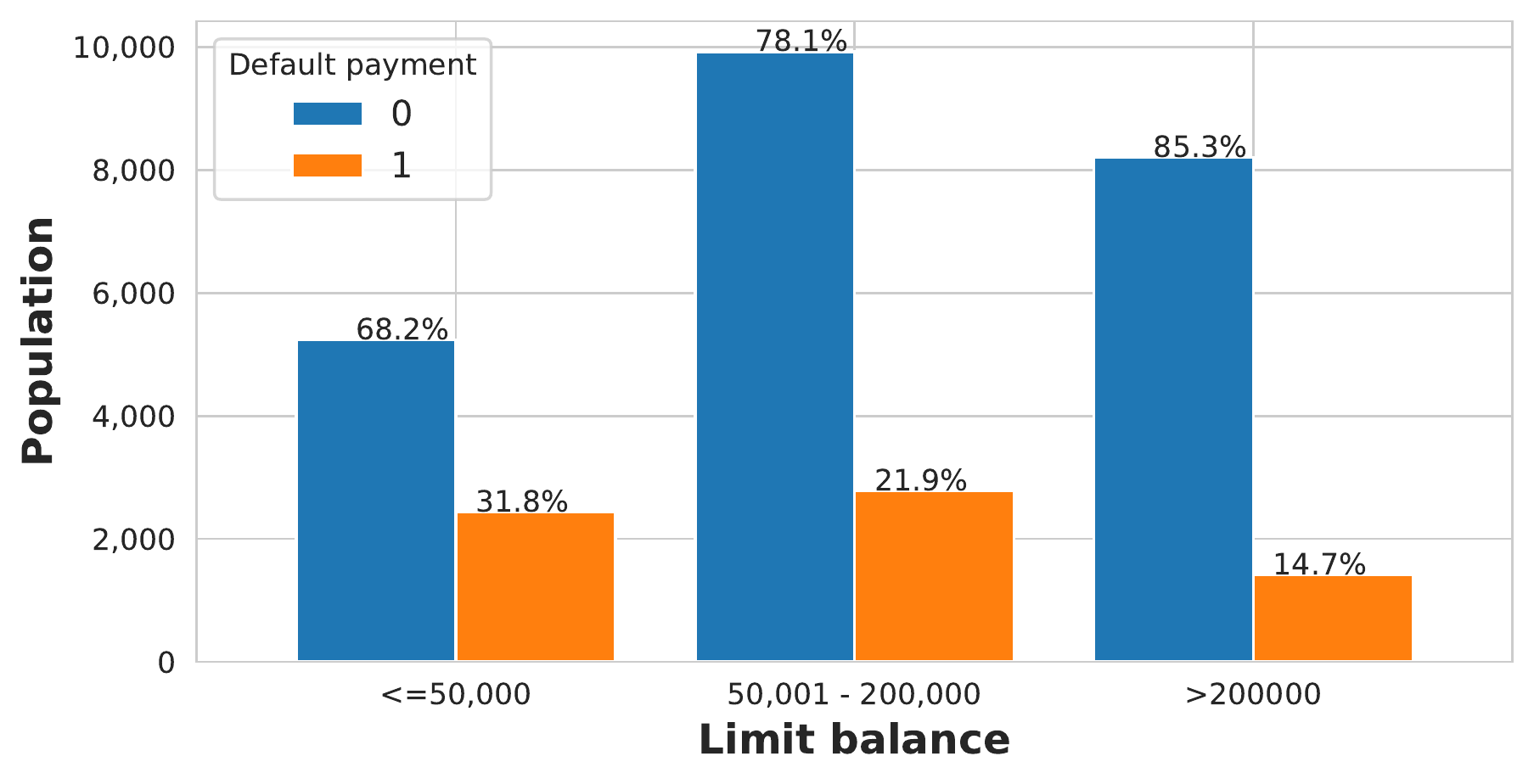}
  \caption{Credit card clients: Relationship between \textit{the credit limit balance} and \textit{default payment}}
  \label{fig:credit_limit_default_payment}
\end{figure}

The class label \textit{default payment} is directly conditionally dependent on the repayment status in July 2005 (attribute \textit{pay\_3}), and the given credit (attribute \textit{limit\_bal}) and indirectly dependent on the amount of bill statements (the attributes with a prefix \textit{bill\_amt}). As demonstrated in Figure~\ref{fig:credit_limit_default_payment}, the ratio of the default payment phenomenon is inversely proportional to the credit limit balance. 
Moreover, we discover that the percentage of males having the default payment in the next month is higher than that of females. In particular, the ratio of males with the default payment is 24.2\% while that of females is only 20.8\%. Interestingly, the protected attributes (\textit{sex, education, marriage}) are conditionally dependent on each other. 

\textbf{Summary of the financial datasets:}
In general, the financial datasets are very diverse as they were collected from several diverse locations (from US, to Taiwan) and at very different time points (from 1994 to 2013). With respect to the collection time, the datasets are pretty old, esp. Adult and KDD census datasets.  
These datasets have been heavily investigated in the related work and under different protected attributes. The most prevalent protected attribute is \emph{sex} followed by \emph{race, age, marriage} and \emph{education}. An interesting observation is that the protected attributes are often related to each other (a strong or weak relationship), e.g., race with education. Due to these dependencies, ensuring fairness for one protected attribute may positively affect fairness for other protected attributes.
Moreover, most of the datasets in this category are imbalanced, with the only exception of the Dutch census dataset which is almost balanced (see Table~\ref{tbl:real_world_datasets}). In terms of class imbalance, datasets demonstrate different imbalance ratios.
         

%% file: realdata_criminology.tex
\subsection{Criminological  datasets}
\label{subsec:criminology}

\subsubsection{COMPAS dataset}
\label{subsubsec:compas}

The COMPAS (Correctional Offender Management Profiling for Alternative Sanctions)~\cite{angwin2016machine} is a recent dataset, compared to the rest of the datasets in our {survey}, which was released by ProPublica\footnote{https://www.propublica.org/datastore/dataset/compas-recidivism-risk-score-data-and-analysis} in 2016 based on the Broward County data (collected from Jan 2013 to Dec 2014). Defendant’s answers to the COMPAS screening survey are used to generate the recidivism risk scores.
The data have been used for crime recidivism risk prediction by a plethora of works (Appendix~\ref{sec:citation}).
\textit{Risk of recidivism} (denoted as \textit{COMPAS recid.})  and \textit{Risk of violent recidivism} (denoted as \textit{COMPAS viol. recid}) subsets are most widely used in the literature. The former has a classification task to predict if an individual is rearrested within two years after the first arrest. The latter predicts if an individual is rearrested for a violent crime within two years.

\noindent\textbf{Dataset characteristics:}
\textit{COMPAS recid.} and \textit{COMPAS viol. recid.} datasets contain 7,214 and 4,743 samples, respectively. Each defendant is described by 52 attributes\footnote{Table~\ref{tbl:COMPAS_attribute} describes attributes used in the Bayesian network and data analysis} (31 categorical, 6 binary, 14 numerical and a null attribute), as shown in Table~\ref{tbl:COMPAS_attribute} and Table~\ref{tbl:COMPAS_attribute_extra} (Appendix~\ref{sec:dataset_characteristics}). 
Missing data is a common phenomenon in both subsets. There are 6,395 rows (88.6\%) containing missing values in the COMPAS recid. subset while this number in the COMPAS viol. revid. subset is 3,748 (79\%). 
Based on \cite{angwin2016machine}, we clean the dataset by removing the missing data, such as \textit{violent\_recid = NULL} or the change date of a crime (attribute \textit{days\_b\_screening\_arrest}) was not within 30 days when he or she was arrested. The cleaned datasets used in our analysis contain 6,172 (COMPAS recid.) and 4,020 (COMPAS viol. recid.) records.

\begin{figure} [h!]
     \centering
     \begin{subfigure}[b]{0.45\linewidth}
         \centering
         \label{fig:compas_recid_race_recid}
         \includegraphics[width=1\linewidth]{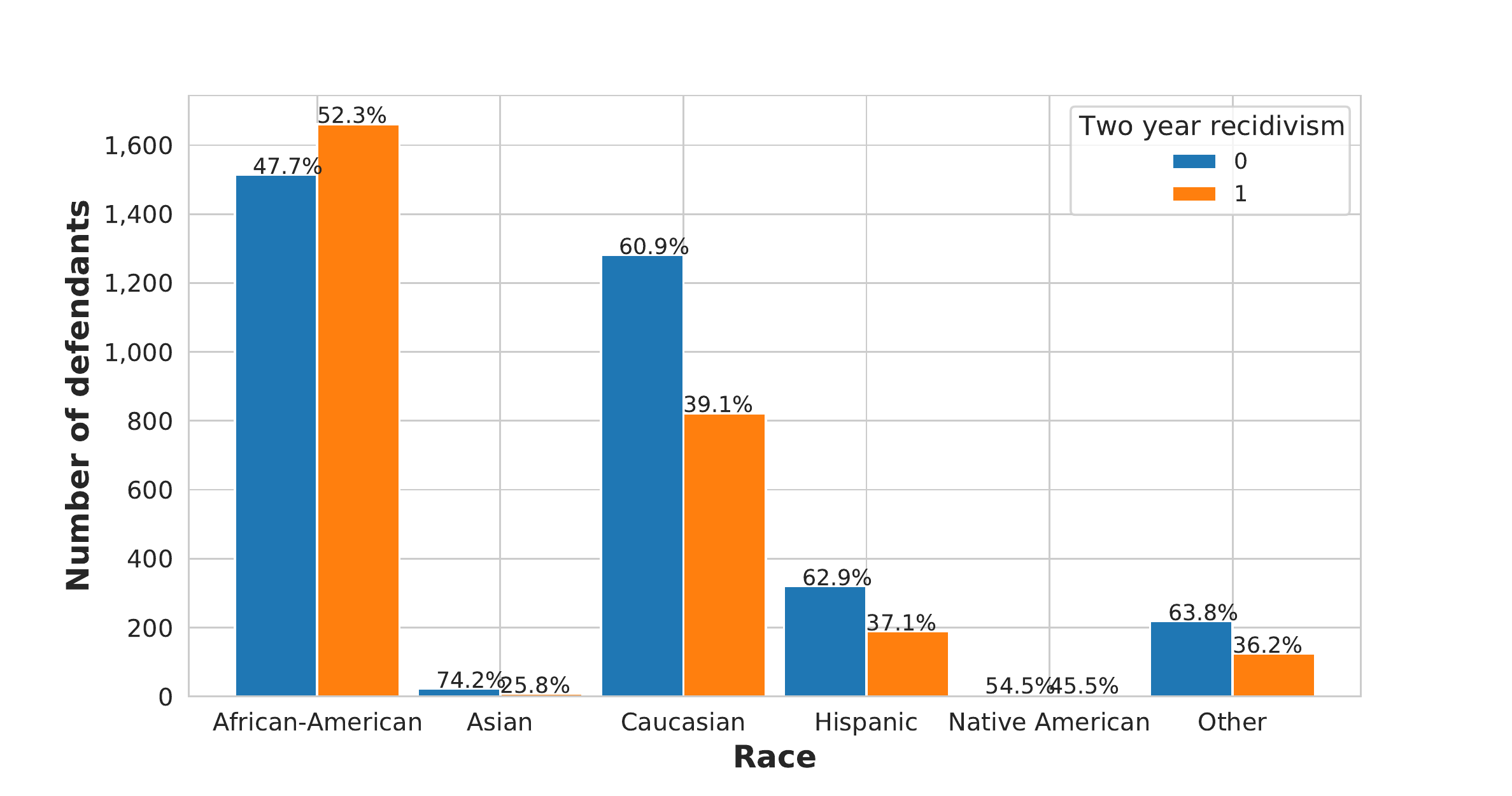}
         \caption{COMPAS recid. subset}    
     \end{subfigure}
     \quad
     \begin{subfigure}[b]{0.45\linewidth}
         \centering
         \includegraphics[width=1\linewidth]{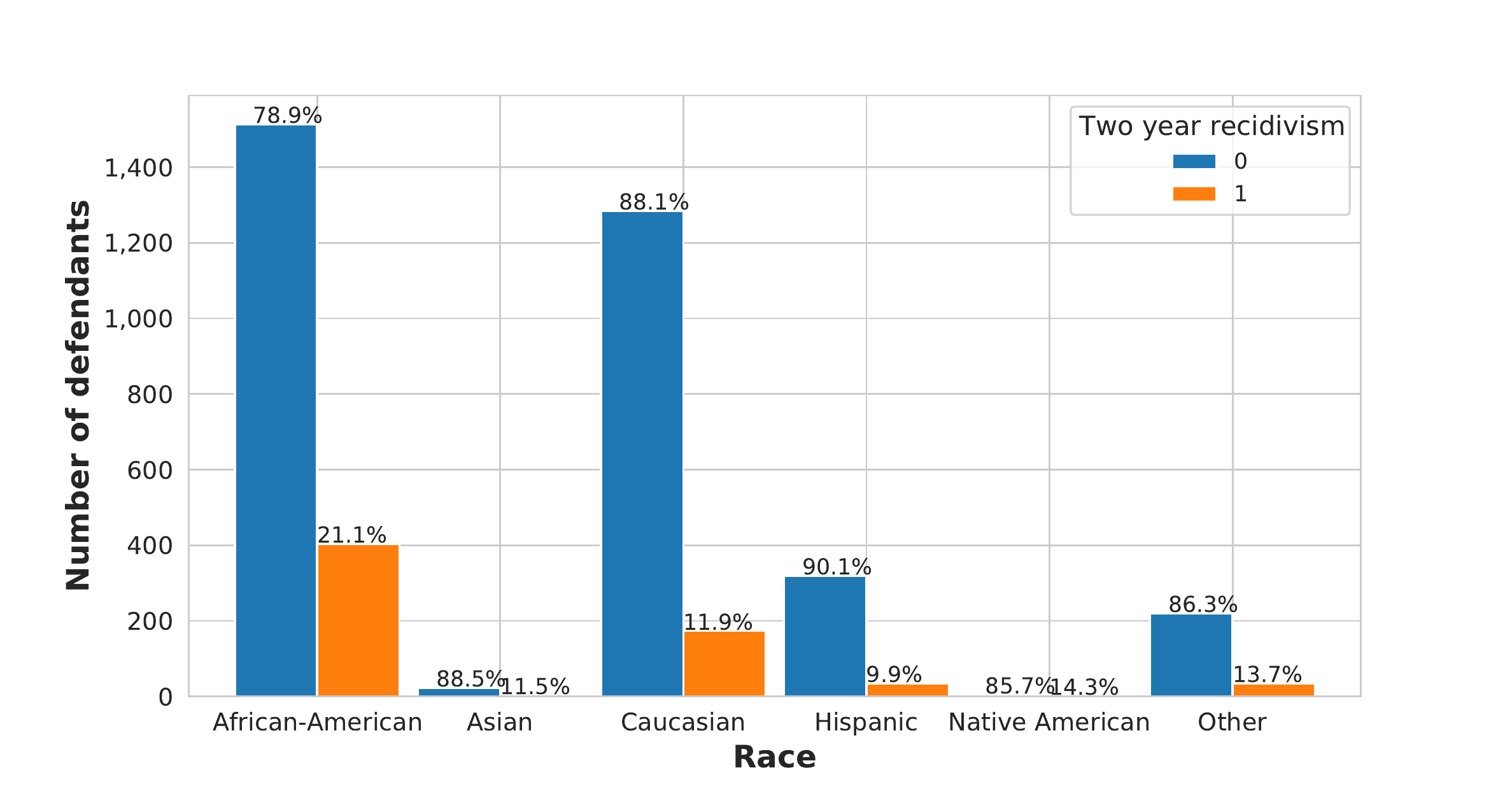}
         \caption{COMPAS viol. recid. subset}
         \label{fig:compas_recid_race_viol}
     \end{subfigure}
     \caption{COMPAS: distribution of \textit{two year recidivism} w.r.t. \textit{race}}
     \label{fig:compas_recid_race}
\end{figure}

\noindent\textbf{Protected attributes:} Typically, \textit{race} is employed as the protected attribute. In both subsets, \emph{black} and \emph{white} are the main races. In the COMPAS recid. subset, the \textit{black:white} ratio is 3,175:2,103 (51.4\%:34\%) (computed on the total number of defendants), while this ratio in the COMPAS viol. recid. subset is 1,918:1,459 (47.7\%:36.3\%). Figure~\ref{fig:compas_recid_race} describes the distribution of defendants w.r.t. \textit{race} attribute. The recidivism rate in the black defendants is higher than that of the white defendants in both subsets.

\emph{Sex} has been also considered as the protected attribute \cite{diana2021minimax,van2021effect,chakraborty2020fairway}. The ratio \textit{male:female} is 4,997:1,175 (81\%:19\%) in the COMPAS recid. subset, while this ratio in the COMPAS viol. recid. subset is 3,179:841 (79.1\%:20.9\%).

\textbf{Class attribute:} The class attribute is \emph{two\_year\_recid} $\in \{0, 1\}$ indicating whether an individual will be rearrested within two years (1) or not (0). The positive class is \emph{1}. The COMPAS recid. subset is fairly balanced with an IR $1:1.20$ (positive:negative) while the COMPAS viol. recid. subset is imbalanced with an IR $1:5.17$. 

\begin{table*}[!ht]
\caption{COMPAS recid: attributes characteristics}
\label{tbl:COMPAS_attribute}
\begin{adjustbox}{width=1\linewidth}
    \begin{tabular}{llccl}
        \hline
        \multicolumn{1}{c}{\textbf{Attributes}} & \multicolumn{1}{c}{\textbf{Type}} & 
        \multicolumn{1}{c}{\textbf{Values}} & 
        \multicolumn{1}{c}{\textbf{\#Missing values}} &
        \multicolumn{1}{c}{\textbf{Description}}
        \\ \hline
sex  & Binary &  \{Male, Female\} &  0 & Sex\\
age & Numerical &   [18 - 96] & 0 & Age in years\\
age\_cat  & Categorical &  3  & 0 & Age category\\
race  & Categorical &  6 &  0 & Race\\
juv\_fel\_count & Numerical &   [0 - 20] & 0 & The juvenile felony count\\
juv\_misd\_count & Numerical   & [0 - 13] & 0 & The juvenile misdemeanor count\\
juv\_other\_count & Numerical   & [0 - 17] & 0 & The juvenile other offenses count\\
priors\_count & Numerical &   [0 - 38] & 0 &The prior offenses count\\
c\_charge\_degree  & Binary &  \{F, M\}  & 0 & Charge degree of original crime\\
score\_text  & Categorical &  3  & 0 & ProPublica-defined category of decile score\\
v\_score\_text  & Categorical &  3  & 0 & ProPublica-defined category of v\_decile\_score\\
two\_year\_recid & Binary   & \{0, 1\} & 0 &  Whether the defendant is rearrested within two years \\
        \hline
    \end{tabular}
\end{adjustbox}
\end{table*}

\begin{figure}[!h]
  \centering
         \includegraphics[width=0.6\linewidth]{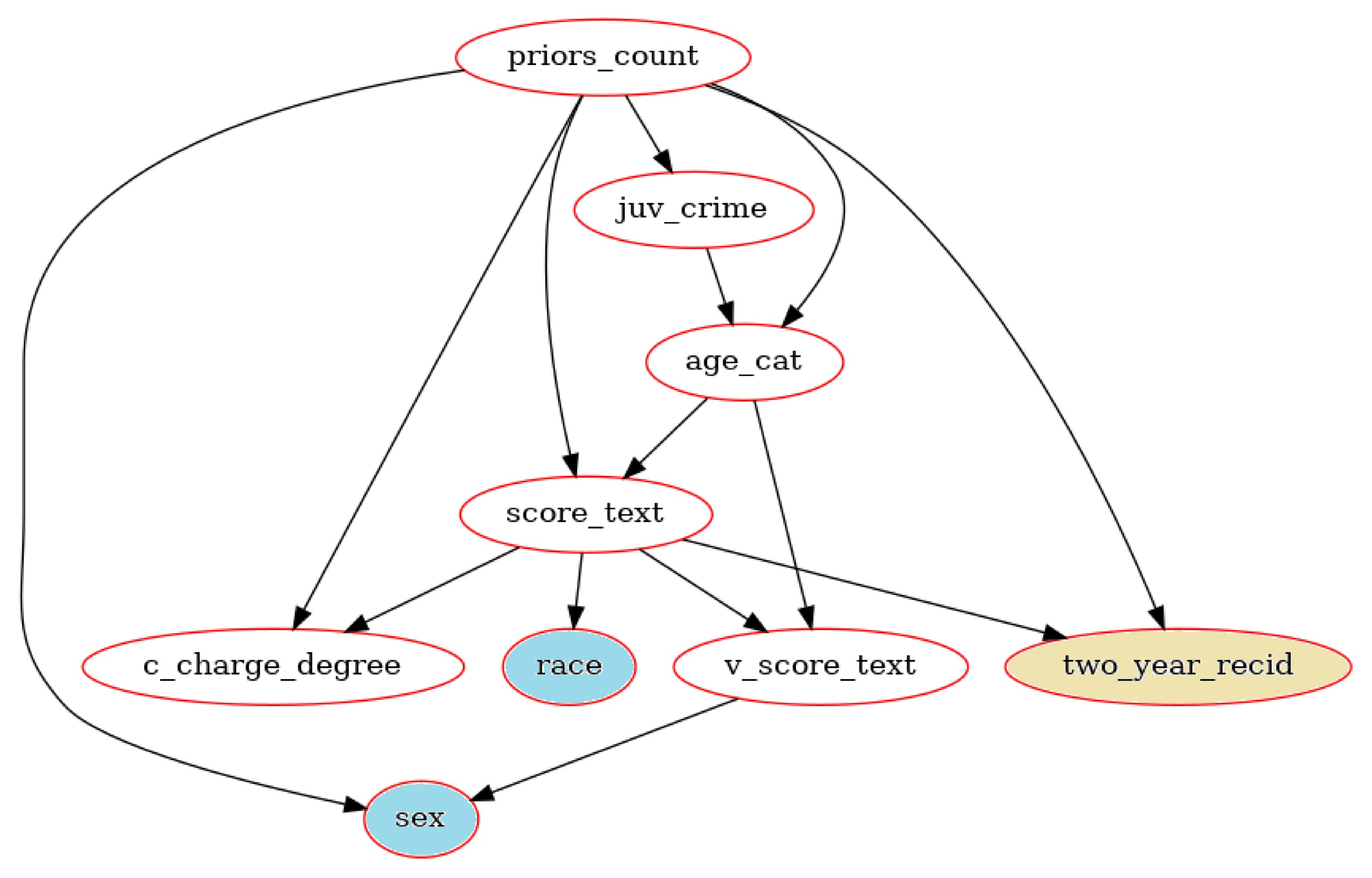}
         \caption{COMPAS recid.: Bayesian network (class label: \textit{two\_year\_recid}, protected attributes: \textit{race}{, \textit{sex}})}
         \label{fig:compas_recid_BN}
\end{figure}

\begin{figure}[!h]
  \centering
         \includegraphics[width=0.6\linewidth]{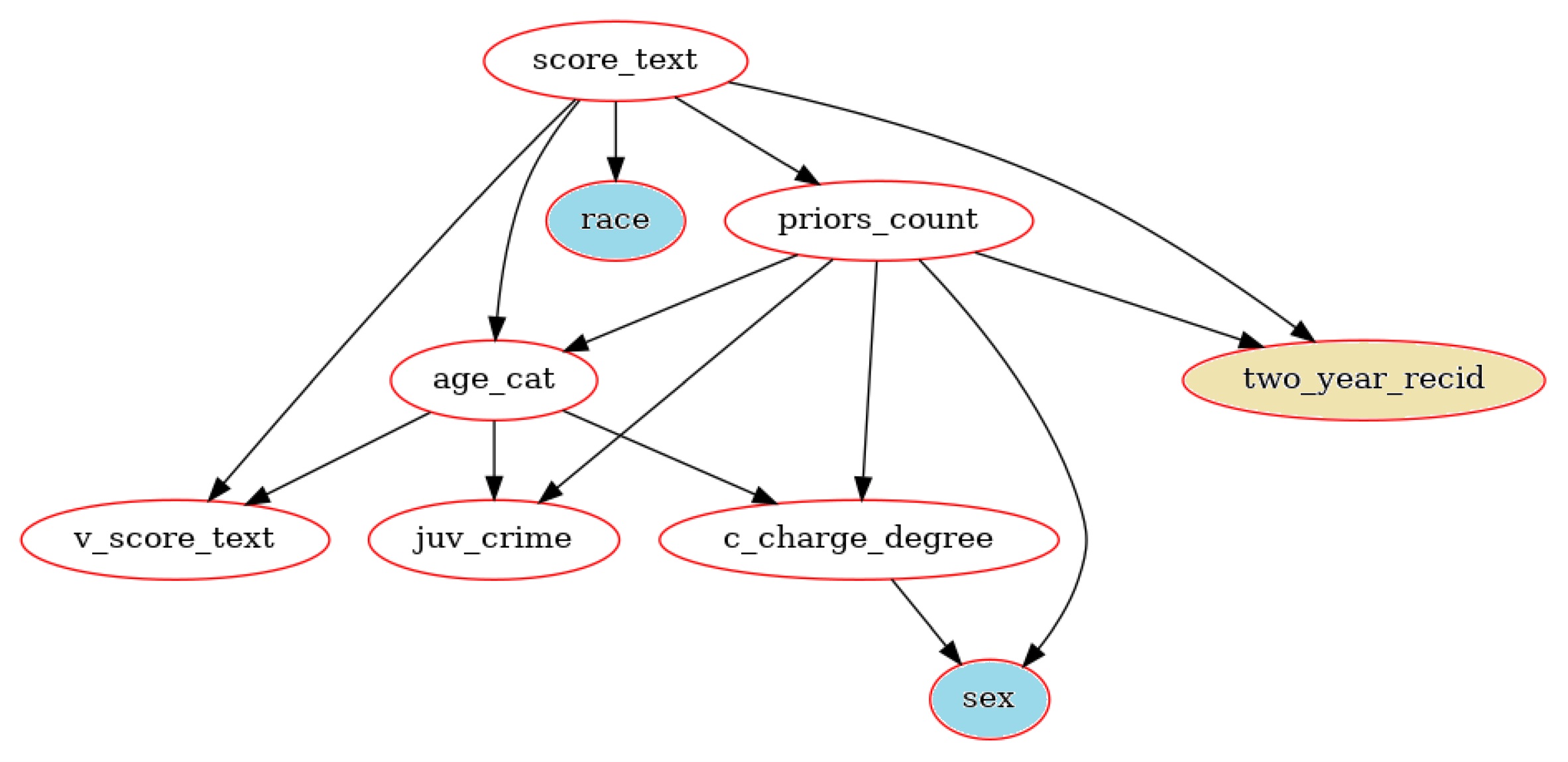}
         \caption{COMPAS viol. recid.: Bayesian network (class label: \textit{two\_year\_recid}, protected attributes: \textit{race}{, \textit{sex}})}
         \label{fig:compas_viol_BN}
\end{figure}

\begin{figure} [htb!]
     \centering
     \begin{subfigure}[b]{0.45\linewidth}
         \centering
         \label{fig:compas_recid_prior_recid}
         \includegraphics[width=\linewidth]{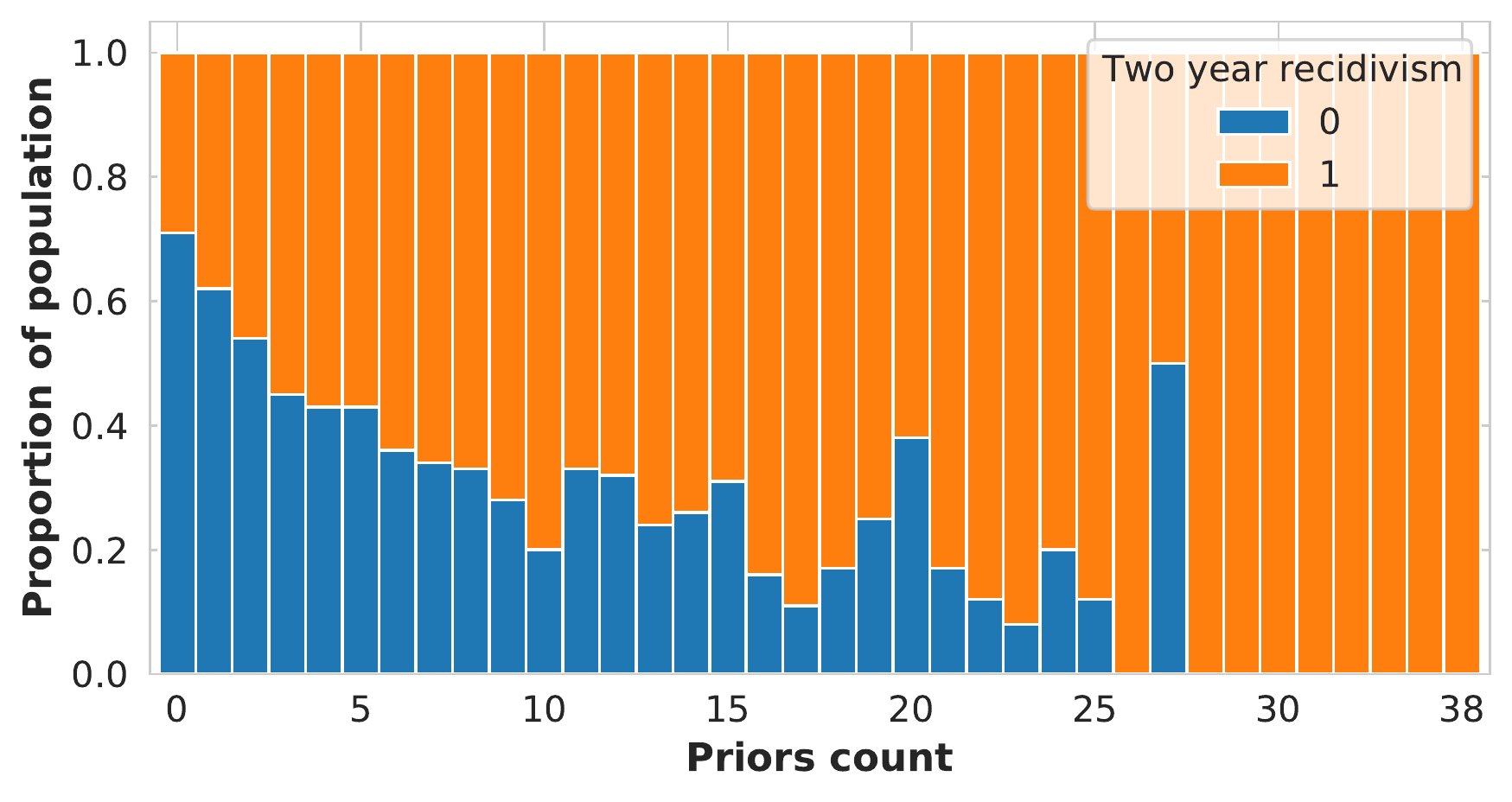}
         \caption{COMPAS recid. subset}
     \end{subfigure}
    \quad
     \begin{subfigure}[b]{0.45\linewidth}
         \centering
         \includegraphics[width=\linewidth]{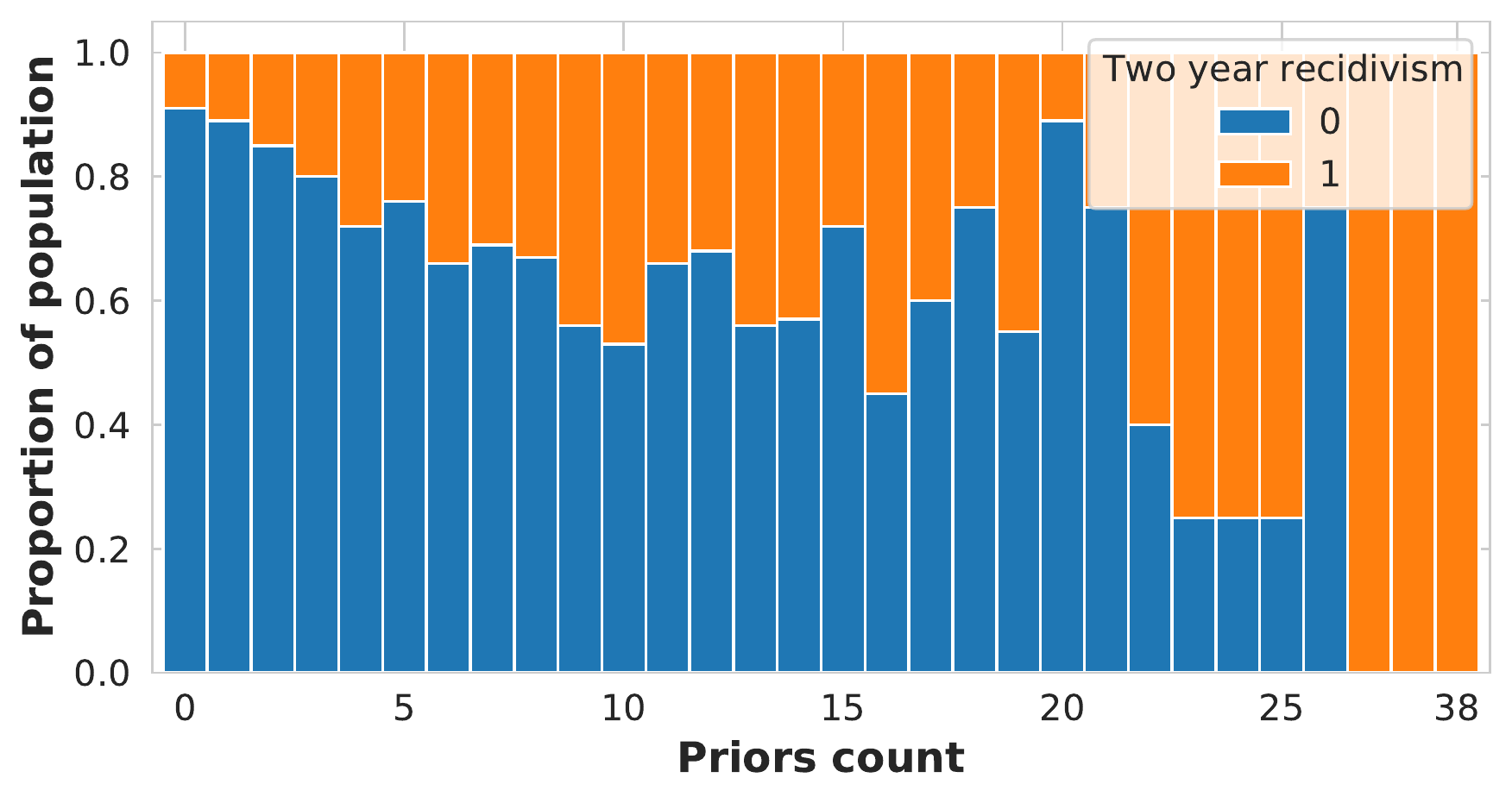}
         \caption{COMPAS viol. recid. subset}
         \label{fig:compas_recid_prior_viol}
         
     \end{subfigure}
     \caption{COMPAS: Relationship between recidivism and priors count}
     \label{fig:compas_recid_prior}
\end{figure}

\noindent\textbf{Bayesian network:} To generate the Bayesian network, we remove the temporal attributes such as \textit{screening\_date} (the date on which the risk of recidivism score was given), \textit{in\_custody} (the date on which individual was brought into custody), and several ID-related attributes. A new attribute \textit{juv\_crime} is computed by the sum of the juvenile felony count (\textit{juv\_fel\_count}) and the juvenile misdemeanor count (\textit{juv\_misd\_count}) and the juvenile other offenses count (\textit{juv\_other\_count}).
We transform the numerical attributes into the categorical type: prior offenses count \textit{priors\_count} = \{0, 1-5, $>$5\}; the juvenile felony count \textit{juv\_crime} = \{0, $>$0\}. 
Figure~\ref{fig:compas_recid_BN} and Figure~\ref{fig:compas_viol_BN} are the Bayesian networks of the COMPAS dataset. The class label \textit{two\_year\_recid} = \{0, 1\} is assigned as a leaf node. It shows the dependency of many attributes such as \textit{sex}, age category (\textit{age\_cat}) on prior offenses count (\textit{priors\_count}) feature. For instance, the number of convictions directly affects the frequency of recidivism, as shown in Figure~\ref{fig:compas_recid_prior}. If a defendant has a long history of convictions, his probability of recidivism is higher, especially when the number of convictions is more than 27 times, the recidivism probability is almost 100\%.

\begin{figure}[htb!]
  \centering
  \includegraphics[width=\linewidth]{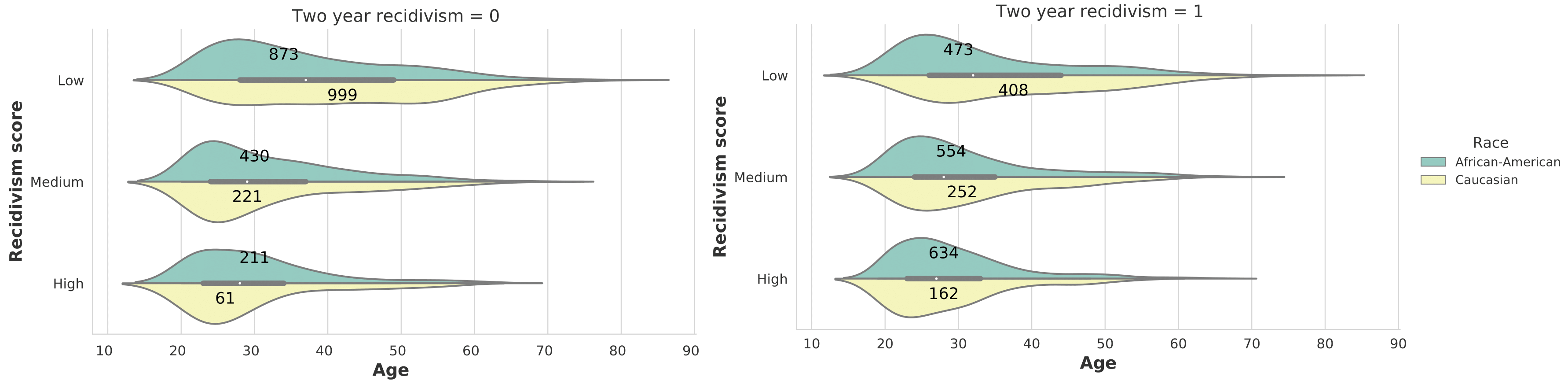}
  \caption{COMPAS recid. : distribution of recidivism score, age w.r.t. race}
  \label{fig:compas_violin}
\end{figure}

Interestingly, \textit{score\_text} attribute (defines the category of the recidivism score) has many ingoing and outgoing edges as depicted in Figure~\ref{fig:compas_viol_BN}. To clarify this phenomenon, we investigate the distribution of age, recidivism score (\textit{score\_text}) w.r.t. \textit{race}, in Figure~\ref{fig:compas_violin}. The majority of recidivists are under the age of 30. In the recidivist group, the number of black criminals is four times and two times more than that of white criminals with a high recidivism score and medium recidivism score, respectively. In the group of defendants with a low recidivism score, the distribution of the \textit{race} is balanced.


\subsubsection{Communities and Crime dataset}
\label{subsubsec:communities-crime}

The Communities and Crime\footnote{http://archive.ics.uci.edu/ml/datasets/communities+and +crime} dataset~\cite{dua2017} was a small dataset containing the socio-economic data from 46 states of the United States in 1990 (the US Census). The law enforcement data come from the 1990 US LEMAS survey, and crime data come from the 1995 FBI {Uniform Crime Reporting (UCR) program}. The goal is to predict the total number of violent crimes per 100 thousand population. Many researchers are investigating the dataset in their experiments (Appendix~\ref{sec:citation}).

\begin{table*}[h!]
\caption{Communities and Crime: attributes characteristics}
\label{tbl:C_C_attribute1}
\begin{adjustbox}{width=1\linewidth}
    \begin{tabular}{llccl}
        \hline
        \multicolumn{1}{c}{\textbf{Attributes}} & \multicolumn{1}{c}{\textbf{Type}} & 
        \multicolumn{1}{c}{\textbf{Values}} & 
        \multicolumn{1}{c}{\textbf{\#Missing values}} &
        \multicolumn{1}{c}{\textbf{Description}}
        \\ \hline
racepctblack & Numerical & [0.0 - 1.0] & 0 & The percentage of population that is African American \\
pctWInvInc & Numerical & [0.0 - 1.0] & 0 & The percentage of households with investment/rent income in 1989\\
pctWPubAsst & Numerical & [0.0 - 1.0] & 0 & The percentage of households with public assistance income in 1989\\
NumUnderPov & Numerical & [0.0 - 1.0] & 0 & The number of people under the poverty level\\
PctPopUnderPov & Numerical & [0.0 - 1.0] & 0 & The percentage of people under the poverty level\\
PctUnemployed & Numerical & [0.0 - 1.0] & 0 & The percentage of people 16 and over, in the labor force, and unemployed\\
MalePctDivorce & Numerical & [0.0 - 1.0] & 0 & The percentage of males who are divorced\\
FemalePctDiv & Numerical & [0.0 - 1.0] & 0 & The percentage of females who are divorced\\
TotalPctDiv & Numerical & [0.0 - 1.0] & 0 & The percentage of population who are divorced\\
PersPerFam & Numerical & [0.0 - 1.0] & 0 & The mean number of people per family \\
PctKids2Par & Numerical & [0.0 - 1.0] & 0 & The percentage of kids in family housing with two parents\\
PctYoungKids2Par & Numerical & [0.0 - 1.0] & 0 & The percentage of kids 4 and under in two parent households\\        
PctTeen2Par & Numerical & [0.0 - 1.0] & 0 & The percentage of kids age 12-17 in two parent households \\
NumIlleg & Numerical & [0.0 - 1.0] & 0 & The number of kids born to never married\\
PctIlleg & Numerical & [0.0 - 1.0] & 0 & The percentage of kids born to never married\\
PctPersOwnOccup & Numerical & [0.0 - 1.0] & 0 & The percentage of people in owner occupied households\\
HousVacant & Numerical & [0.0 - 1.0] & 0 & The number of vacant households\\
PctHousOwnOcc & Numerical & [0.0 - 1.0] & 0 & The percentage of households owner occupied\\
PctVacantBoarded & Numerical & [0.0 - 1.0] & 0 & The percentage of vacant housing that is boarded up\\
NumInShelters & Numerical & [0.0 - 1.0] & 0 & The number of people in homeless shelters\\
NumStreet & Numerical & [0.0 - 1.0] & 0 & The number of homeless people counted in the street\\
ViolentCrimesPerPop & Numerical & [0.0 - 1.0] & 0 & The total number of violent crimes per 100,000 popuation\\
        \hline
    \end{tabular}
\end{adjustbox}
\end{table*}

\begin{figure*}[h!]
  \centering
  \includegraphics[width=0.65\linewidth]{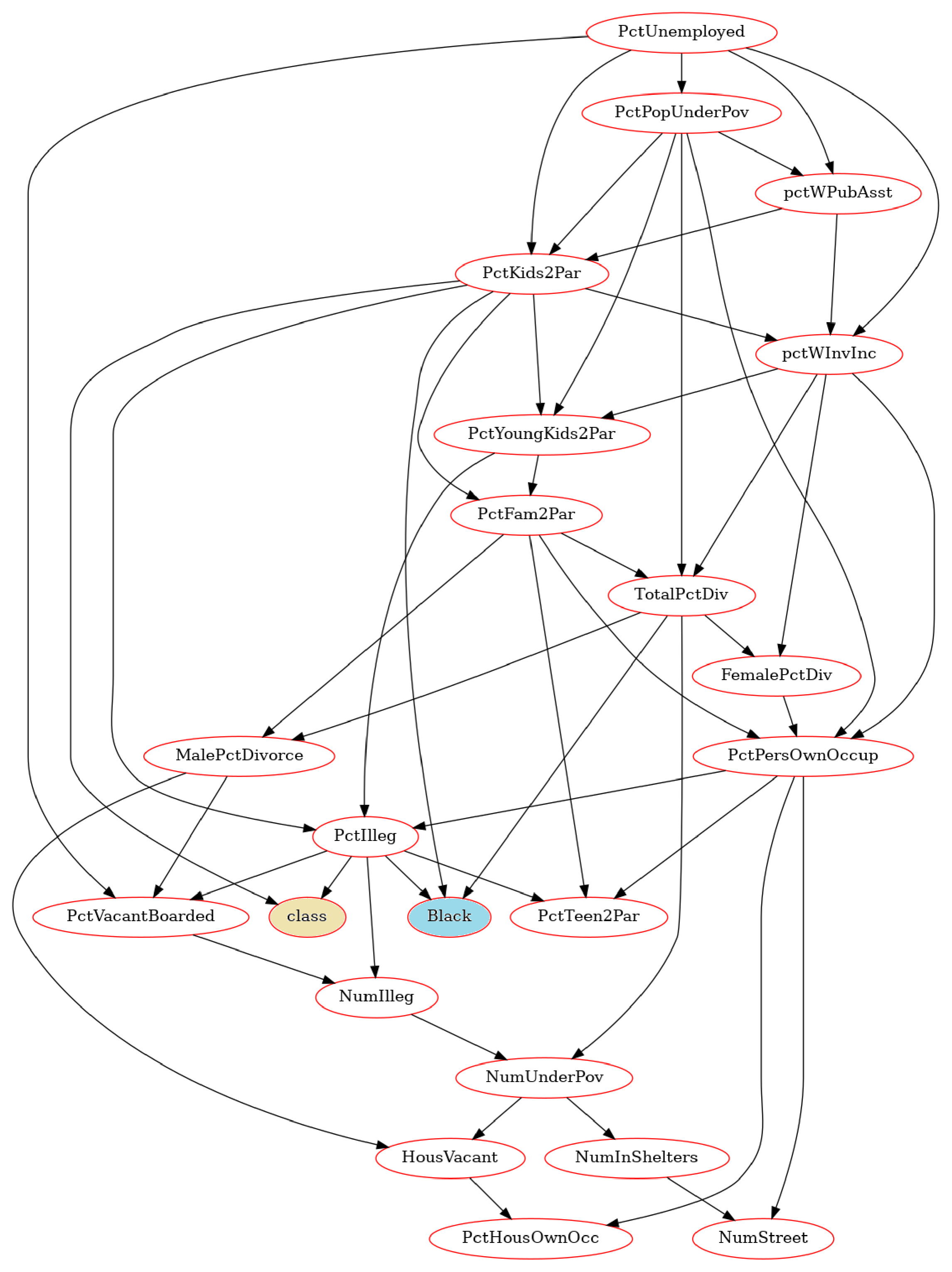}
  \caption{Communities and Crime: Bayesian network (class label: \textit{class}, protected attribute: \textit{black})}
  \label{fig:communities_crime_NB}
\end{figure*}

\noindent\textbf{Dataset characteristics:} The dataset contains only 1,994 samples; each instance is described by 127 attributes (4 categorical and 123 numerical attributes). 
A description of attributes\footnote{Table~\ref{tbl:C_C_attribute1} contains attributes used in the Bayesian network} is illustrated in Table~\ref{tbl:C_C_attribute1}, Table~\ref{tbl:C_C_attribute_extra1} and Table~\ref{tbl:C_C_attribute_extra2} (Appendix \ref{sec:dataset_characteristics}).

There is a very high proportion (84\%) of missing values in 25 attributes, as demonstrated in Table~\ref{tbl:C_C_attribute_extra2}. Based on the suggestions from the literature  \cite{heidari2018fairness,calders2013controlling} , we remove all columns containing missing values. We create a new binary class label namely \textit{class} based on \textit{ViolentCrimesPerPop} attribute (the total number of violent crimes per 100,000 population). As illustrated in the related work {\cite{kearns2018preventing}}, a label ``high-crime'' is set if the crime rate of the communities {(possitive class)} is greater than 0.7, otherwise, ``low-crime'' is given. The ratio of \textit{high-crime:low-crime} is: 122:1,872 (6.1\%:93.9\%).  Therefore,  the dataset is very imbalanced with an IR $1:15.34$.

\noindent\textbf{Protected attributes:}
In the literature {\cite{kamishima2012fairness,kamiran2013quantifying}}, typically, researchers derive a new attribute, namely \textit{Black}, which is {considered as} the protected attribute, in order to divide the communities according to race by thresholding the attribute \textit{racepctblack} (the percentage of the population that is African American) at 0.06. The ratio of \textit{black:non-black} is 1,038:956 (52.1\%:47.9\%). The interesting point in the data is that 94.3\% (115/122) of the class ``high-crime'' are communities dominated by \emph{blacks}.

\noindent\textbf{Bayesian network:} The dataset contains 122 numerical attributes normalized in the range of $(0,1)$, which is not competent to the Bayesian network. Hence, we use the median value 0.5 as a threshold to transform these attributes into categorical with two values $\{\leq0.5,>0.5\}$. Besides, to ensure the visibility of the chart and the computation time, we use 21 attributes that have a high correlation {(at a threshold of 0.25)} with the class label. The Bayesian network is visualized in Figure~\ref{fig:communities_crime_NB}.
In which, the percentage of \textit{kids born to never married} (\textit{PctIlleg}) and the percentage of \textit{kids in family housing with two parents} (\textit{PctKids2Par}) have a direct impact on the class label and the race. Looking into the dataset, we discover that 92.4\% of the communities are dominated by \emph{black} people, where the percentage of \textit{kids in family housing with two parents} less than 50\%), while only 55.6\% of the communities are dominated by \emph{black} people, where the percentage of \textit{kids in family housing with two parents} greater than 50\%.

\textbf{Summary of the criminological datasets:}
In summary, the criminological datasets were only surveyed in the US. 

\emph{Race} and \emph{sex} are considered as protected attributes, with 
\emph{race} being the most prevalent protected attribute. Historical bias w.r.t \emph{race} has been detected in the data, but comprises a challenge for ML models.
Furthermore, the datasets consists of many attributes (the richer description among all datasets, see Table~\ref{tbl:real_world_datasets}); hence, a careful selection of attributes for fairness-aware learning is required.

%% file: realdata_society.tex
\subsection{Healthcare and social datasets}
\label{subsec:social}
\subsubsection{Diabetes dataset}
\label{subsubsec:diabetes}
The diabetes\footnote{https://archive.ics.uci.edu/ml/datasets/diabetes+130-us+hospitals+for+years+1999-2008} dataset~\cite{strack2014impact} describes the clinical care at 130 US hospitals and integrated delivery networks from 1999 to 2008. The classification task is to predict whether a patient will readmit within 30 days. The dataset is investigated in several studies (Appendix~\ref{sec:citation}).

\noindent\textbf{Dataset characteristics:} The dataset contains 101,766 patients described by 50 attributes (10 numerical, 7 binary and 33 categorical). Characteristics of all attributes\footnote{Table~\ref{tbl:diabetes_attributes} describes attributes used in the Bayesian network} are summarized in Table~\ref{tbl:diabetes_attributes} and Table~\ref{tbl:diabetes_attributes_extra} (in Appendix~\ref{sec:dataset_characteristics}). The attributes \textit{encounter\_id} and \textit{patient\_nbr} should not be considered in the learning tasks since they are the ID of the patients. Typically, \textit{weight, payer\_code, medical\_specialty} attributes are removed because they contains at least 40\% of missing values. Furthermore, we eliminate the missing values in \textit{race, diag\_1, diag\_2, diag\_3} columns. The class label \textit{readmitted} contains 54,864 rows with ``no record of readmission'', hence, these rows should be clean. The clean version of the dataset contains 45,715 records.
\begin{table*}[!ht]
\caption{Diabetes: attributes characteristics}
\label{tbl:diabetes_attributes}
\begin{adjustbox}{width=1\linewidth}
    \begin{tabular}{llccl}
        \hline
        \multicolumn{1}{c}{\textbf{Attributes}} & \multicolumn{1}{c}{\textbf{Type}} & 
        \multicolumn{1}{c}{\textbf{Values}} & 
        \multicolumn{1}{c}{\textbf{\#Missing values}} &
        \multicolumn{1}{c}{\textbf{Description}} 
        \\ \hline
race  & Categorical &  6 & 2,273 & Race (Caucasian, Asian, African American, Hispanic, and other)\\
gender  & Categorical &  3 & 0 & Gender (male, female, and unknown/invalid)\\
age  & Categorical &  10 & 0 & Grouped in 10-year intervals\\
time\_in\_hospital & Numerical & [1 - 14] & 0 & The number of days between admission and discharge\\
num\_procedures & Numerical & [0 - 6] & 0 & The number of procedures (other than lab tests) performed during the encounter\\
num\_medications & Numerical & [1 - 81] & 0 & The number of distinct generic names administered during the encounter \\
number\_outpatient & Numerical & [0 - 42] & 0 & The number of outpatient visits of the patient in the year preceding the encounter\\
number\_emergency & Numerical & [0 - 76] & 0 & The number of emergency visits of the patient in the year preceding the encounter\\
number\_inpatient & Numerical & [0 - 21] & 0 & The number of inpatient visits of the patient in the year preceding the encounter \\
A1Cresult  & Categorical &  4 & 0 & The range of the results or if the test was not taken\\
metformin  & Categorical &  4 & 0 & Whether the drug was prescribed or there was a change in the dosage\\
chlorpropamide  & Categorical &  4 & 0&Whether the drug was prescribed or there was a change in the dosage \\
glipizide  & Categorical &  4 & 0&Whether the drug was prescribed or there was a change in the dosage \\
rosiglitazone  & Categorical &  4 & 0&Whether the drug was prescribed or there was a change in the dosage \\
acarbose  & Categorical &  4 & 0&Whether the drug was prescribed or there was a change in the dosage \\
miglitol  & Categorical &  4 & 0& Whether the drug was prescribed or there was a change in the dosage\\
diabetesMed  & Binary & \{Yes, No\}& 0 & Was there any diabetic medication prescribed?\\
readmitted  & Categorical &  3 & 0 & The number of days to inpatient readmission (No, $<30$, $>30$)\\
        \hline
    \end{tabular}
\end{adjustbox}
\end{table*}

\noindent\textbf{Protected attributes:}
Typically \textit{Gender} = \{\textit{male, female}\} is chosen as the protected attribute. The ratio of \textit{male:female} is 20,653:25,062 (45.2\%:54.8\%). The ratio of males or females who have to readmit hospital in less than 30 days is approximately 24\%.

\textbf{Class attribute:} The class attribute is \emph{readmitted} $\in \{<30, >30\}$ indicating whether a patient will readmit within 30 days. The positive class is ``$<30$''. The dataset is imbalanced with an IR $1:3.13$ (positive:negative).

\begin{figure*}[h!]
  \centering
  \includegraphics[width=1\linewidth]{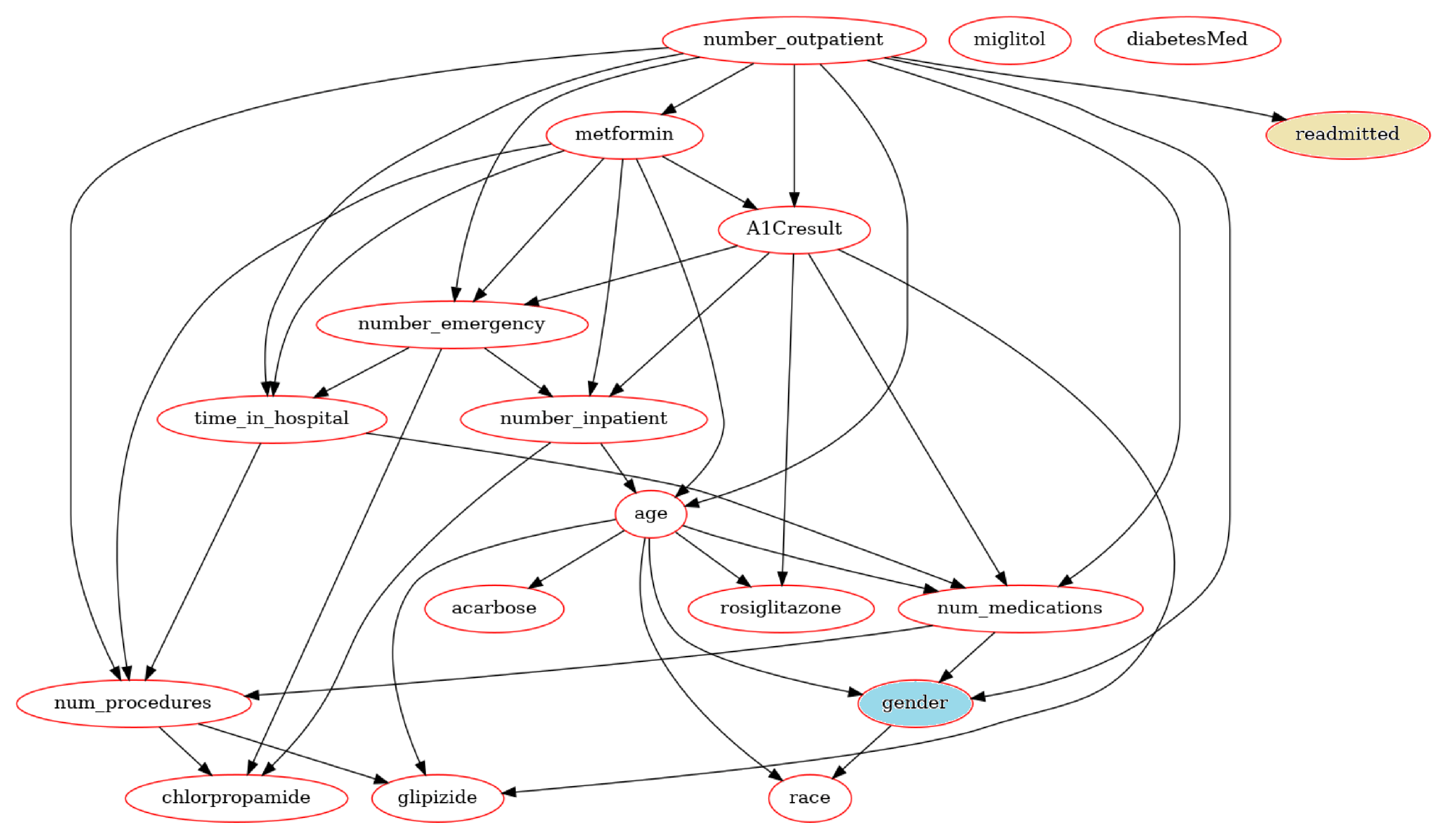}
  \caption{Diabetes: Bayesian network (class label: \textit{readmitted}, protected attribute: \textit{gender})}
  \label{fig:diabetes_NB}
\end{figure*}

\noindent\textbf{Bayesian network:}
To prepare the dataset for Bayesian network generating process, we encode the attributes: \textit{age} = \{$<$40, 40-59, 60-79, 80-99\}; \textit{time\_in\_hospital} = \{$\leq$5, $>$5\}; \textit{num\_lab\_procedures} = \{$\leq$50, 50\}; \textit{num\_procedures} = \{$\leq$1, $>$1\};  \textit{number\_outpatient} = \{0, $>$0\}; \textit{num\_medications} = \{$\leq$15, $>$15\}; \textit{number\_emergency} = \{0, $>$0\}; \textit{number\_inpatient} = \{0, $>$0\}; \textit{number\_diagnoses} = \{0, $>$0\}. 
To reduce the computation time, we use 17 attributes that have an absolute correlation coefficient higher than 0.005 with \emph{gender} and \emph{readmitted} attributes to generate the Bayesian network in Figure~\ref{fig:diabetes_NB}.

The class label \textit{readmitted} is directly conditionally dependent on the number of outpatient visits of the patient in the year preceding the encounter (\textit{number\_outpatient}). The attribute \textit{number\_outpatient} also has an impact on 8 other features. Interestingly, there is no connection between the protected attribute \textit{gender} and the class label. 


\subsubsection{Ricci dataset}
\label{subsubsec:ricci}
The Ricci\footnote{https://www.key2stats.com/data-set/view/690} dataset was generated by the Ricci v.DeStefano case~\cite{ricci2009}, in which they investigated the results of a promotion exam within a fire department in Nov 2003 and Dec 2003. Although it is a relatively small dataset, it has been employed for fairness-aware classification tasks in many studies (Appendix~\ref{sec:citation}).
The classification task is to predict whether an individual obtains a promotion based on the exam results.

\noindent\textbf{Dataset characteristics:}
The dataset consists of 118 samples, where each sample is characterized by 6 attributes (3 numerical and 3 binary attributes), as presented in Table~\ref{tbl:ricci_attributes}.

\begin{table*}[!ht]
\caption{Ricci: attributes characteristics}
\label{tbl:ricci_attributes}
\centering
\begin{adjustbox}{width=0.9\linewidth}
    \begin{tabular}{llccl}
        \hline
        \multicolumn{1}{c}{\textbf{Attributes}} & \multicolumn{1}{c}{\textbf{Type}} & 
        \multicolumn{1}{c}{\textbf{Values}} & 
        \multicolumn{1}{c}{\textbf{\#Missing values}}&
        \multicolumn{1}{c}{\textbf{Description}}
        \\ \hline
Position  & Binary &  \{Lieutenant, Captain\}  & 0 & The desired promotion \\
Oral & Numerical &   [40.83 - 92.08] & 0 & The oral exam score\\
Written & Numerical   & [46 - 95] & 0 & The written exam score\\
Race  & Binary &  \{White, Non-White\}  & 0 & Race\\
Combine & Numerical &   [45.93 - 92.80] & 0 & The combined score (the written exam gets 60\% weight)\\
Promoted & Binary &  \{True, False\} & 0 & Whether an individual obtains a promotion or not \\
        \hline
    \end{tabular}
\end{adjustbox}
\end{table*}

\noindent\textbf{Protected attributes:}
In this dataset, only attribute \textit{race} can be used as a protected attribute. \textit{Race} contains three values (\textit{black, white}, and \textit{hispanic}). As described in the literature, ``black'' and ``hispanic'' are grouped as ``non-white'' community. The ratio of \textit{white:non-white} is 68:50 (57.6\%:42.4\%).

\textbf{Class attribute:} The class attribute is \emph{promoted} $\in \{True, False\}$ revealing whether an individual achieves a promotion or not. The positive class is ``True''. The dataset is almost balanced with an IR $1:1.11$ (positive:negative).

\begin{figure}[!htb]
  \centering
  \includegraphics[width=0.2\linewidth]{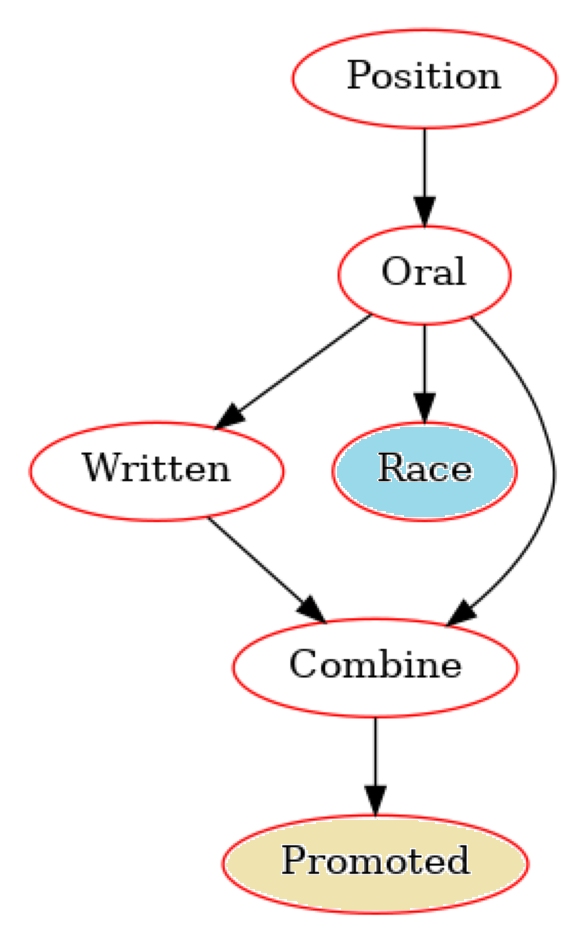}
  \caption{Ricci: Bayesian network (class label: \textit{promoted}, protected attribute: \textit{race})}
  \label{fig:ricci_NB}
\end{figure}

\noindent\textbf{Bayesian network:}
We encode 3 numerical attributes \textit{oral, written} and \textit{combine} as following: \textit{oral} = \{$<$70, $\geq$70\}, \textit{written} = \{$<$70, $\geq$70\}, \textit{combine} = \{$<$70, $\geq$70\}. The Bayesian network of the Ricci dataset is demonstrated in Figure~\ref{fig:ricci_NB}.

\begin{figure}[h!]
  \centering
  \includegraphics[width=0.45\linewidth]{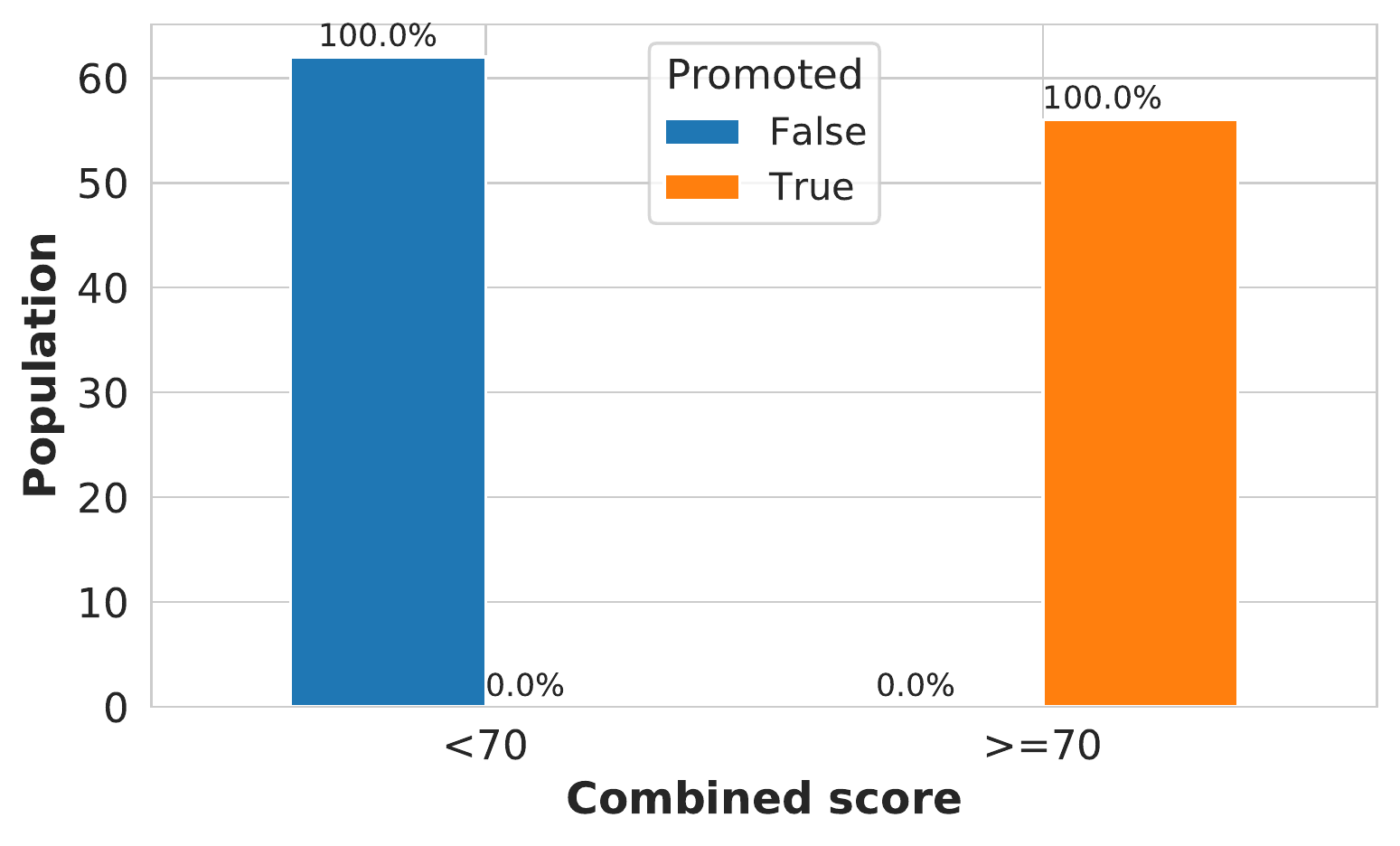}
  \caption{Ricci: Relationship between \textit{combined score} and \textit{promotion status}}
  \label{fig:ricci_combine_promoted}
\end{figure}
\begin{figure}[h!]
  \centering
  \includegraphics[width=1\linewidth]{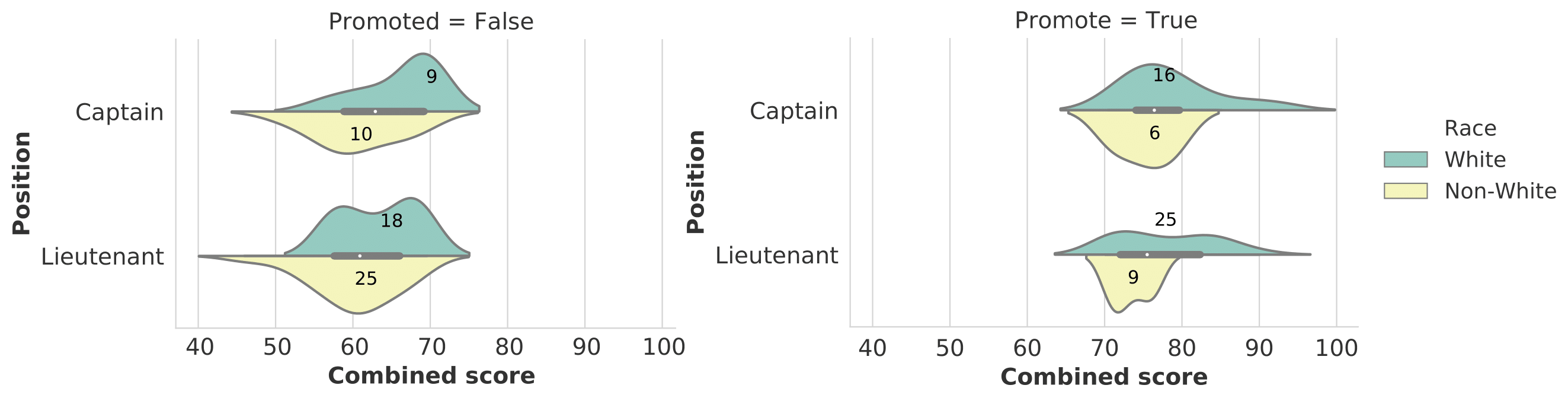}
  \caption{Ricci: Distribution of \textit{combined score}, \textit{position} and \textit{promotion decision} w.r.t. \textit{race}}
  \label{fig:ricci_violin}
\end{figure}

It is easy to observe that the combined grade (attribute \textit{combine}) has a direct effect on the class label (\textit{promoted}). Figure~\ref{fig:ricci_combine_promoted} illustrates the relationship between the combined grade and the promotion status. 100\% of people whose combined oral and written exams are equal to or above 70 are promoted. Besides, as depicted in Figure~\ref{fig:ricci_violin}, the number of promotions are granted for \emph{white} people is higher than that for \emph{non-white} people. The opposite trend is true in the group of candidates with no promotion.

\textbf{Summary of the healthcare and social datasets:} 
In summary, the datasets in healthcare and society domains were only surveyed in the US. \emph{Race} and \emph{gender} are considered as protected attributes. In terms of class imbalance, these datasets are less imbalanced than datasets in other domains, although the Diabetes dataset is still imbalanced. Interestingly, there is no connection between the protected attribute and the class label in both two datasets, which implies fairness can be observed in the results of fairness-aware ML models.

%% file: realdata_education.tex
\subsection{Educational datasets}
\label{subsec:educational}

\subsubsection{Student performance dataset}
\label{subsubsec:student}
The student performance dataset~\cite{cortez2008using} described students' achievement in the secondary education of two Portuguese schools in 2005 - 2006 with two distinct subjects: Mathematics and Portuguese.\footnote{https://archive.ics.uci.edu/ml/datasets/student+performance}. The regression task is to predict the final year grade of the students. It is investigated in several researches
(Appendix~\ref{sec:citation}) with {fairness-aware} regression and clustering approaches. 

\noindent\textbf{Dataset characteristics:}
The dataset contains information of 395 (Mathematics subject) and 649 (Portuguese subject) students described by 33 attributes (4 categorical, 13 binary and 16 numerical attributes). Characteristics of all attributes is described in Table~\ref{tbl:student_attributes}. To simply the classification problem, we create a class label based on attribute \textit{G3}, \textit{class} = \{\textit{Low, High}\}, {corresponding} to \textit{G3} = \{$<$10, $\geq$10\}.  The positive class is ``High''. The dataset is imbalanced with imbalance ratios 1:2.04 (Mathematics subject) and 1:5.09 (Portuguese subject).

\begin{table*}[!ht]
\caption{Student performance: attributes characteristics}
\label{tbl:student_attributes}
\begin{adjustbox}{width=1\linewidth}
    \begin{tabular}{llccl}
        \hline
        \multicolumn{1}{c}{\textbf{Attributes}} & \multicolumn{1}{c}{\textbf{Type}} & 
        \multicolumn{1}{c}{\textbf{Values}} &
        \multicolumn{1}{c}{\textbf{\#Missing values}}&
        \multicolumn{1}{c}{\textbf{Description}}
        \\ \hline
school  & Binary & \{GP, MS\}& 0 & The student's school {(`GP': Gabriel Pereira, `MS': Mousinho da Silveira)}\\
sex  & Binary &  \{Male, Female\} & 0 & Sex\\
age & Numerical & [15 - 22] & 0 & Age (in years)\\
address  & Binary & \{U, R\}& 0 & The address type {(`U': urban, `R':rural)}\\
famsize  & Binary & \{LE3, GT3\}& 0 & The family size {(`LE3': less or equal to 3, `GT3': greater than 3)}\\
Pstatus  & Binary & \{T, A\}& 0 & The parent's cohabitation status {( `T': living together, `A': apart)}\\
Medu & Numerical & [0 - 4] & 0 & Mother's education \\
Fedu & Numerical & [0 - 4] & 0 & Father's education\\
Mjob  & Categorical &  5 & 0 & Mother's job\\
Fjob  & Categorical &  5 & 0 & Father's job\\
reason  & Categorical &  4 & 0 & The reason to choose this school\\
guardian  & Categorical &  3 & 0 & The student's guardian {(mother, father, other)}\\
traveltime & Numerical & [1 - 4] & 0 & The travel time from home to school\\
studytime & Numerical & [1 - 4] & 0 & The weekly study time\\
failures & Numerical & [0 - 3] & 0 & The number of past class failures\\
schoolsup  & Binary & \{Yes, No\} & 0 & Is there an extra educational support? \\
famsup  & Binary & \{Yes, No\} & 0 & Is there any family educational support?\\
paid  & Binary & \{Yes, No\} & 0 & Is there an extra paid classes within the course subject (Math or Portuguese)\\
activities  & Binary & \{Yes, No\} & 0 &Are there extra-curricular activities?\\
nursery  & Binary & \{Yes, No\} & 0 & Did the student attend a nursery school? \\
higher  & Binary & \{Yes, No\} & 0 & Does the student want to take a higher education? \\
internet  & Binary & \{Yes, No\} & 0 & Does the student have an Internet access at home? \\
romantic  & Binary & \{Yes, No\} & 0 & Does the student have a romantic relationship with anyone?\\
famrel & Numerical & [1 - 5] & 0 & The quality of family relationships {(from 1: very bad to 5: excellent)}\\
freetime & Numerical & [1 - 5] & 0 & Free time after school {(from 1: very low to 5: very high)}\\
goout & Numerical & [1 - 5] & 0 & How often does the student go out with friends? {(from 1: very low to 5: very high)}\\
Dalc & Numerical & [1 - 5] & 0 & The workday alcohol consumption {(from 1: very low to 5: very high)}\\
Walc & Numerical & [1 - 5] & 0 & The weekend alcohol consumption {(from 1: very low to 5: very high)}\\
health & Numerical & [1 - 5] & 0 & The current health status {(from 1: very bad to 5:very good)}\\
absences & Numerical & [0 - 32] & 0 & The number of school absences\\
G1 & Numerical & [0 - 19] & 0 & The first period grade\\
G2 & Numerical & [0 - 19] & 0 & The second period grade\\
G3 & Numerical & [0 - 19] & 0 & The final grade\\
        \hline
    \end{tabular}
\end{adjustbox}
\end{table*}

\begin{figure}[h!]
  \centering
  \includegraphics[width=0.5\linewidth]{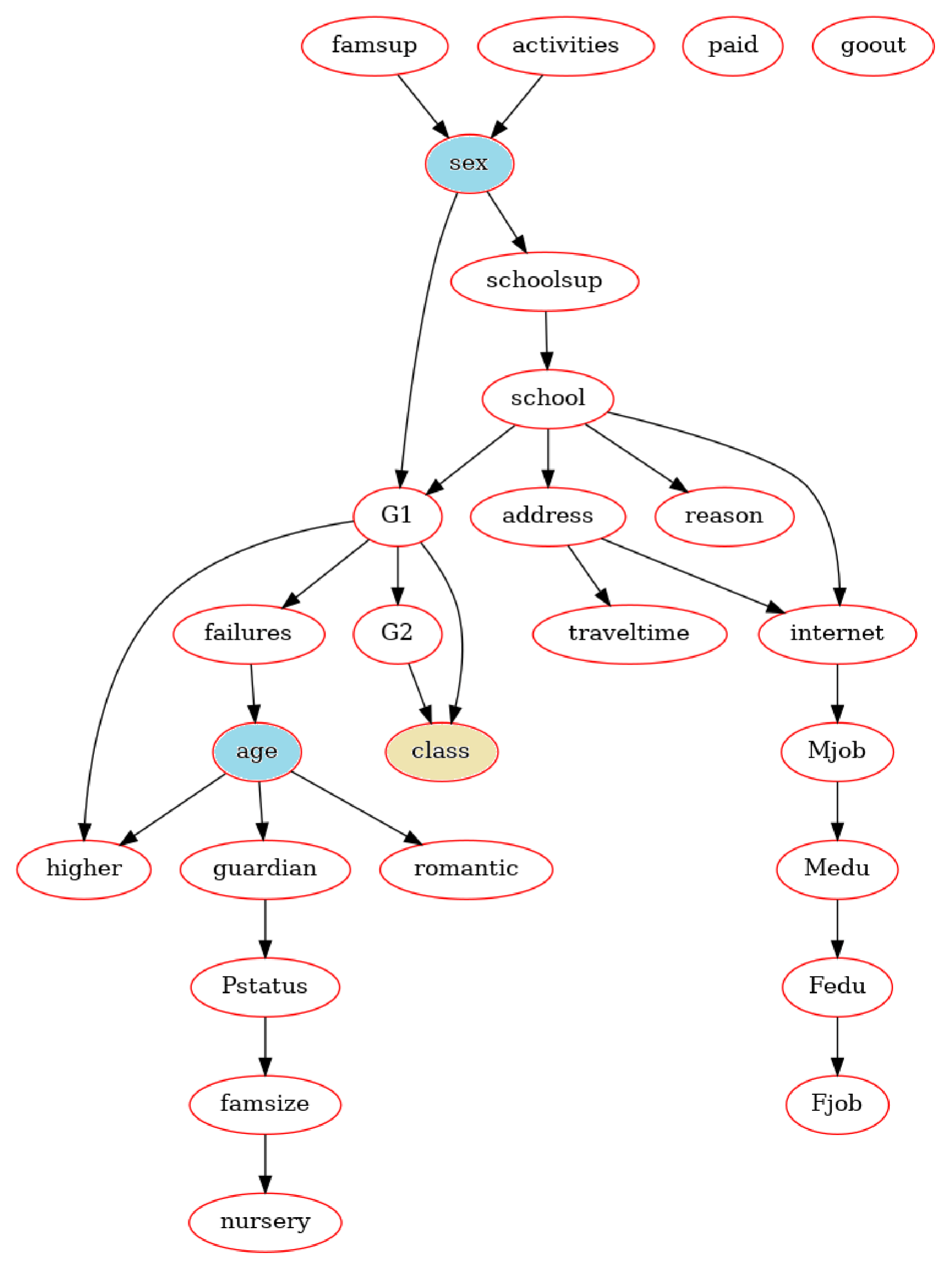}
  \caption{Student performance - Portuguese subject: Bayesian network (class label: \textit{class}, protected attributes: \textit{age, sex})}
  \label{fig:student_por_BN}
\end{figure}

\begin{figure*}[h!]
  \centering
  \includegraphics[width=0.9\linewidth]{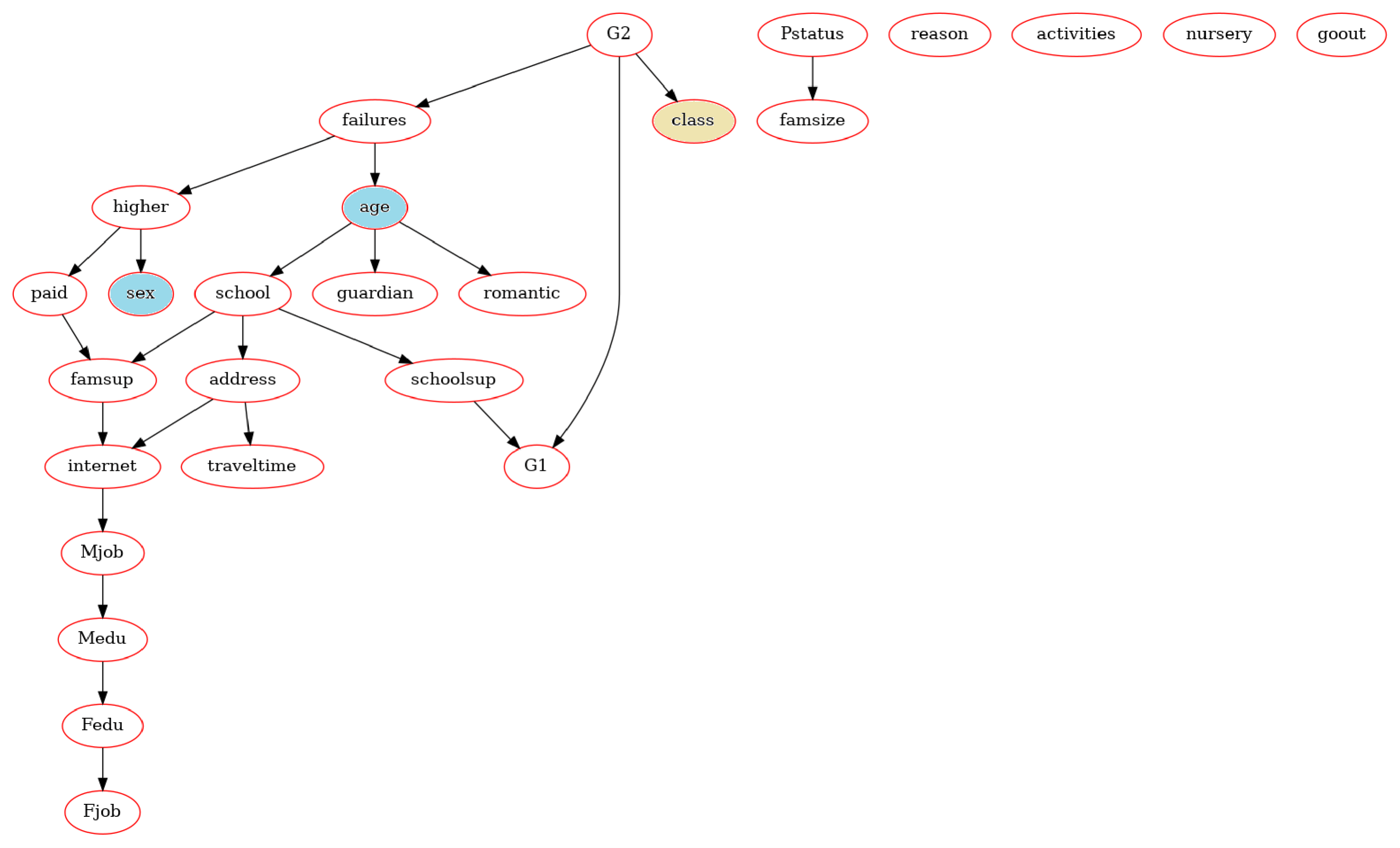}
  \caption{Student performance - Mathematics subject: Bayesian network (class label: \textit{class}, protected attributes: \textit{age, sex})}
  \label{fig:student_mat_BN}
\end{figure*}

\noindent\textbf{Protected attributes:}
Typically, in the literature, \textit{sex} is considered as the protected attribute. In the work of ~\cite{kearns2019empirical,deepak2020fair}, they also select \textit{age} as the protected attribute. Especially, in the research \cite{kearns2019empirical}, they consider atttributes \emph{romatic} (relationship) and \emph{dalc, walc} (alcohol consumption) as the protected attributes. However, because of the unpopularity of these attributes, we did not consider those within the scope of this paper.
\begin{itemize}
    \item \emph{sex} = \{\textit{male, female}\}: the dataset is dominated by female students. The ratios of \textit{male:female} are 208:187 (52.7\%:47.3\%) and 383:266 (59\%:41\%) for the Mathematics subject and Portuguese subject, respectively.
    \item \emph{age} = \{\textit{$<$18, $\geq$ 18}\}: young students (less than 18 years old) are the majority with the ratios of ``$<18$'':``$\geq 18$'' are 284:111 (71.9\%: 28.1\%) and 468:181 (72.1\%:27.9\%) for the Mathematics subject and Portuguese subject, respectively.
\end{itemize}
\noindent\textbf{Bayesian network:}
We perform a transformation of numerical variables: the number of school absences, \textit{absences} = \{0-5, 6-20, $>$20\}; grade \textit{G1} = \{$<$10, $\geq$10\}; \textit{G2} = \{$<$10, $\geq$10\}. Due to the computation of the Bayesian network generator and the correlation coefficient with the class label (with a threshold of 0.02), we select 26 variables for the network. The Bayesian networks of the dataset on Portuguese and Mathematics subjects are visualized in Figure~\ref{fig:student_por_BN} and Figure~\ref{fig:student_mat_BN}, respectively.


\begin{figure}[h!]
    \centering
    \begin{subfigure}[t]{0.35\textwidth}
        \centering
        \includegraphics[width=1\linewidth]{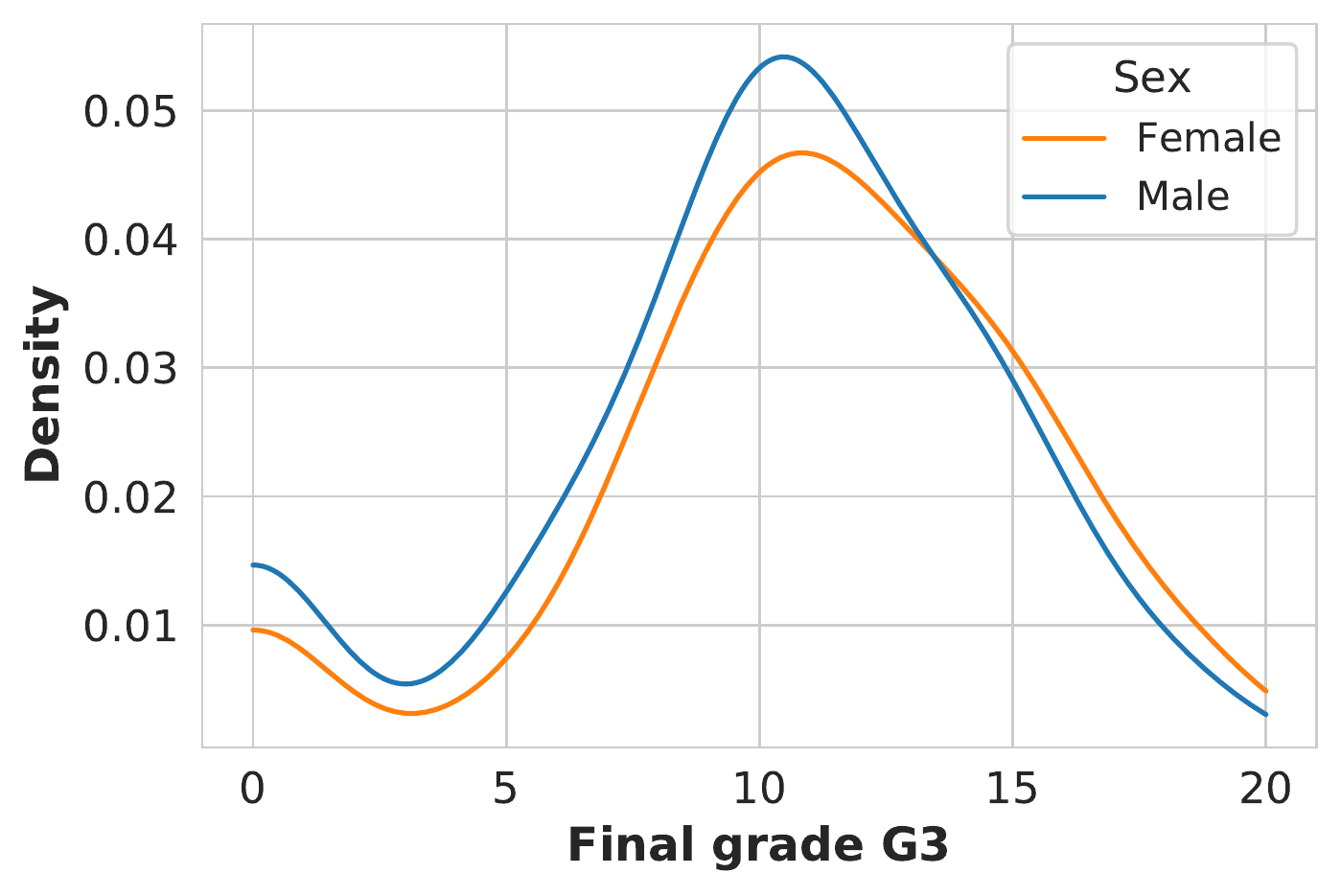}
        \caption{Mathematics subject}
    \end{subfigure}
    \quad
    \begin{subfigure}[t]{0.35\textwidth}
        \centering
        \includegraphics[width=1\linewidth]{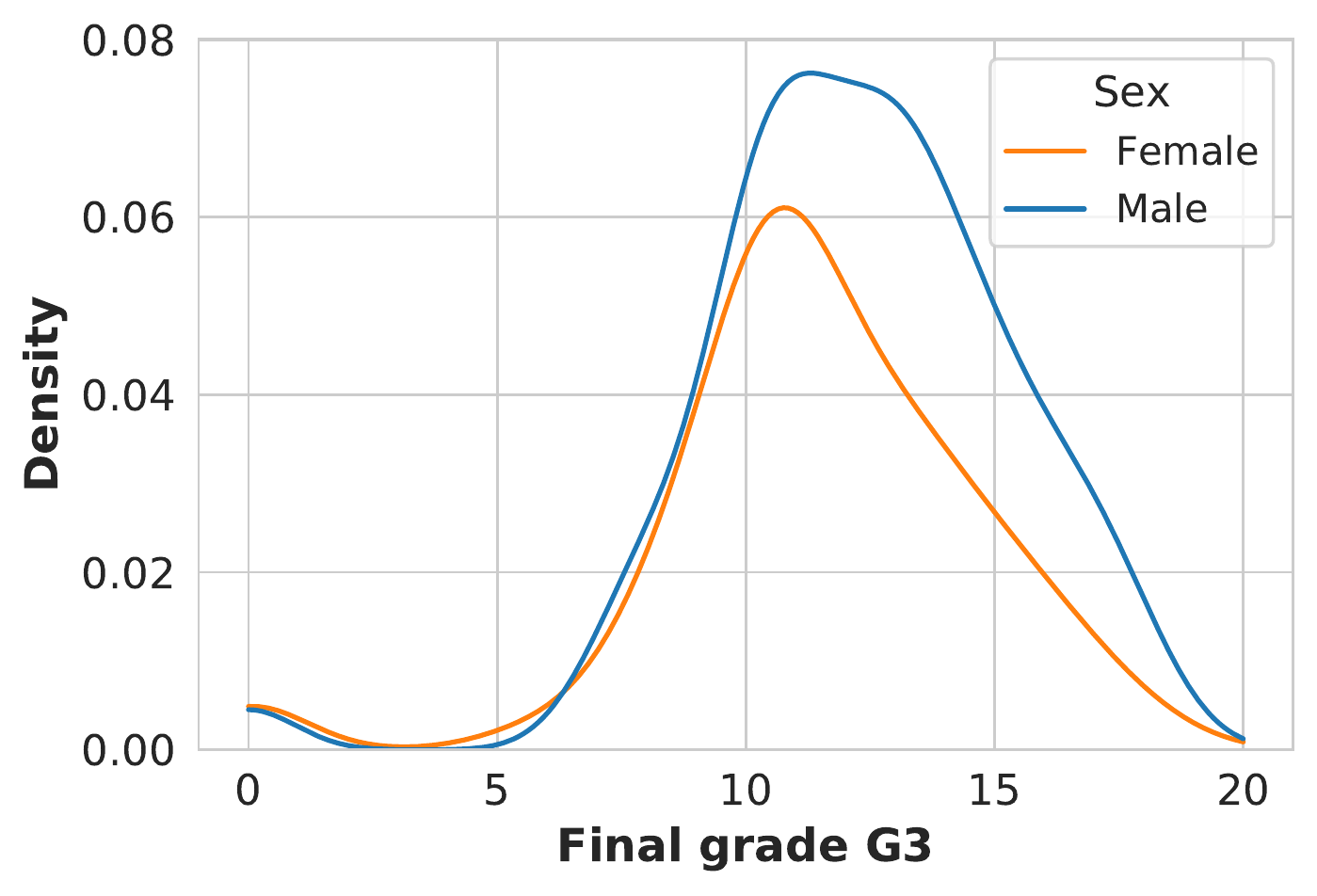}
        \caption{Portuguese subject}
    \end{subfigure}
    \caption{Student performance: Distribution of the final grade G3 w.r.t. \textit{sex}}
    \label{fig:student_G3_sex}
\vspace{-10pt}
\end{figure}

The \textit{class label} attribute is conditionally dependent on the grade \textit{G2} in both subsets (Mathematics and Portuguese subjects). This is explained by a very high correlation coefficient (above 90\%) between \textit{G2} and \textit{G3} variables. In addition, we investigate the distribution of the final grade \textit{G3} on \textit{sex} because the attribute \textit{sex} has an indirect relationship with the \textit{class label}. Figure~\ref{fig:student_G3_sex} 
reveals that the male students tend to receive high scores in the Portuguese subject, while the scores of Math are relatively evenly distributed across both sexes.

\subsubsection{OULAD dataset}
\label{subsubsec:oulad}
The Open University Learning Analytics (OULAD) dataset\footnote{https://analyse.kmi.open.ac.uk/open\_dataset} was collected from the OU analysis project~\cite{kuzilek2017open} of The Open University (England) in 2013 - 2014. The dataset contains information of students and their activities in the virtual learning environment (VLE) for 7 courses. The dataset is investigated in several papers (Appendix~\ref{sec:citation}),
on fairness-aware problems. The goal is to predict the success of students.

\noindent\textbf{Dataset characteristics:}
The dataset contains information of 32,593 students characterized by 12 attributes (7 categorical, 2 binary and 3 numerical attributes). An overview of all attributes is illustrated in Table~\ref{tbl:oulad_attributes}. {Attribute} \textit{id\_student} should be ignored in the analysis. Typically, in the related work, they consider the prediction task on the class label \textit{final\_result} = \{\textit{pass, fail}\}. Therefore, we investigate the cleaned dataset with 21,562 instances after removing the missing values and rows with \textit{final\_result = ``withdrawn''}. {``Pass'' is the positive class.} The ratio of \textit{pass:fail} is 14,655:6,907 (68\%:32\%). {In other words, the dataset is imbalanced with the IR is 2.12:1 (positive:negative).}
\begin{table*}[!ht]
\caption{OULAD: attributes characteristics}
\label{tbl:oulad_attributes}
\begin{adjustbox}{width=1\linewidth}
    \begin{tabular}{llccl}
       \hline
        \multicolumn{1}{c}{\textbf{Attributes}} & \multicolumn{1}{c}{\textbf{Type}} & 
        \multicolumn{1}{c}{\textbf{Values}} & 
        \multicolumn{1}{c}{\textbf{\#Missing values}}&
        \multicolumn{1}{c}{\textbf{Description}}
        \\ \hline
code\_module  & Categorical &  7  & 0 & The identification code of the module on which the student is registered \\
code\_presentation  & Categorical &  4  & 0 & The identification code of the presentation on which the student is registered\\
id\_student & Numerical &  [3,733 - 2,716,795] & 0 & A unique identification number for the student\\
gender  & Binary &    \{Male, Female\} & 0 & Gender\\
region  & Categorical &  13  & 0 & The geographic region\\
highest\_education  & Categorical &  5  & 0 & The highest student education level\\
imd\_band  & Categorical &  10 &  1111 & The index of multiple deprivation (IMD) band of the place where the student lived\\
age\_band  & Categorical &  3 &  0 & The category of the student's age\\
num\_of\_prev\_attempts & Numerical &   [0 - 6] & 0 & The number times the student has attempted this module\\
studied\_credits & Numerical &   [30 - 655] & 0 & The total number of credits for the modules the student is currently studying\\
disability  & Binary &    \{Yes, No\} & 0 & Whether the student has declared a disability\\
final\_result  & Categorical &  4  & 0 & The student’s final result (in the module-presentation)\\
        \hline
    \end{tabular}
\end{adjustbox}
\end{table*}

\noindent\textbf{Protected attributes:}
\textit{gender} = \{\textit{male, female}\} is considered as the protected attribute, in the literature. Male is the majority group with the ratio \textit{male:female} is 11,568:9994 (56.6\%:46.4\%).

\begin{figure}[!htb]
  \centering
  \includegraphics[width=0.6\linewidth]{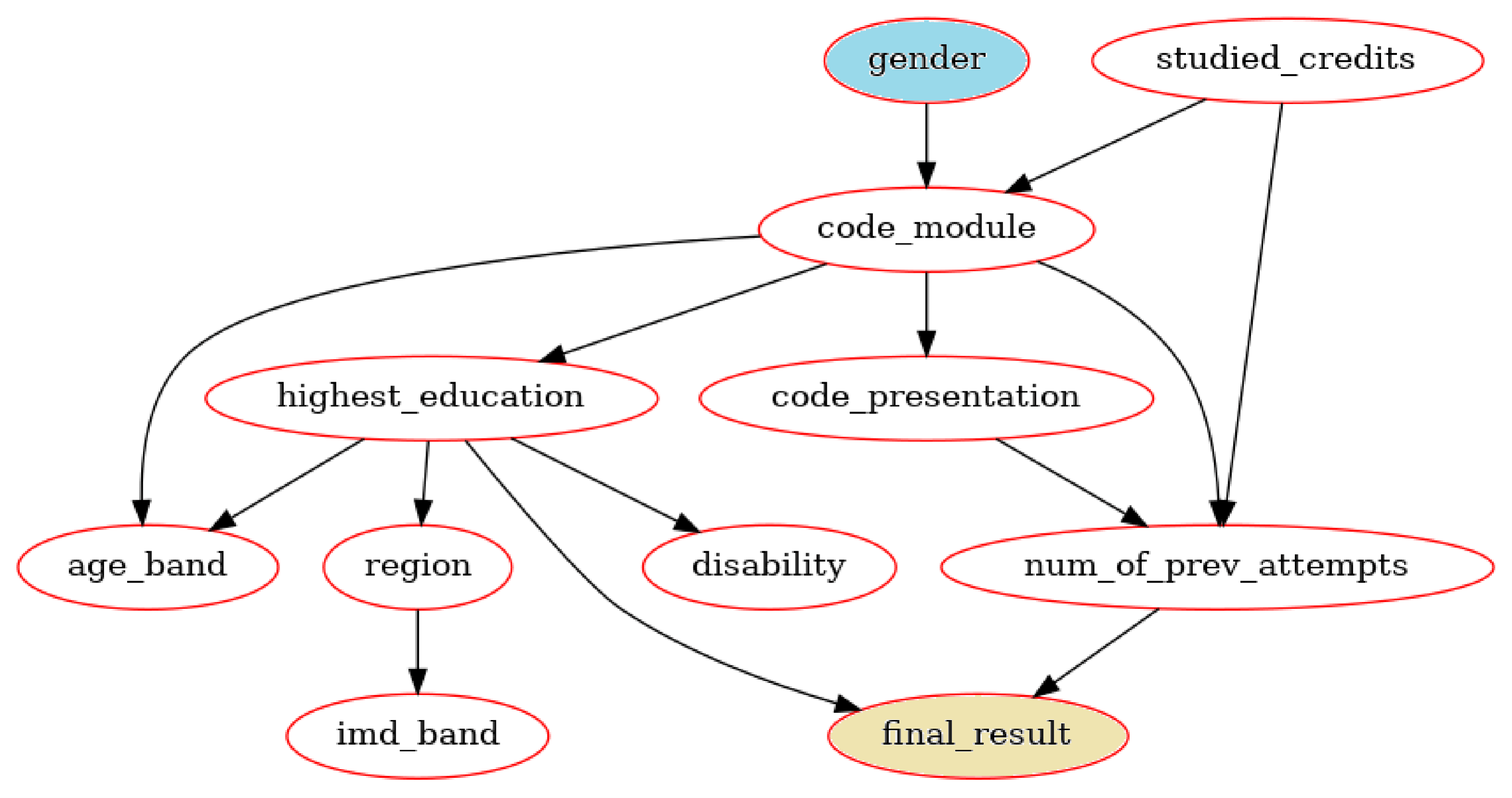}
  \caption{OULAD: Bayesian network (class label: \textit{final\_result}, protected attributes: \textit{gender})}
  \label{fig:OULAD_BN}
\end{figure}

\noindent\textbf{Bayesian network:}
The numerical attributes are encoded for generating the Bayesian network: \textit{num\_of\_prev\_attempts} = \{0, $>$0\}, \textit{studied\_credits} = \{$\leq$100, $>$100\}. The network is depicted in Figure~\ref{fig:OULAD_BN}. The final result attribute is directly conditionally dependent on the highest education level (\textit{highest\_education}) and the number times the student has attempted the module (\textit{num\_of\_prev\_attempts}) attributes, while \textit{gender} has a more negligible effect on the outcome.
\begin{figure}[!htb]
  \centering
  \includegraphics[width=\linewidth]{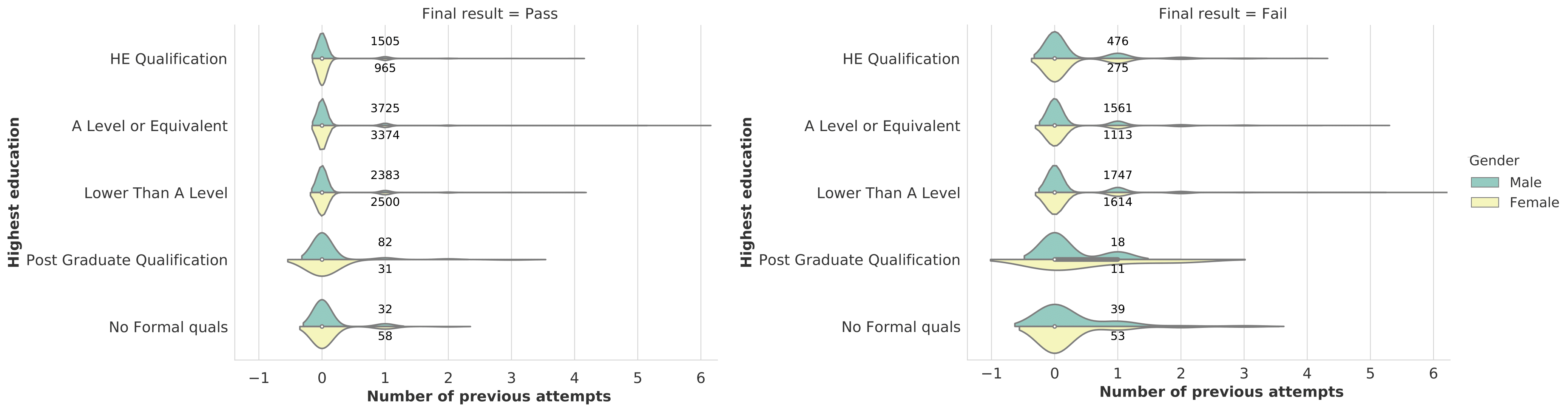}
  \caption{OULAD: Distribution of \textit{the number of previous attempts}, \textit{the highest education} and \textit{the final result} w.r.t. \textit{gender}}
  \label{fig:OULAD_violin}
\end{figure}

We perform the analysis on the relationship of the highest education, number of previous attempts and the final result for each gender. As demonstrated in Figure~\ref{fig:OULAD_violin}, students have a higher probability of failing when they tried to attempt the exam many times in the past. The ratio of male students having the \textit{highest education} is ``A-level or equivalent'' or ``higher education (HE) qualification'' is around 1.5 times higher than that of female students.

\subsubsection{Law school dataset}
\label{subsubsec:college_admission}

The Law school\footnote{https://github.com/tailequy/fairness\_dataset/tree/main/Law\_school} dataset~\cite{wightman1998lsac} was conducted by a  Law School Admission Council (LSAC) survey across 163 law schools in the United States in 1991. The dataset contains the law school admission records. The prediction task is to predict whether a candidate would pass the bar exam or predict a student's first-year average grade (FYA). The dataset is investigated in a variety of studies (Appendix~\ref{sec:citation}).

\noindent\textbf{Dataset characteristics:}
The dataset contains information of 20,798 students characterized by 12 attributes (3 categorical, 3 binary and  6 numerical attributes). An overview of all attributes is depicted in Table~\ref{tbl:law_attributes}. 

\begin{table*}[!h]
\caption{Law schoool: attributes characteristics}
\label{tbl:law_attributes}
\begin{adjustbox}{width=1\linewidth}
    \begin{tabular}{llccl}
        \hline
        \multicolumn{1}{c}{\textbf{Attributes}} & \multicolumn{1}{c}{\textbf{Type}} & 
        \multicolumn{1}{c}{\textbf{Values}} & 
        \multicolumn{1}{c}{\textbf{\#Missing values}} &
        \multicolumn{1}{c}{\textbf{Description}}
        \\ \hline
decile1b & Numerical &   [1.0 - 10.0] & 0 & The student's decile in the school given his grades in Year 1 \\
decile3 & Numerical &   [1.0 - 10.0] & 0 & The student's decile in the school given his grades in Year 3\\
lsat & Numerical &   [11.0 - 48.0] & 0 & The student's LSAT score\\
ugpa & Numerical &   [1.5 - 4.0] & 0 & The student's undergraduate GPA\\
zfygpa & Numerical &   [-3.35 - 3.48] & 0 & The first year law school GPA\\
zgpa & Numerical &   [-6.44 - 4.01] & 0 & The cumulative law school
GPA\\
fulltime & Binary &   \{1, 2\} & 0 & Whether the student will work full-time or part-time\\
fam\_inc & Categorical & 5   & 0 & The student's family income bracket\\
male & Binary &  \{0, 1\} & 0 & Whether the student is a male or female\\
tier & Categorical & 6   & 0 & Tier\\
{race}  & Categorical  & 6   & 0 & Race\\
pass\_bar & Binary &  \{0, 1\} & 0 & Whether the student passed the bar exam on the first try\\
        \hline
    \end{tabular}
\end{adjustbox}
\end{table*}

\noindent\textbf{Protected attributes:}
In the literature, \textit{race}~\cite{bechavod2017penalizing,lahoti2020fairness,russell2017worlds,kusner2017counterfactual,chzhen2020fair,kearns2019empirical,Ruoss2020,yang2020fairness} and \textit{male}~\cite{berk2017convex,lahoti2020fairness,kusner2017counterfactual,kearns2019empirical,yang2020fairness} are considered as the protected attributes.
\begin{itemize}
    \item \textit{male} = \{1, 0\}. \emph{Male} is the majority group. The ratio of \textit{male} (1):\textit{female} (0) is 11,675:9,123 (56.1\%:43.9\%).
    \item \textit{race} = \{\textit{white, black, Hispanic, Asian, other}\}. As introduced in the related work, we encode \textit{race} = \{\textit{white, non-white}\} based on the original attribute. \textit{Non-white} is the minority group with the ratio \textit{white:non-white} is 17,491:3,307 (84\%:16\%).
\end{itemize}

\textbf{Class attribute:} The class label \textit{pass\_bar} = \{0, 1\} is used for the classification task. The positive class is \emph{1 - pass}. The dataset is imbalanced with an imbalance ratio 8.07:1 (positive:negative).

\begin{figure}[h!]
  \centering
  \includegraphics[width=0.45\linewidth]{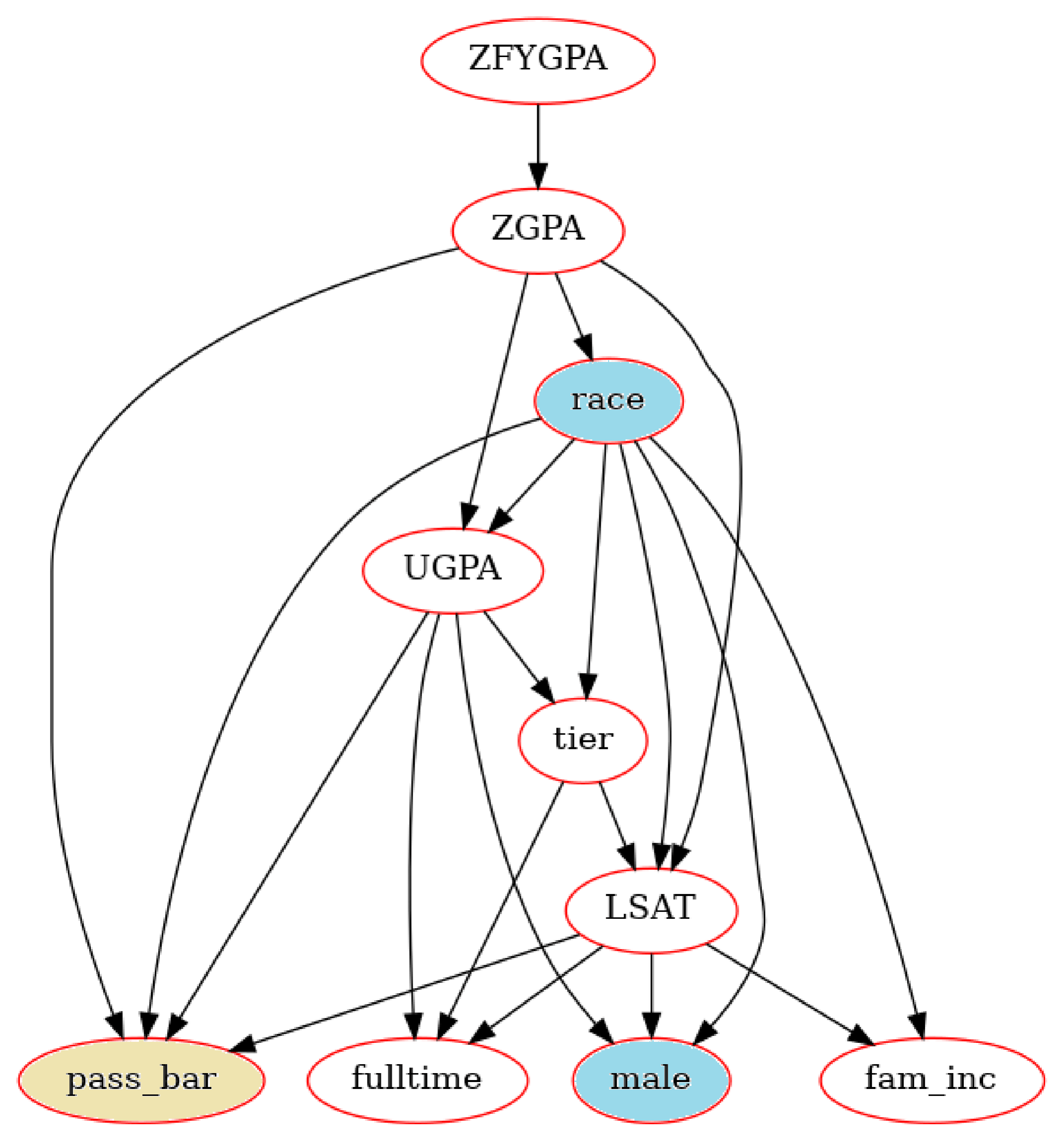}
  \caption{Law school: Bayesian network (class label: \textit{pass\_bar}, protected attributes: \textit{male, race})}
  \label{fig:law_BN}
\end{figure}

\begin{figure}[h!]
    \centering
    \begin{subfigure}[t]{0.4\textwidth}
        \centering
        \includegraphics[width=1\linewidth]{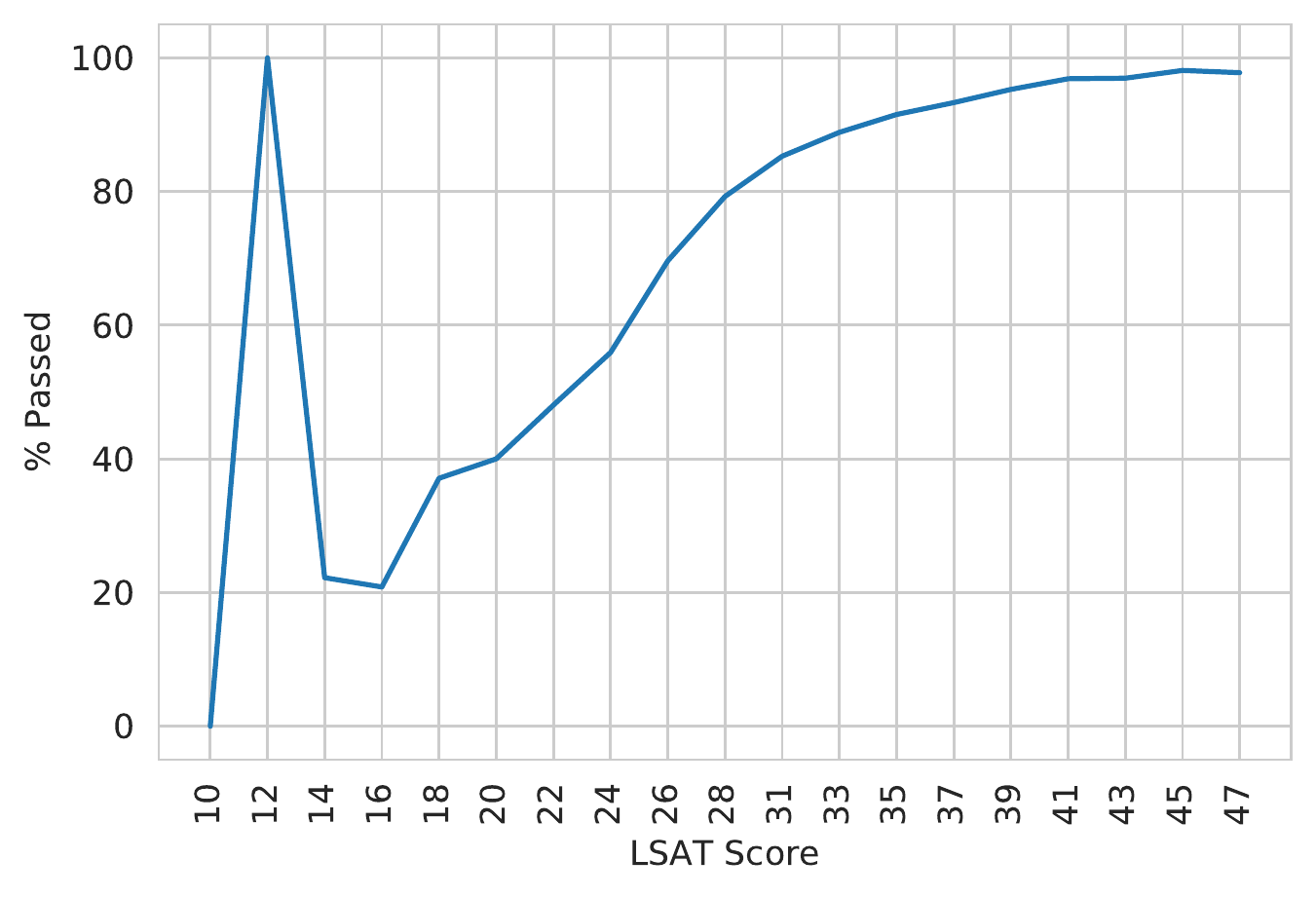}
        \caption{LSAT score}
    \end{subfigure}
    \quad
    \begin{subfigure}[t]{0.4\textwidth}
        \centering
        \includegraphics[width=1\linewidth]{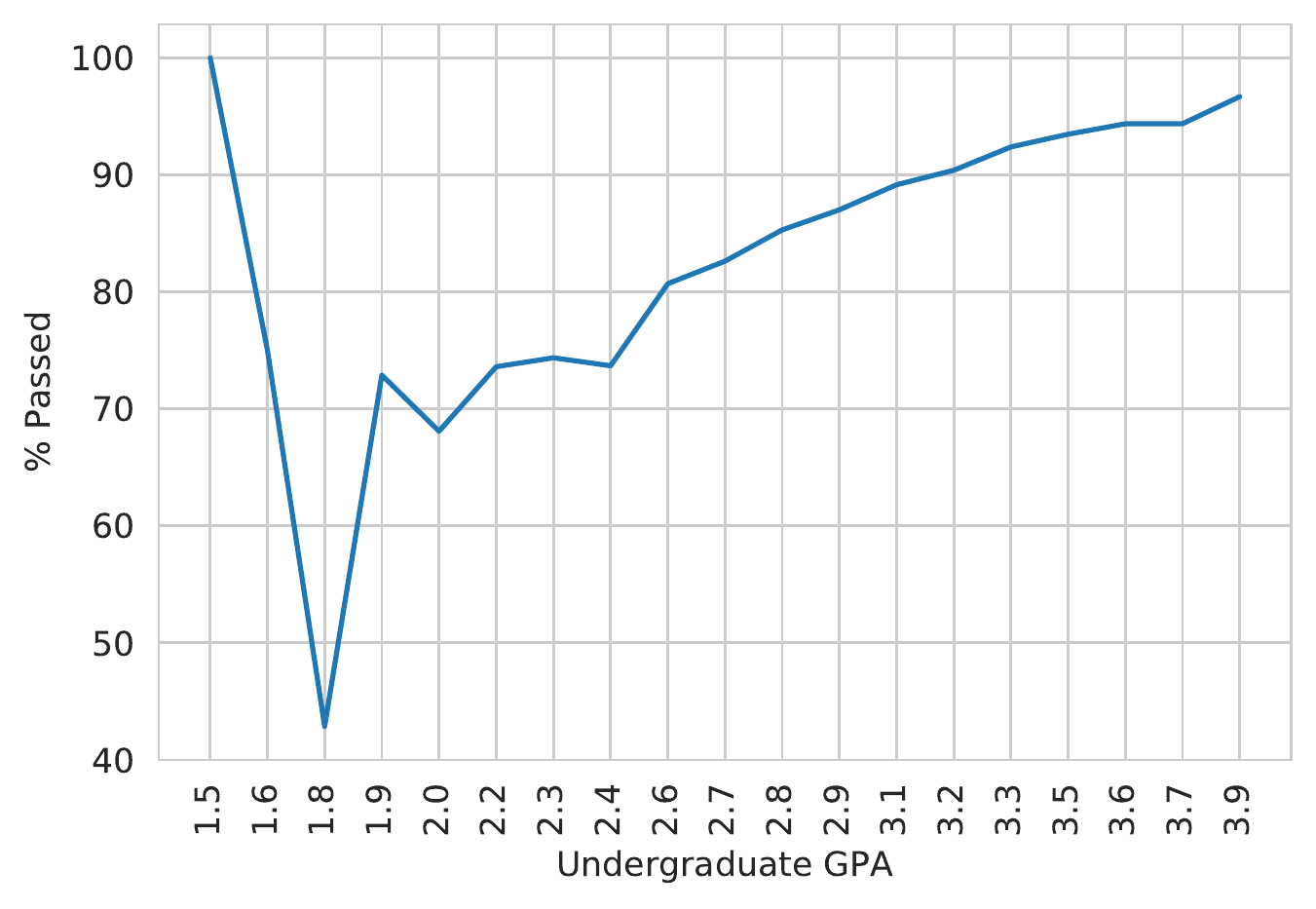}
        \caption{Undergraduate GPA}
    \end{subfigure}
    \caption{Law school: The percentage of students that passed the bar exam by LSAT and UGPA scores}
    \label{fig:law_lsat_ugpa_pass}
\end{figure}

\noindent\textbf{Bayesian network:}
To generate the Bayesian network, we encode the numerical attributes as follows: \textit{decile1b} = \{$\leq$5, $>$5\}, \textit{decile3} = \{$\leq$5, $>$5\}, \textit{lsat} = \{37, $>$37\}, \textit{ugpa} = \{$<$3.3, $\geq$3.3\}, \textit{zgpa} = \{$\leq$0, $>$0\}, \textit{zfygpa} = \{$\leq$0, $>$0\}. The Bayesian network is visualized in Figure~\ref{fig:law_BN}.

It is easy to observe that the {bar exam's result} is conditionally dependent on the law school admission test (LSAT) score, undergraduate grade point average (UGPA)  and \textit{Race}. We discover that 92.1\% of \emph{white} students (16,114/17,491) pass the bar exam, while this ratio in \emph{non-white} students is only 72.3\%. In general, the percentage of students who passed the bar exam is increased in proportion to the LSAT and UGPA scores, which is depicted in Figure~\ref{fig:law_lsat_ugpa_pass}.

\textbf{Summary of the educational datasets:} The educational datasets were collected in many countries around the world. \emph{Gender} is the most popular protected attribute, followed by \emph{age} and \emph{race}. The typical learning task is to predict students' outcome or grades. Therefore, many machine learning tasks are applied to the datasets, such as classification, regression, or clustering. All datasets are imbalanced with very different imbalance ratios in terms of class imbalance. The bias is observed in the datasets w.r.t protected attributes, i.e., \emph{race, sex}; hence, fairness-aware algorithms need to take into account these attributes to achieve fairness in education

%% file: experiment.tex
\section{Experimental evaluation}
\label{sec:experiments}

The goal of our survey is to summarize the different datasets on fairness-aware learning in terms of their application domain, fairness-aware and learning-related challenges. An experimental evaluation of the different fairness-aware learning methods (pre-,in-and post-processing) is beyond the scope of this survey. 
However, in order to characterize the different datasets in terms of the difficulty of the fairness-aware learning task, in this section, we present a short fairness-vs-predictive performance evaluation\footnote{The source code is available at: https://github.com/tailequy/fairness\_dataset} using a popular classification method (namely, logistic regression).

\subsection{Evaluation setup}
\label{subsec:setup}
\noindent\textbf{Predictive model}. As our classification model, we use  {\textit{logistic regression}~\cite{cox1958regression}}, a statistical model using a logistic function to model a binary dependent variable. To simplify the task, we apply the logistic regression model to the binary classification problem.

\noindent\textbf{Metrics}. Based on the confusion matrix in Figure~\ref{fig:confusion_matrix} (in which, \textit{prot} and \textit{non-prot} stand for \textit{protected, non-protected}, respectively), we report the performance of the predictive model on the following measures.

\newcommand\MyBox[2]{
  \fbox{\lower0.75cm
    \vbox to 1.6cm{\vfil
      \hbox to 3.0cm{\hfil\parbox{2.4cm}{#1\\#2}\hfil}
      \vfil}%
  }%
}
\begin{center}

\renewcommand\arraystretch{1.5}
\setlength\tabcolsep{0pt}
\begin{figure}[H]
\centering
\begin{tabular}{c >{\bfseries}r @{\hspace{0.7em}}c @{\hspace{0.4em}}c @{\hspace{0.7em}}l}
  \multirow{10}{*}{\rotatebox{90}{\parbox{2.4cm}{\bfseries\centering Actual class\\\phantom{123} }}} & 
    & \multicolumn{2}{c}{\bfseries Predicted class} & \\
  & & \bfseries Positive  & \bfseries Negative &  \\
  & Positive & \MyBox{\centering True Positive (TP)}{$TP_{prot} + TP_{non-prot}$} & \MyBox{\centering False Negative (FN)}{$FN_{prot} + FN_{non-prot}$} &  \\ [2.7em]
  & Negative & \MyBox{\centering False Positive (FP)}{$FP_{prot} + FP_{non-prot}$} & \MyBox{\centering True Negative (TN)}{$TN_{prot} + TN_{non-prot}$} &  \\
  
\end{tabular}
\caption{The confusion matrix, including protected/ non-protected groups.}
\label{fig:confusion_matrix}
\end{figure}
\end{center}

\begin{itemize}
    \item Accuracy
    \begin{equation}
    {Accuracy} = \frac{TP + TN}{TP + TN + FP + FN}   
    \end{equation}
    \item Balanced accuracy
    \begin{equation}
    {Balanced \: accuracy}  = \frac{1}{2}\times \left(\frac{TP}{TP+FN}+\frac{TN}{TN+FP}    \right)
    \end{equation}
    \item True positive rate (TPR) on protected group
    \begin{equation}
    TPR_{prot} = \frac{TP_{prot}}{TP_{prot} + FN_{prot}}   
    \end{equation}
    \item TPR on non-protected group
    \begin{equation}
    TPR_{non-prot} = \frac{TP_{non-prot}}{TP_{non-prot} + FN_{non-prot}}   
    \end{equation}
    \item True negative rate (TNR) on protected group
    \begin{equation}
    TNR_{prot} = \frac{TN_{prot}}{TN_{prot} + FP_{prot}}   
    \end{equation}
    \item TNR on non-protected group
    \begin{equation}
    TNR_{non-prot} = \frac{TN_{non-prot}}{TN_{non-prot} + FP_{non-prot}}   
    \end{equation}
    \item Statistical parity (Eq. \ref{eqn:statistical_parity})
    \item Equalized odds (Eq. \ref{eq:equalized_odds})
    \item ABROCA (Eq. \ref{eq:abroca})
\end{itemize}

\noindent\textbf{Training/test set spliting}. The ratio of training set and test set in our experiment is 70\%:30\% (single split) applied for each dataset. 

\begin{figure*}[!htbp]
\centering
\vspace{-3pt}
\begin{subfigure}{.28\linewidth}
    \centering
    \includegraphics[width=\linewidth]{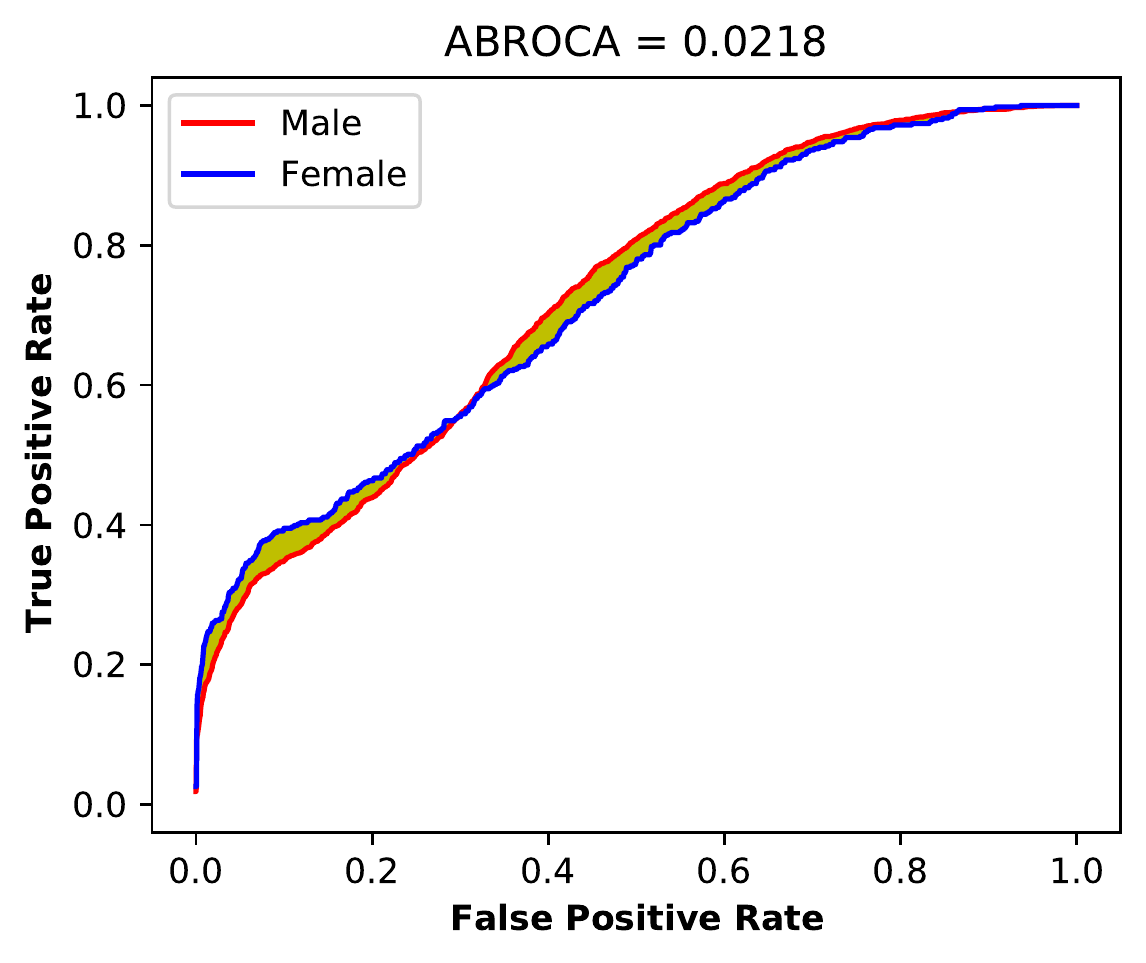}
    \caption{Adult}\label{fig:adult_abroca}
\end{subfigure}
\hfill
\vspace{-3pt}
\begin{subfigure}{.28\linewidth}
    \centering
    \includegraphics[width=\linewidth]{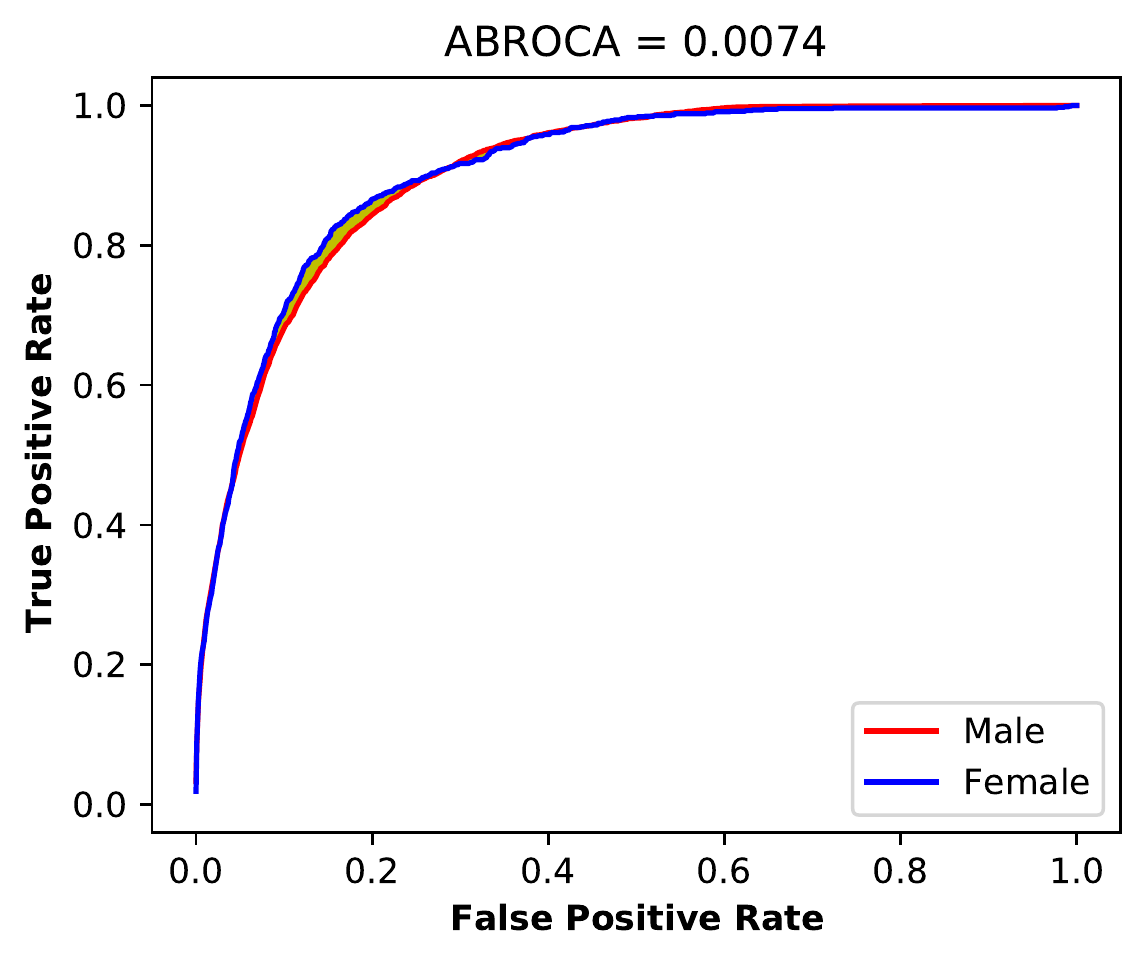}
    \caption{KDD Census-Income}\label{fig:kdd_abroca}
\end{subfigure}    
\hfill
\begin{subfigure}{.28\linewidth}
    \centering
    \includegraphics[width=\linewidth]{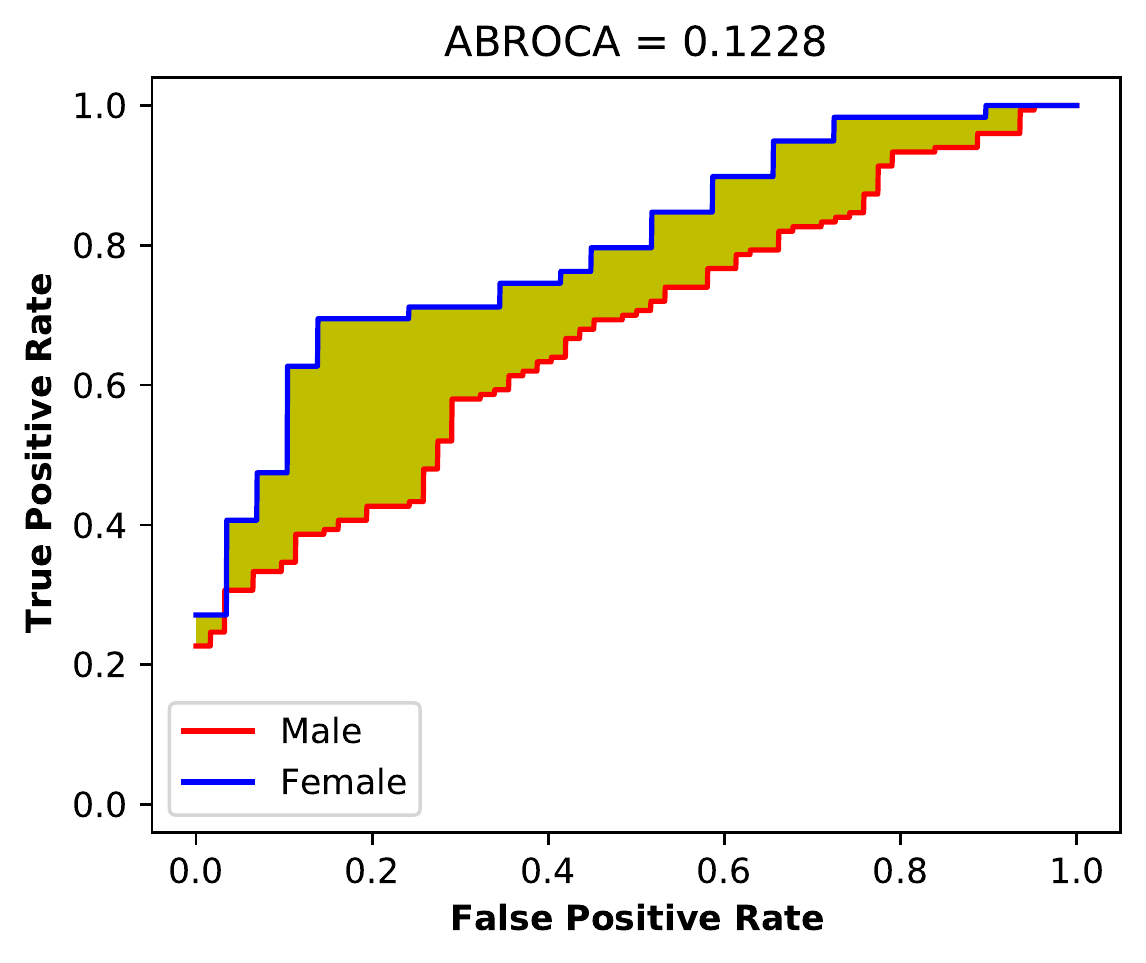}
    \caption{German credit}\label{fig:german_abroca}
\end{subfigure}
\bigskip
\vspace{-3pt}
\begin{subfigure}{.28\linewidth}
    \centering
    \includegraphics[width=\linewidth]{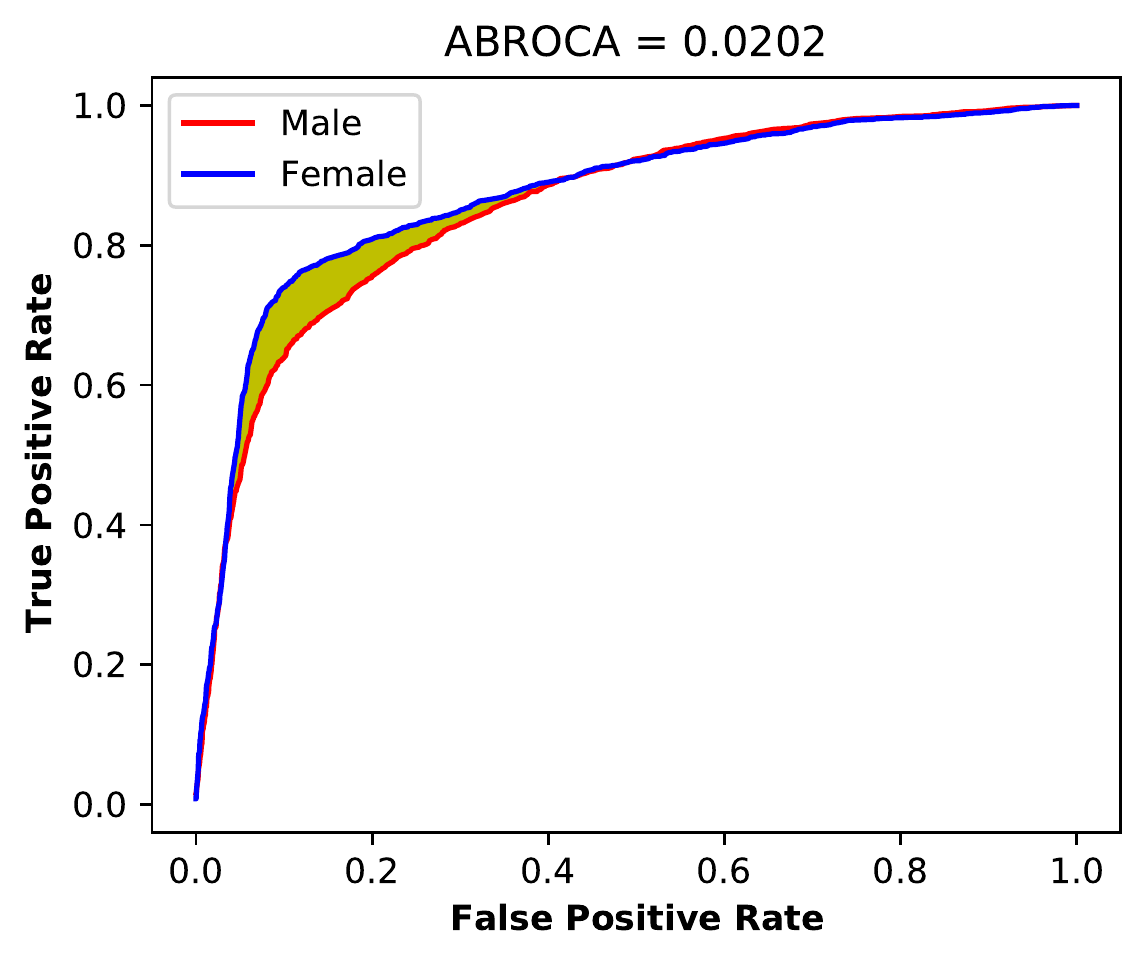}
    \caption{Dutch census}\label{fig:dutch_abroca}
\end{subfigure}
\hfill
\vspace{-3pt}
\begin{subfigure}{.28\linewidth}
    \centering
    \includegraphics[width=\linewidth]{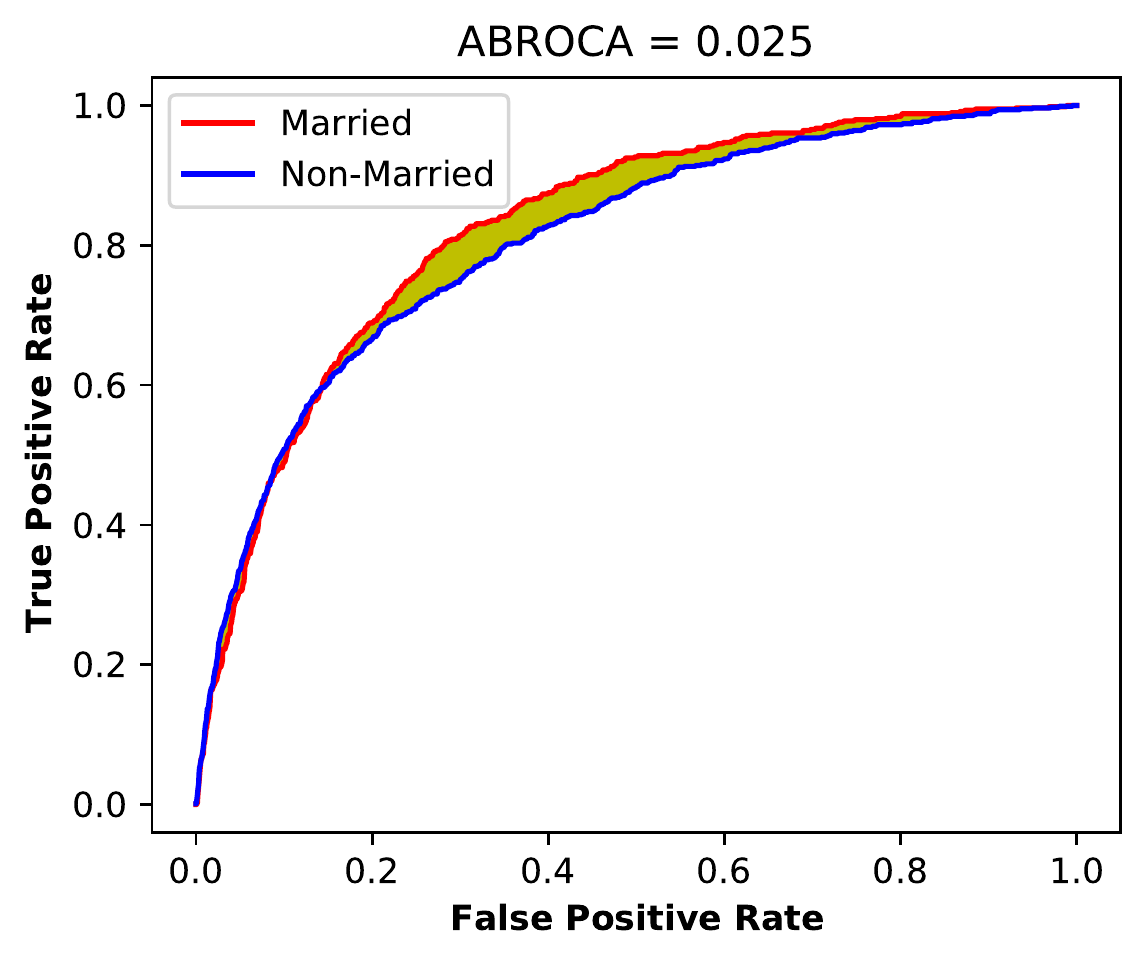}
    \caption{Bank marketing}\label{fig:bank_abroca}
\end{subfigure}
\vspace{-3pt}
 \hfill
\begin{subfigure}{.28\linewidth}
    \centering
    \includegraphics[width=\linewidth]{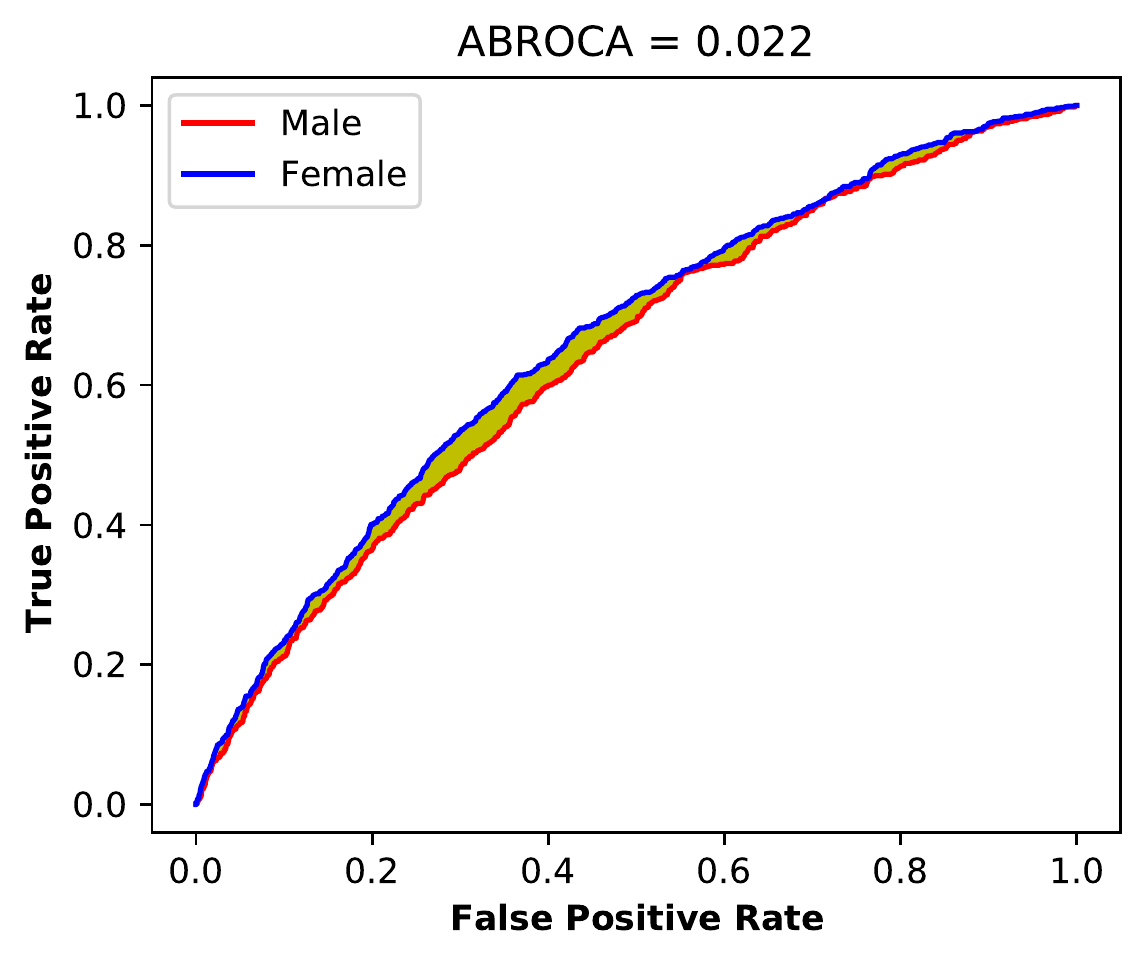}
    \caption{Credit card clients}\label{fig:credit_abroca}
\end{subfigure}
\bigskip
\vspace{-3pt}
\begin{subfigure}{.28\linewidth}
    \centering
    \includegraphics[width=\linewidth]{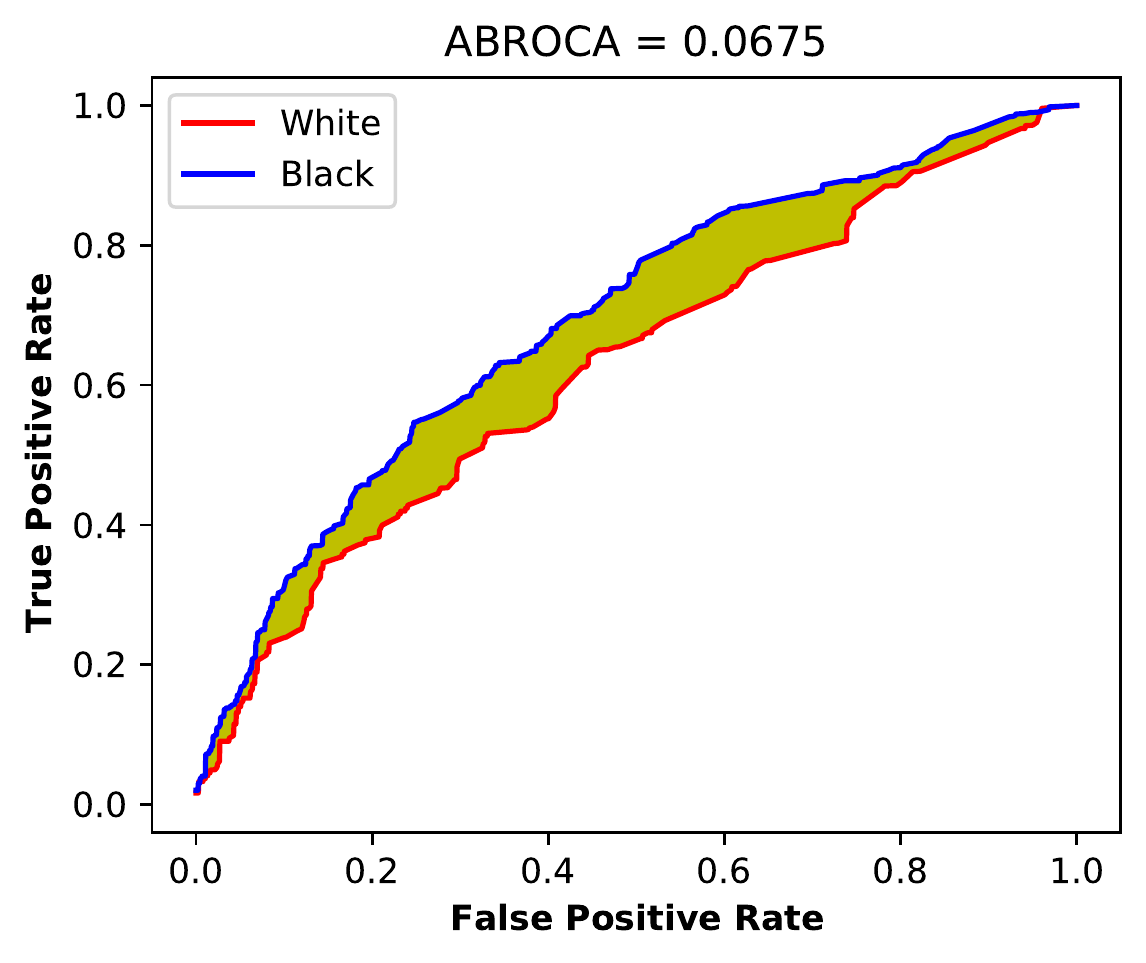}
    \caption{COMPAS recid.}\label{fig:compas_recid_abroca}
\end{subfigure}
   \hfill
\vspace{-3pt}
\begin{subfigure}{.28\linewidth}
    \centering
    \includegraphics[width=\linewidth]{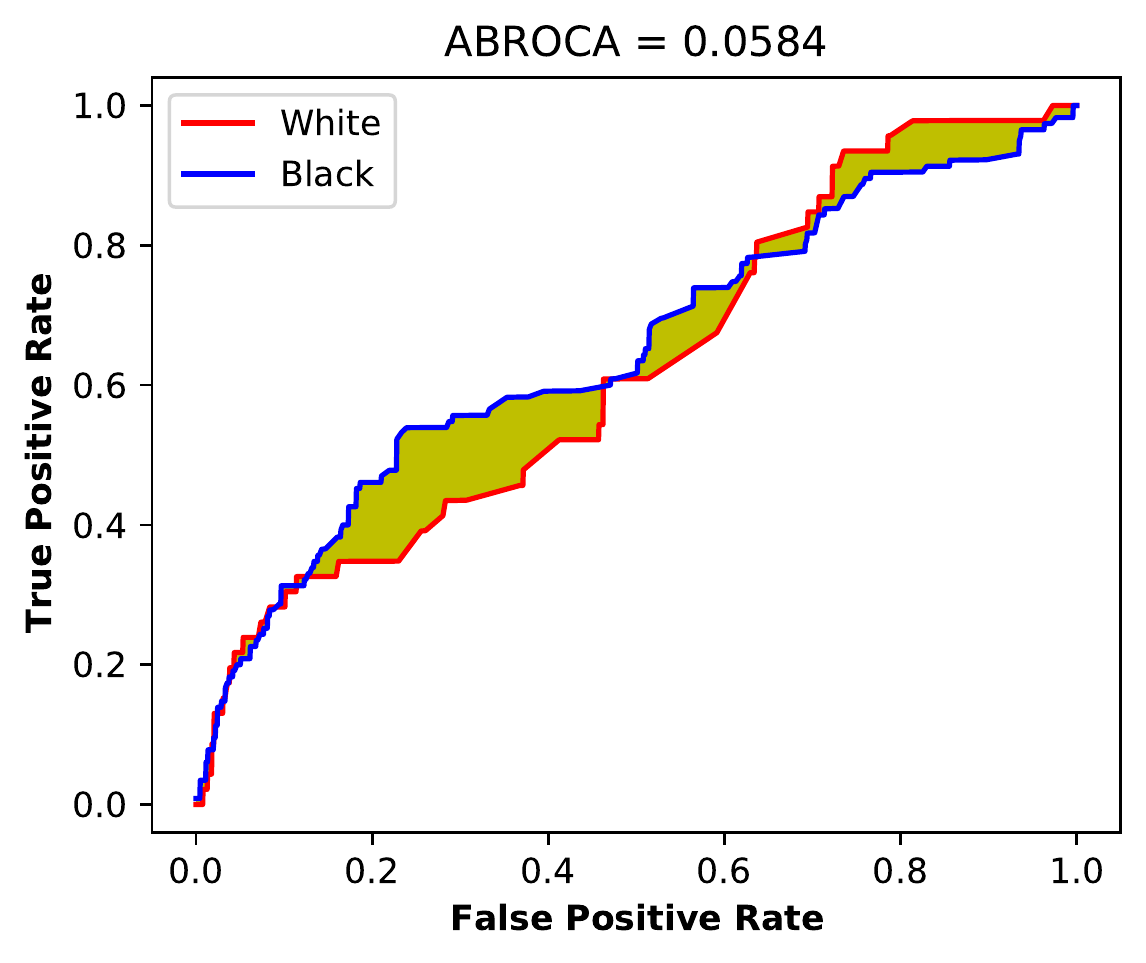}
    \caption{COMPAS viol. recid.}\label{fig:compas_viol_abroca}
\end{subfigure}
\hfill
\vspace{-3pt}
\begin{subfigure}{.28\linewidth}
    \centering
    \includegraphics[width=\linewidth]{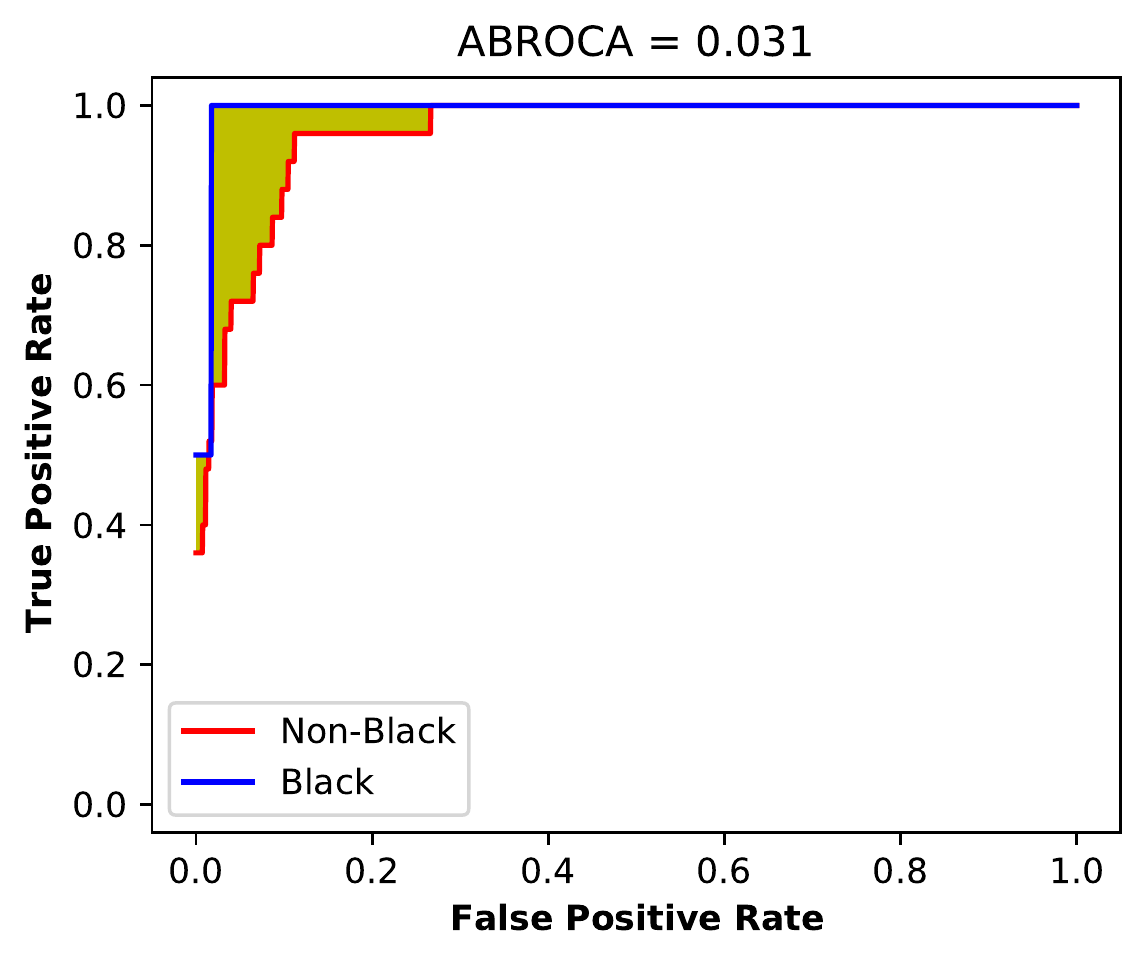}
    \caption{Communities \& Crime}\label{fig:crime_abroca}
\end{subfigure}
\bigskip
\vspace{-3pt}
\begin{subfigure}{.28\linewidth}
    \centering
    \includegraphics[width=\linewidth]{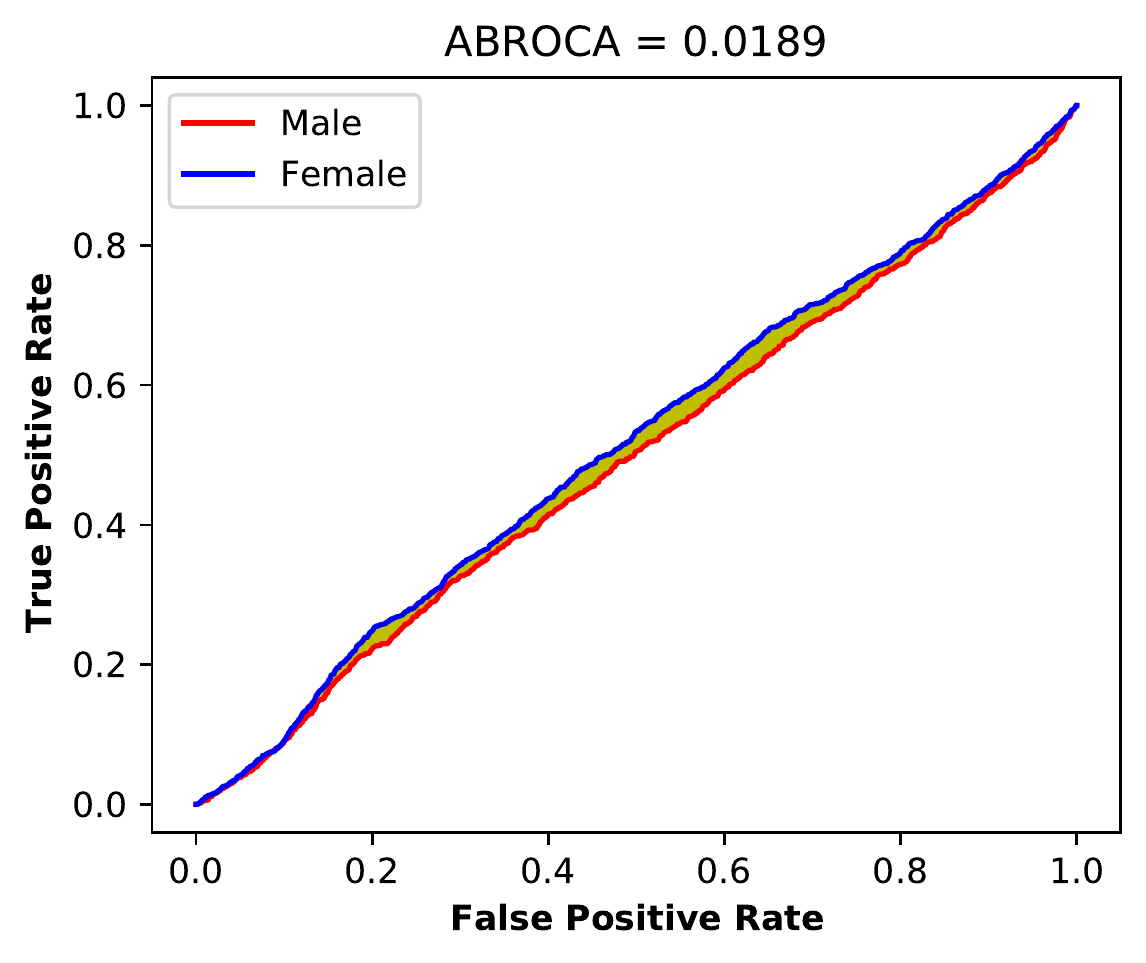}
    \caption{Diabetes}\label{fig:diabetes_abroca}
\end{subfigure}
    \hfill
\vspace{-3pt}
\begin{subfigure}{.28\linewidth}
    \centering
    \includegraphics[width=\linewidth]{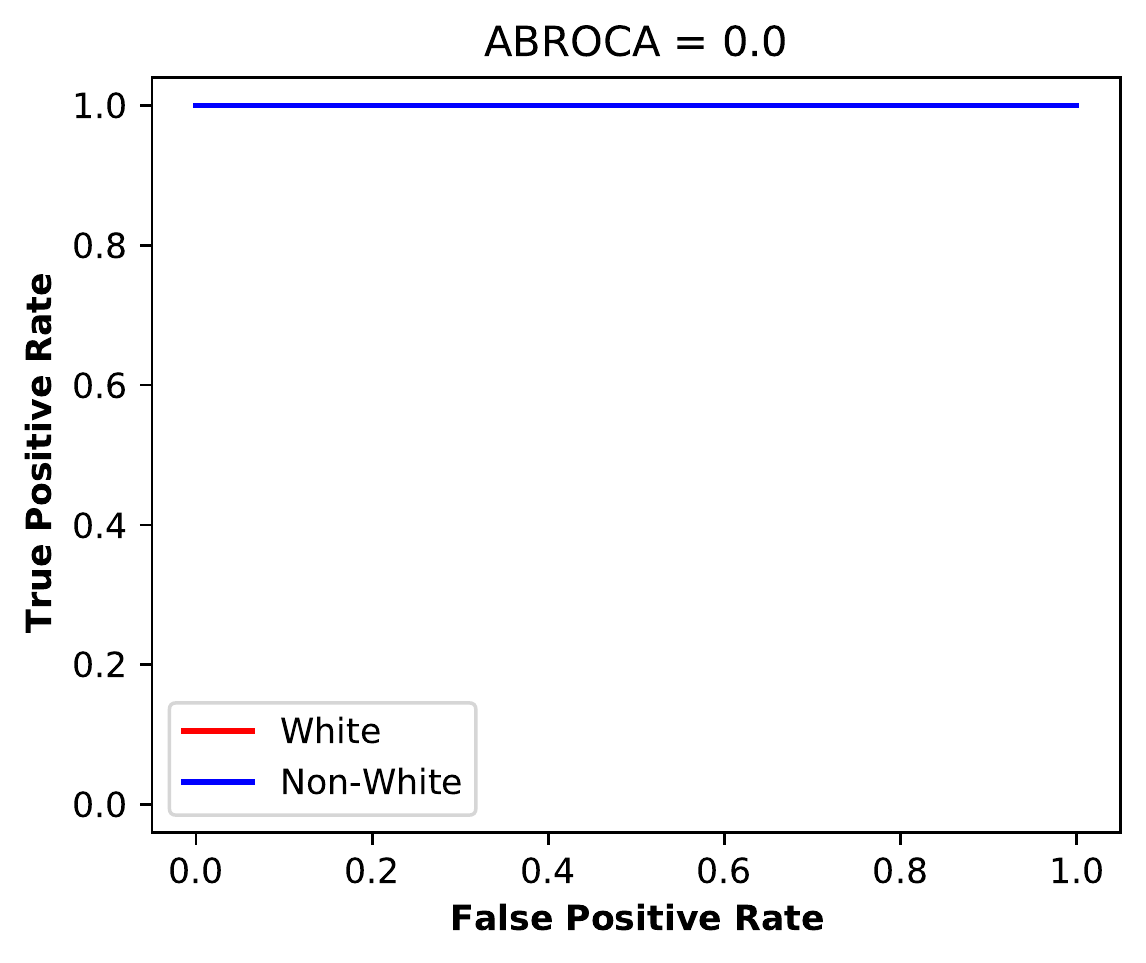}
    \caption{Ricci}\label{fig:ricci_abroca}
\end{subfigure}
   \hfill
\vspace{-3pt}
\begin{subfigure}{.28\linewidth}
    \centering
    \includegraphics[width=\linewidth]{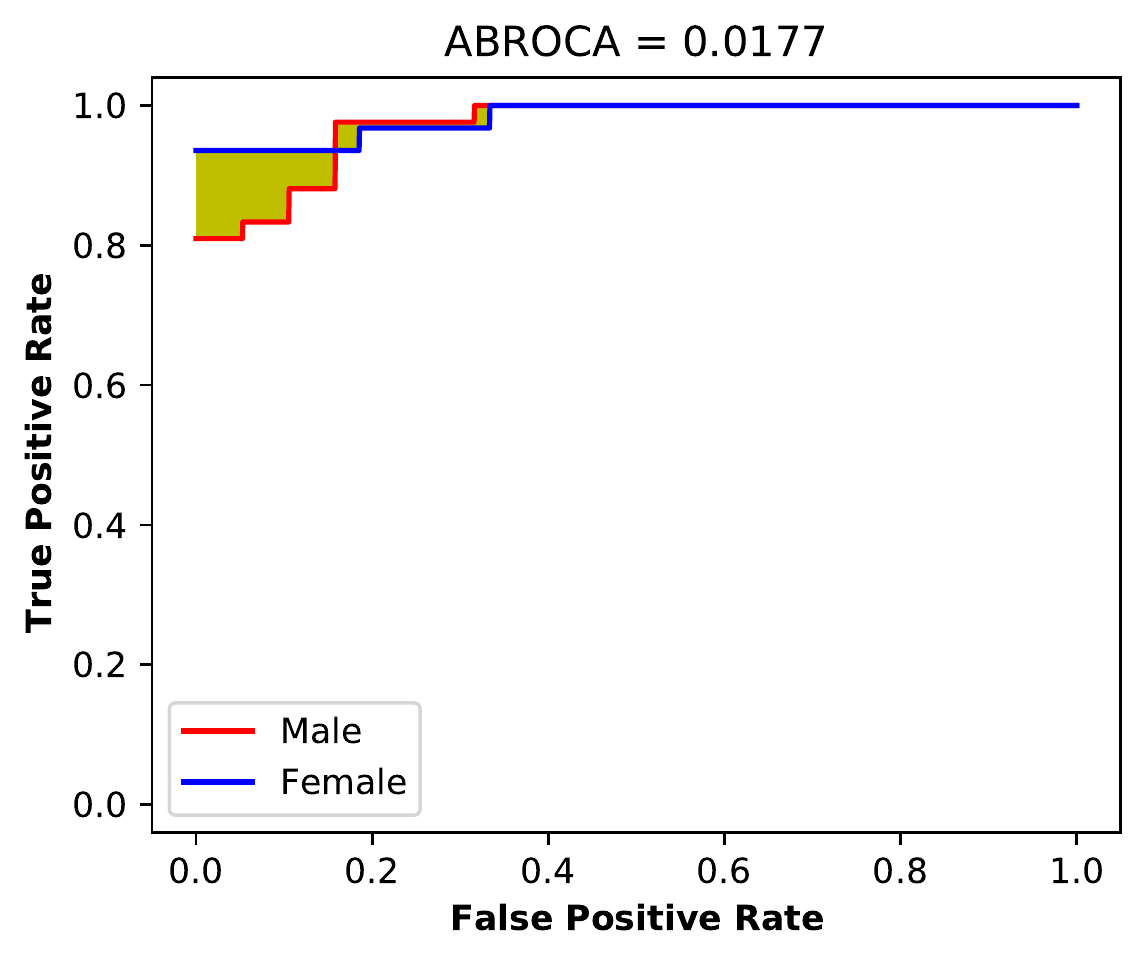}
    \caption{Student-Mathematics}\label{fig:student_mat}
\end{subfigure}
\bigskip  
\vspace{-3pt}
\begin{subfigure}{.28\linewidth}
    \centering
    \includegraphics[width=\linewidth]{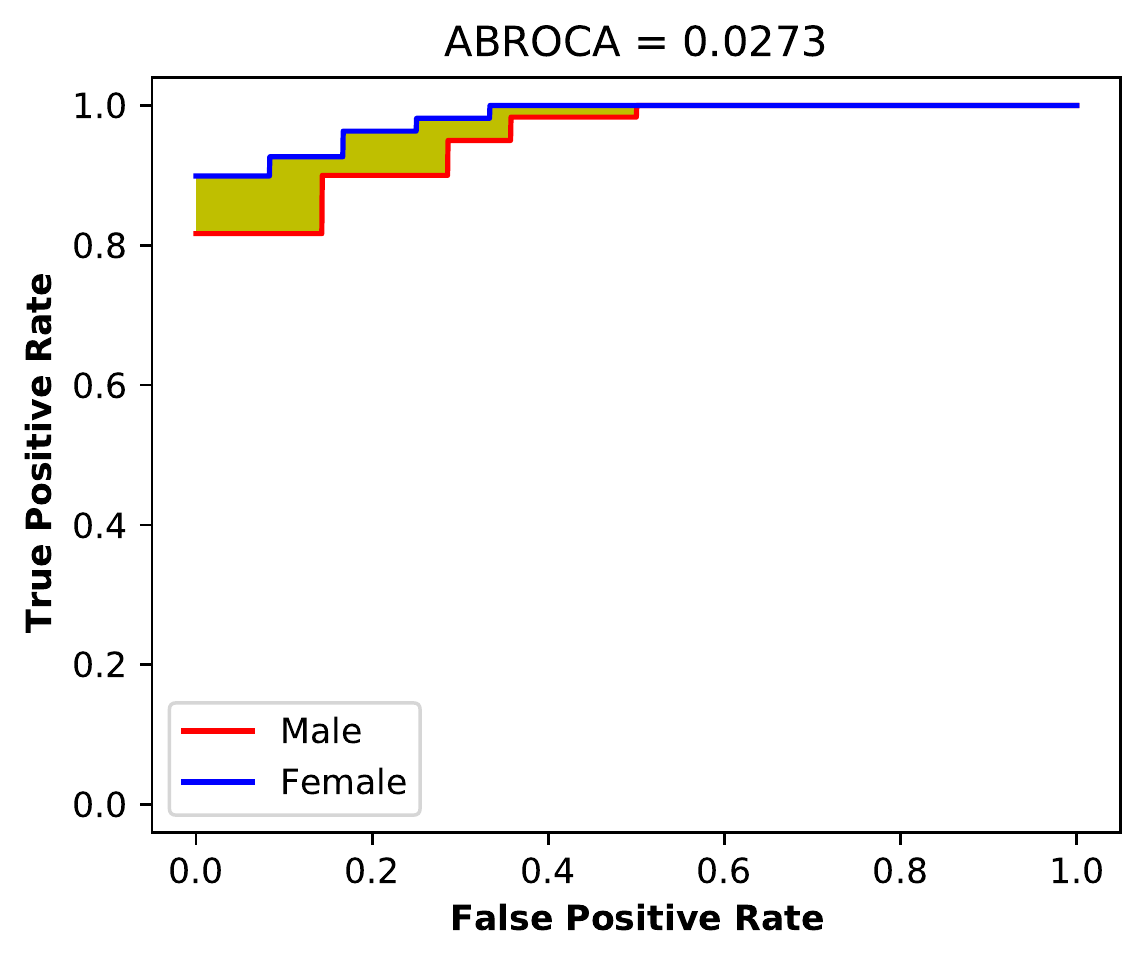}
    \caption{Student-Portuguese}\label{fig:student_por}
\end{subfigure}
\hfill
\vspace{-3pt}
\begin{subfigure}{.28\linewidth}
    \centering
    \includegraphics[width=\linewidth]{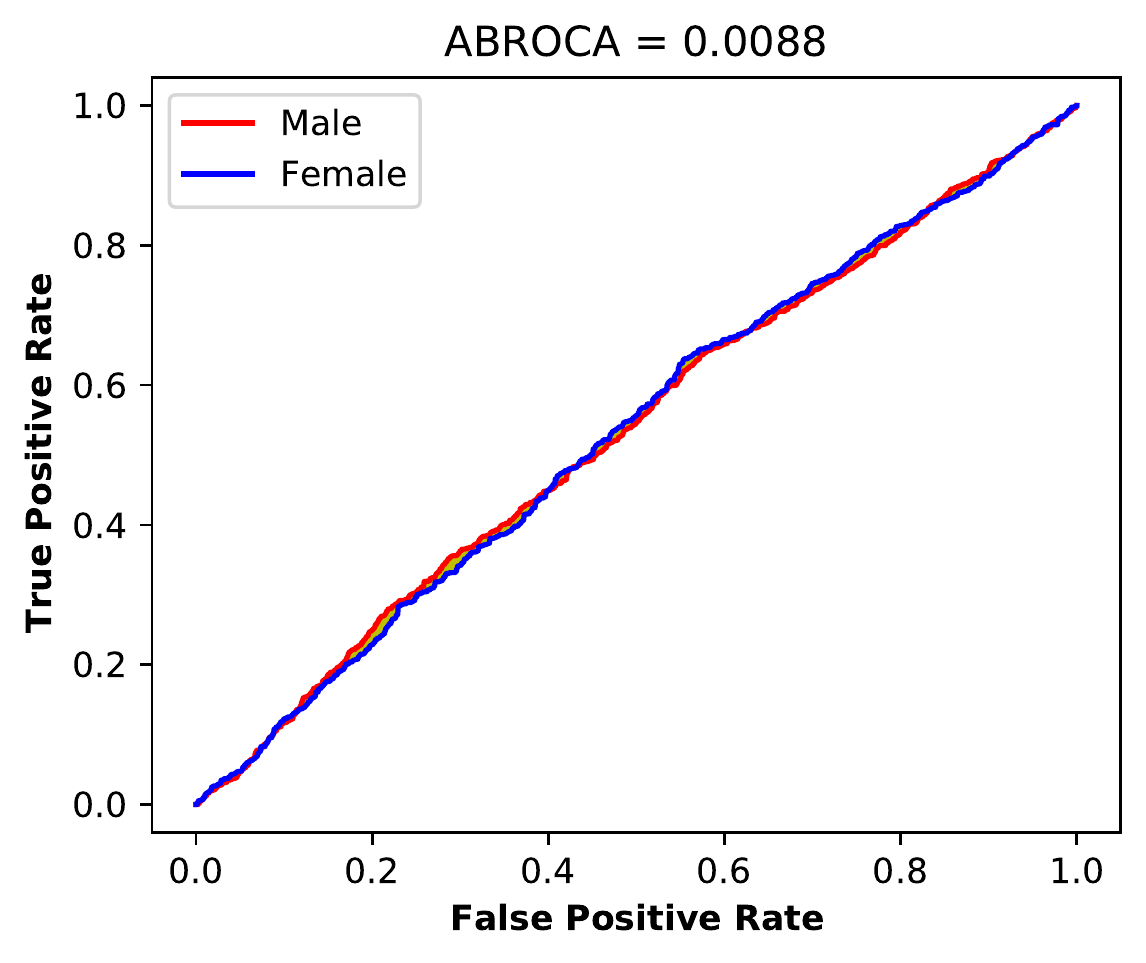}
    \caption{OULAD}\label{fig:oulad_abroca}
\end{subfigure}
    \hfill
\vspace{-3pt}
\begin{subfigure}{.28\linewidth}
    \centering
    \includegraphics[width=\linewidth]{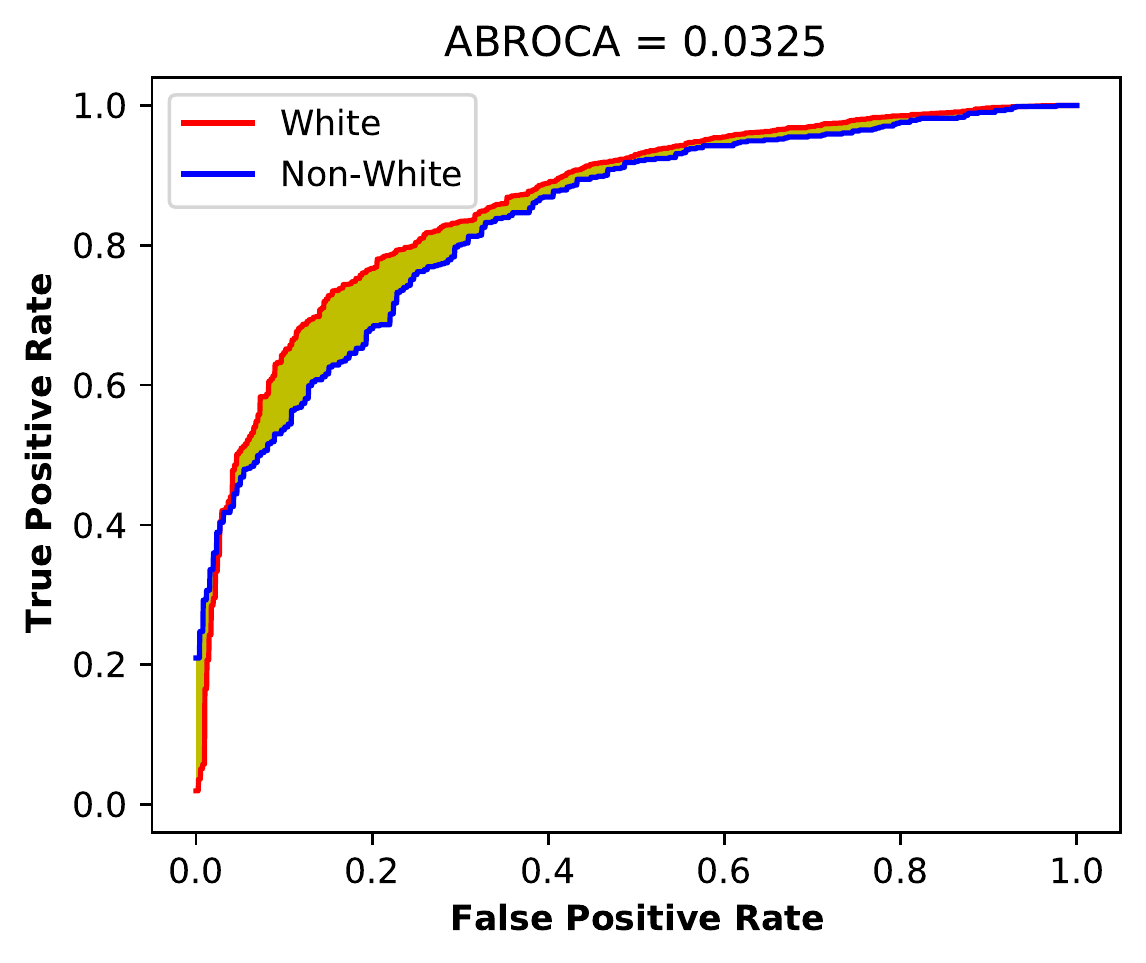}
    \caption{Law school}\label{fig:law_abroca}
\end{subfigure}
\caption{ABROCA slice plot on datasets}
\label{fig:abroca1}
\end{figure*}

\subsection{Experimental results}
\label{subsec:results}
Table~\ref{tbl:experiments} describes the performance of the logistic regression model on all datasets. We believe that our experimental results can be considered as the baseline for the researchers' future studies.
\begin{table*}[!htb]
\centering
\caption{Predictive- and fairness-relatedperformance of logistic regression model}
\label{tbl:experiments}
\begin{adjustbox}{width=1\textwidth}
    \begin{tabular}{lccccccccccc}
        \hline
        \multicolumn{1}{c}{\textbf{Dataset}} &
        \multicolumn{1}{c}{\begin{tabular}[c]{@{}c@{}}\textbf{Protected}\\\textbf{attribute}\end{tabular}} &
        \multicolumn{1}{c}{\begin{tabular}[c]{@{}c@{}c@{}}\textbf{Group}\\\textbf{distribution (\%)}\\{[$s_{+}$, $s_{-}$, $\overline{s}_{+}$, $\overline{s}_{-}$]}\end{tabular}} &
        \multicolumn{1}{c}{\textbf{Accuracy}} &
        \multicolumn{1}{c}{\begin{tabular}[c]{@{}c@{}}\textbf{Balanced}\\\textbf{accuracy} \end{tabular}} &
        \multicolumn{1}{c}{\begin{tabular}[c]{@{}c@{}}\textbf{Statistical}\\\textbf{Parity} \end{tabular}} &
        \multicolumn{1}{c}{\begin{tabular}[c]{@{}c@{}}\textbf{Equalized}\\\textbf{odds} \end{tabular}} &
        \multicolumn{1}{c}{\textbf{ABROCA}} &
        \multicolumn{1}{c}{\begin{tabular}[c]{@{}c@{}}\textbf{TPR}\\\textbf{prot.} \end{tabular}} &
        \multicolumn{1}{c}{\begin{tabular}[c]{@{}c@{}}\textbf{TPR}\\\textbf{non-prot.} \end{tabular}} &
        \multicolumn{1}{c}{\begin{tabular}[c]{@{}c@{}}\textbf{TNR}\\\textbf{prot.} \end{tabular}} &
        \multicolumn{1}{c}{\begin{tabular}[c]{@{}c@{}}\textbf{TNR}\\\textbf{non-prot.} \end{tabular}} \\
        \hline
        Adult & Sex & [3.7, 28.8, 21.1, 46.4] & 0.7864  & 0.6249  & {0.0555}  & 0.0281 & 0.0218 & 0.3194 & 0.3007 & 0.9521 & 0.9426\\
        KDD Census-Income & Sex & [1.3, 50.7, 4.8, 43.2]  & 0.9474 &  0.6031  & {0.0198} & 0.0403 & 0.0074 & 0.1825 & 0.2195 & 0.9961 & 0.9928\\
        German credit & Sex & [20.1, 10.9, 49.9, 19.1] & 0.6967 & 0.5713 & {-0.0770} & 0.1634 & 0.1228 & 0.9831 & 0.8533 & 0.2759 & 0.2419\\
        Dutch census & Sex & [16.4, 33.7, 31.2, 18.7] & 0.8149 & 0.8138 & {0.3568} & 0.3746 & 0.0202 & 0.6984 & 0.8382 & 0.9219 & 0.6871 \\
        Bank marketing & Marital& [5.6, 34.2, 6.1, 54.1] & 0.8855  & 0.5720 & {0.0153}  & 0.0261 & 0.025 & 0.1527 & 0.1726 & 0.9849 & 0.9787\\
        Credit card clients & Sex & [12.5, 47.8, 9.7, 30.0] & 0.7822 & 0.5 & \textbf{{0.0}} & \textbf{0.0} & 0.0220 & 0.0 & 0.0  & \textbf{1.0} & \textbf{1.0}\\ 
        COMPAS recid. & Race & [31.5, 28.7, 15.5, 24.3] & 0.6414 & 0.6299 & {-0.3398} & 0.6452 & 0.0675 & 0.5996 & 0.2058 & 0.6793  &  0.9307\\
        COMPAS viol. recid. & Race & [12.0, 44.8, 5.2, 38.0] & 0.8432 & 0.5541  & {-0.0659} & 0.2195 & 0.0584 & 0.1826 &  0.0 & 0.9606 & 0.9975\\
        Communites \& Crime & Black & [5.8, 46.3, 0.3, 47.6] & 0.9683  & 0.7011  & {0.0396} & 0.4507 & 0.031 & 0.0 & 0.44 & 1.0 & 0.9892\\
        Diabetes & Gender & [11.1, 34.1, 13.1, 41.7] & 0.7584 & 0.5 &  \textbf{{0.0}}  & \textbf{0.0} & 0.0189 & 0.0 & 0.0 & \textbf{1.0}& \textbf{1.0}\\ 
        Ricci & Race & [12.7, 29.7, 34.7, 22.9] & \textbf{1.0} & \textbf{1.0} & {0.1714} & \textbf{0.0} & \textbf{0.0} & \textbf{1.0} & \textbf{1.0} & \textbf{1.0}& \textbf{1.0}\\ 
        Student - Mathematics &Sex & [33.7, 19.0, 33.4, 13.9]  & 0.9412 & 0.9360 & {0.2041} & 0.1616 & 0.0177 & 0.9354  & 0.9762  & 0.9630 & 0.8421\\
        Student - Portuguese & Sex & [51.3, 7.7, 33.3, 7.7]  & 0.9282 & 0.8447 & {-0.0682} & 0.0490 & 0.0273 &  0.9633 & 0.95 & 0.75 & 0.7143\\
        OULAD & Gender & [32.1, 14.2, 35.9, 17.8] & 0.6751 & 0.5 & \textbf{{0.0}} & \textbf{0.0} & 0.0088 & \textbf{1.0} & \textbf{1.0} & 0.0 & 0.0\\ 
        Law School & Race & [11.5, 4.4, 77.5, 6.6]  & 0.9072 & 0.6260 & {0.1937} & 0.5043 & 0.0325 & 0.9100 & 0.9955 & 0.5251 & 0.1063\\ 
        \hline
    \end{tabular}
\end{adjustbox}
\end{table*}

In general, a significant difference in terms of predictive performance and fairness measures is observed between the datasets. In particular, the \textit{Ricci} dataset is an exception where the performance of the predictive model reaches the peak regarding both accuracy and fairness measures. Apart from that, the logistic regression model shows the best performance on the \textit{Communities \& Crime} dataset in terms of accuracy. The worst accuracy is seen in the result of the model on the \textit{OULAD} dataset. Regarding balanced accuracy, the \textit{Student - Mathematics} is the dataset showing the best result of the predictive model, followed by the \textit{Student - Portuguese} and the \textit{Dutch census} datasets. Logistic regression model shows the worst balanced accuracy on the \textit{Credit card clients, Diabetes} and \textit{OULAD} datasets.

Regarding the statistical parity measure, in general, {10/15} datasets have an absolute value of statistical parity less than {0.1}. The \textit{Diabetes, Credit card clients} and \textit{OULAD}  datasets have the best value {(0.0)} of statistical parity while the \textit{Bank marketing} dataset has the worst value. Interestingly, in terms of the equalized odds measure, the best value (0.0) is observed in four datasets (\textit{Credit card clients, Diabetes, OULAD} and \textit{Ricci}). The predictive model results in the worst performance on the \textit{COMPAS recid.} dataset with a high value of equalized odds, followed by the \textit{Law school} and the \textit{Communities \& Crime} datasets.

In addition, we plot the ABROCA slicing of all datasets in Figure~\ref{fig:abroca1}. In the Figure, the \textit{red} ROC curve represents the non-protected group (e.g., Male) while the \textit{blue} ROC is the curve of the protected group (e.g., Female). The best value of the ABROCA is seen in the \textit{Ricci} dataset, followed by the \textit{OULAD} and the \textit{KDD Census-Income} datasets. The worst cases are the \textit{German credit} and the \textit{COMPAS} datasets.

%% file: other-issues.tex
\section{Open issues on datasets for fairness-aware ML}
\label{sec:other_issues}
In the previous sections, we have summarized the most popular datasets for fairness-aware learning. In this section, we extend the discussion to also include recently proposed (and therefore, not adequately exploited) real datasets (Section~\ref{sec:adult_new_dataset}), synthetic datasets (Section~\ref{subsec:synthetic_dataset} and datasets for sequential decision making (Section~\ref{subsec:temporal_dataset}). We advocate that the community should focus more on new datasets representing diverse fairness scenarios, in parallel to new methods and algorithms for fairness-aware learning.

\subsection{Adult reconstruction and ACS PUMS datasets}
\label{sec:adult_new_dataset}
The Adult reconstruction dataset\footnote{https://github.com/zykls/folktables/}\cite{ding2021retiring} is a new dataset reconstructed from the Current Population Survey (CPS) data~\cite{flood2020} from 1994. The dataset consists of 49,531 instances with 14 attributes, in which 13 of 15 attributes of the Adult dataset (see Section~\ref{subsubsec:adult}) are matched. 
Differently from the vanilla Adult dataset, the class attribute  \emph{income} is now represented as a continuous variable. A possible prediction task of the Adult reconstruction dataset is to decide whether an individual earns annually more than  50,000 US dollars. Apart from the Adult reconstruction dataset, \cite{ding2021retiring} introduce further datasets based on the American Community Survey (ACS) Public Use Microdata Sample (PUMS)\footnote{https://www2.census.gov/programs-surveys/acs/data/pums/} with five new prediction tasks w.r.t. \emph{income, public health insurance, residential address, employment} and \emph{commuting time to workplace}.

\begin{figure*}[h!]
  \centering
  \includegraphics[width=0.5\linewidth]{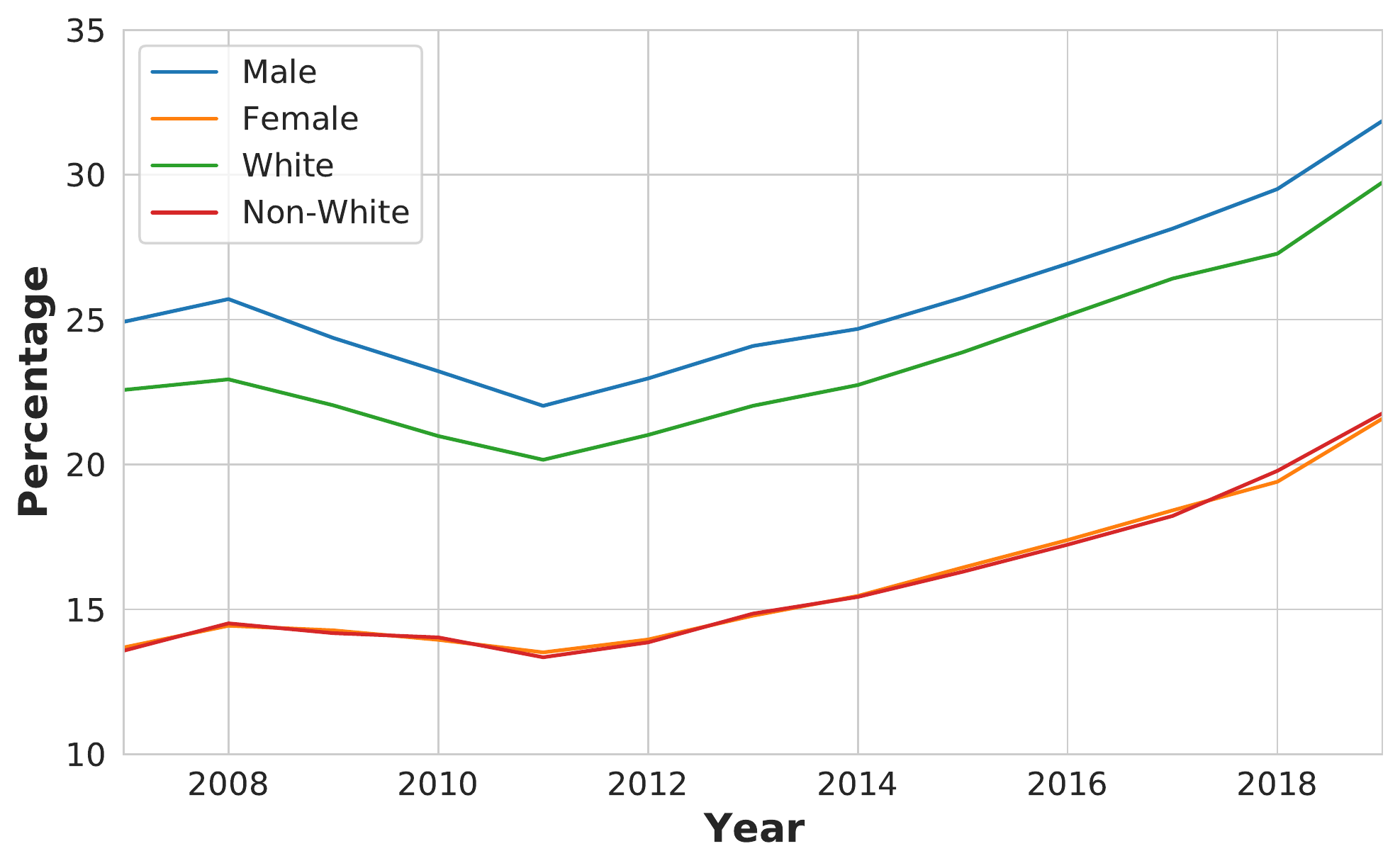}
  \caption{ACS PUMS dataset (California, 2007-2019): Proportion of people with an income above 50K\$ in each group over the years}
  \label{fig:ACS-ratio}
\end{figure*}

For our study, we focus on a particular state (California) for a specific period (2007-2019), using the provided tool \cite{ding2021retiring} and \emph{income} as the prediction task. We consider two protected attributes: \emph{Sex = \{male, female\}} and \emph{race = \{white, non-white\}}, with "female" and "non-white" being the corresponding protected values.

\begin{figure*}[h!]
  \centering
  \includegraphics[width=0.8\linewidth]{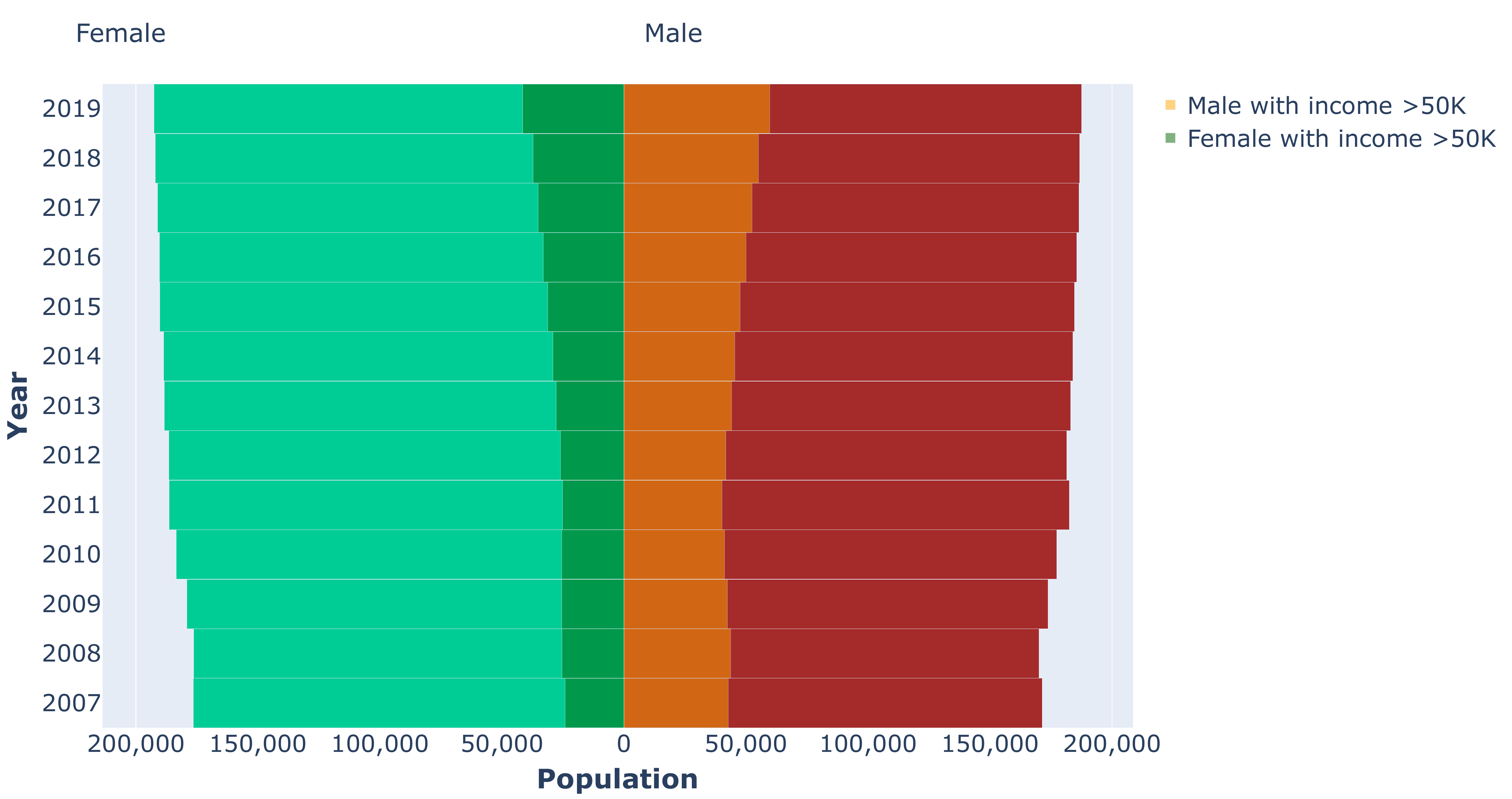}
  \caption{ACS PUMS dataset (California, 2007-2019): The distribution of people w.r.t \emph{sex} and \emph{income} over the years}
  \label{fig:ACS-pyramid}
\end{figure*}

In Figure~\ref{fig:ACS-ratio}, we depict the proportion of people in each income class over time split per gender and race. 
It is easy to observe a lower representation of the protected groups (female, non-white) over the years.
In relation to the population size, which is shown in Figure \ref{fig:ACS-pyramid}, the number of people with an income above 50K\$ gradually increases in both sexes. However, the growth rate in the male group is slightly higher than that in the female group.

\begin{figure}[h!]
    \centering
    \begin{subfigure}[t]{0.47\textwidth}
        \centering
        \includegraphics[width=1\linewidth]{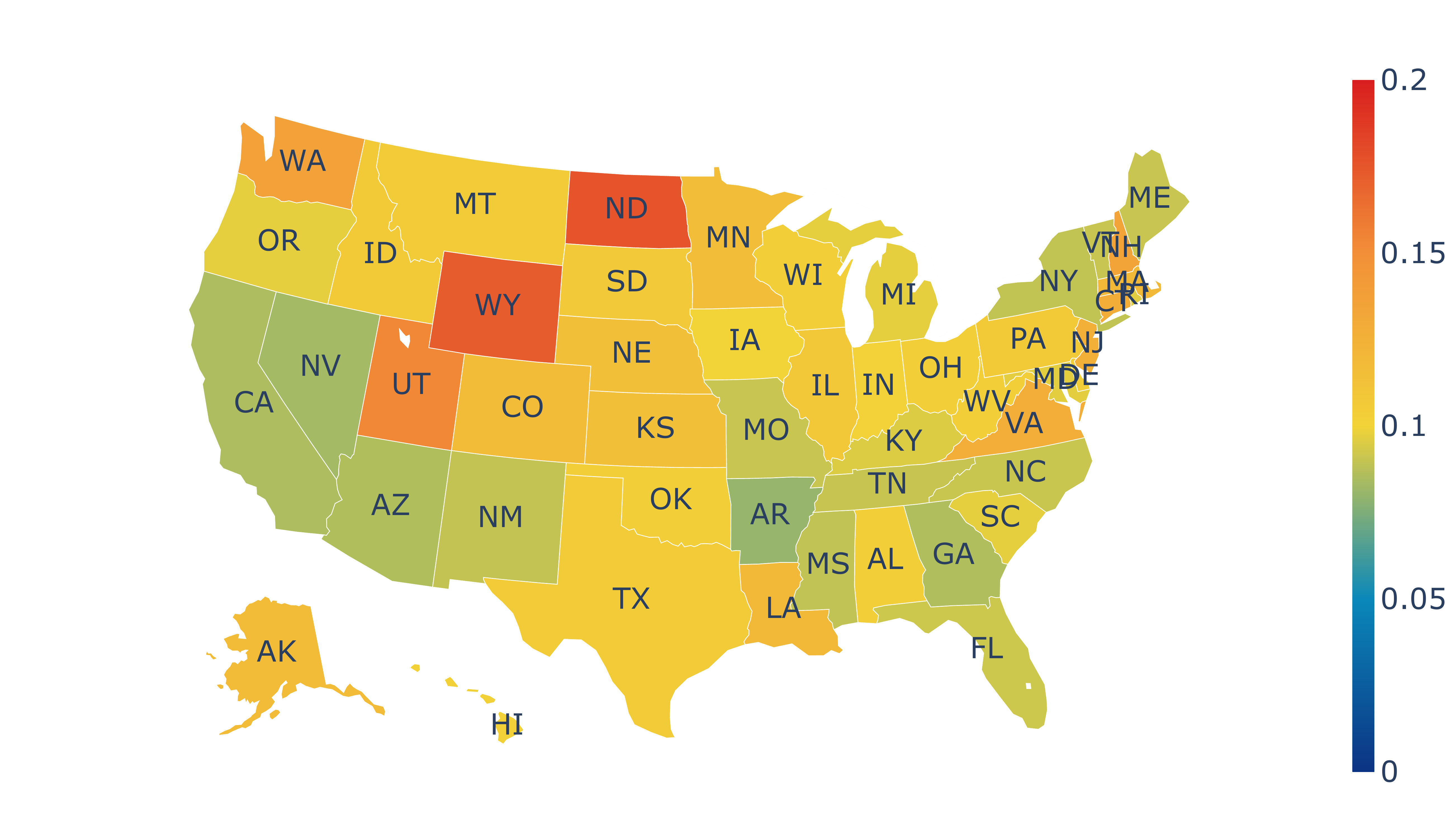}
        \caption{2011}
    \end{subfigure}
    \quad
    \begin{subfigure}[t]{0.47\textwidth}
        \centering
        \includegraphics[width=1\linewidth]{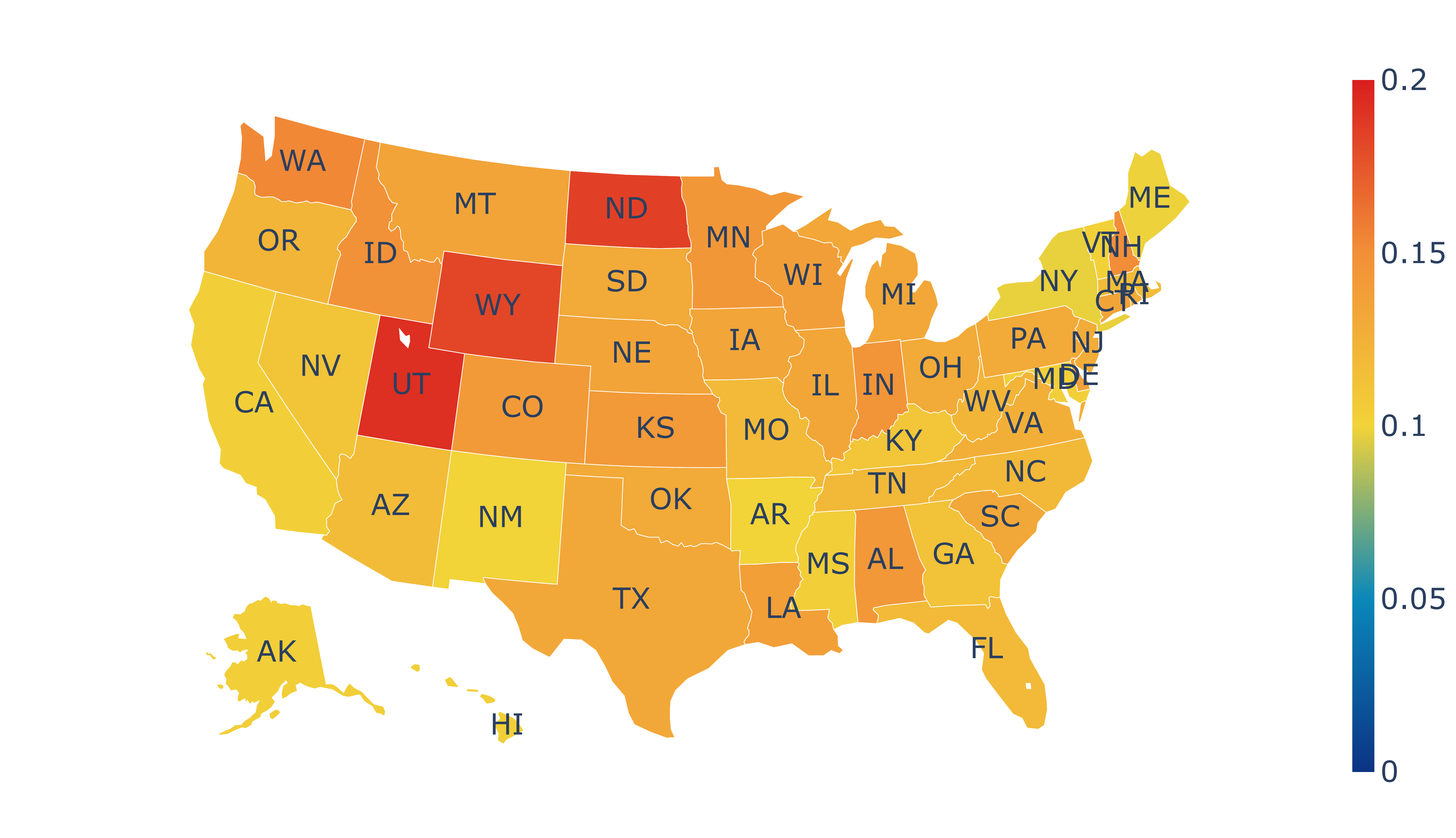}
        \caption{2019}
    \end{subfigure}
    \caption{ACS PUMS dataset: Gender differences (\%males-\%females) in the positive class ($>50K$ income) for different US states}
    \label{fig:ACS-heatmap}
\end{figure}

The ACS PUMS datasets were only recently proposed~\cite{ding2021retiring}. We believe they comprise a very interesting collection since they also contain spatial and temporal information, albeit only for the US, and can therefore be used to analyze the dynamics of discrimination across space and time.
As a preliminary investigation, in Figure \ref{fig:ACS-heatmap}, we illustrate the gender percentage differences in the positive class, i.e., income over 50K\$, for different US states in 2011 and 2019. Many states have low \emph{gender differences} (depicted in green) in 2011. However, the \emph{gender differences} increase over the years, as seen in 2019. A further investigation of the potential effect of spatial and temporal parameters is of course required.

\subsection{Synthetic datasets}
\label{subsec:synthetic_dataset}
Apart from using real-world datasets, it is typical for machine learning evaluation~\cite{ntoutsi2019evaluation} to also employ synthetic data which allow for evaluation under different learning complexity scenarios. Synthetic datasets have been also used for the evaluation of fairness-aware learning methods
~\cite{zafar2017fairness,loh2019subgroup,d2020longtermfairness,tu2020fairlong,reddy2021benchmarking} to produce desired testing scenarios, which may not yet be captured by the existing real-world datasets, but are essential for the development and evaluation of theoretically sound fair algorithms. 

For example, the works~\cite{d2020longtermfairness,tu2020fairlong} study the long-term effects of a currently fair decision-making system and therefore require data that capture the decision of the classifier continuously through time and change the underlying population accordingly. To this end, they simulate dataset changes over time.

In a different direction, \cite{iosifidisdealing} use synthetic data augmentation to increase the representation of the underrepresented protected groups in the overall population. The synthetic instances are generated via SMOTE~\cite{chawla2002smote} by interpolating between original instances.

\subsection{Sequential datasets}
\label{subsec:temporal_dataset}
While algorithmic fairness in decision-making has been mostly studied in static/batch settings, increasing attention has been gained in sequential decision-making environments~\cite{liu2018delayed,heidari2018preventing,wen2021algorithms}, where a sequence of instances possibly infinitely arrives continuously over time which widely exists in many real-world applications such as when making decisions about lending and employment. In contrast to the batch based static environments, sequential decision-making requires the operating model processes each new individual at each time step while making an irrevocable decision based on observations made so far~\cite{zhang2020feat,zhang2019faht}. Often times, the processing also needs to be on the fly and without the need for storage and reprocessing~\cite{zhang2021farf}. 

The aforementioned unique characteristics require the datasets being used for fair sequential decision-making studies fulfill these additional demanding requirements. Among the previously discussed datasets, the \textit{Adult}~\cite{kohavi1996scaling} and \textit{Census}~\cite{asuncion2007uci} are rendered as discriminated data streams to fit for this purpose by processing the individuals in sequence~\cite{zhang2020feat}. In addition, the datasets are ordered based on the sensitive attribute of their particular task at hand before sequential processing to further simulating the potential concept and fairness drifts in the online settings~\cite{zhang2021farf}. Revevantly, the \textit{Crime and Communities} dataset~\cite{asuncion2007uci} is also sequentially processed for sequential fairness-aware studies~\cite{heidari2018preventing}. However, sequential-friendly datasets, due to their magnified requirements, are still in scarce, albeit their significance for the development of fair sequential models which are widely applicable in many real-world applications~\cite{zhang2019fairness}. A continuation on fair sequential datasets efforts is therefore required for a unified fairness-aware research. The new Adult dataset(s) (see Section~\ref{sec:adult_new_dataset}) might be suitable for sequential learning as they contain temporal information (year of data collection).

More recently, the uncertainty due to censorship in fair sequential decision-making has also been researched~\cite{zhang2021fair,zhang2022fair}. Distinct from existing fairness studies assuming certainty on the class label by designed, this line of works addresses fairness in the presence of uncertainty on the class label due to censorship. Take the motivating clinical prediction therein as the example (e.g., SUPPORT dataset~\cite{knaus1995support}), whether the patient relapses/discharges (event of interest) could be unknown for various reasons (e.g., loss to followup) leading to uncertainty on the class label, i.e., censorship~\cite{zhang2016using}. This problem extends beyond the medical domain with examples in marketing analytics (e.g., KKBox dataset~\cite{kvamme2019time}) and recidivism prediction instruments (e.g., ROSSI~\cite{fox2012rcmdrplugin} and COMPAS dataset~\cite{angwin2016machine}). The censorship information, including survival time and an event indicator, in addition to the observed features, is thus also included, which is normally excluded in fairness studies that do not consider censorship. As the exclusion of censorship information could lead to important information loss and introduce substantial bias~\cite{wang2021harmonic}, more attention on the censorship of fairness datasets is warranted.     

Related to the topic of fairness, is the topic of explainability. Explainability tools can help debugging ML models and uncover biased decision making. For sequential decision making, the notion of sequential counterfactuals~\cite{Naumann2021} seems prominent as it takes into account longer-term consequences of feature-value changes. The experiments were conducted on the Adult dataset, however the fairness of the decisions were not investigated. Further research in this direction is required.

%% file: conclusion.tex
\section{Conclusion and outlook}
\label{sec:conclusion}
There are several approaches and discussions that can be implemented in studies on fairness-aware ML. First, in this survey, we investigate the tabular data as the most prevalent data representation. However, in practice, other data types such as text~\cite{zhao2018gender} and images~\cite{buolamwini2018gender} are also used in fairness-aware machine learning problems. Obviously, these data types are closely related to the domain, and the method of handling data sets is also very different and specialized. This requires the fairness-aware algorithms to be tweaked to apply to different datasets.

Second, by generating the Bayesian network, we discover the relationship between
attributes showing their conditional dependence. The results from data analysis and experiments show that the bias may appear in the data itself and/or in the outcome of predictive models. It is understandable that if a dataset contains bias and discrimination, it would be difficult for fairness-aware algorithms to find the trade-off between fairness requirement and performance. Furthermore, based on our experimental results, a significant variation in outcomes between the datasets suggests that the fairness-aware models need to be performed on the diverse datasets.

Third, bias and discrimination are the common problems of almost all domains in reality. In this paper, we study the well-known datasets describing the important aspects of social life such as finance, education, healthcare and criminology. The definition of fairness, of course, is different across domains. It isn't easy to evaluate the efficiency of fairness-aware algorithms because they must be based on such fairness notions. Therefore, it is crucial and necessary to select or define the appropriate fairness notions for each problem in each domain because there is no universal fairness notion for every problem. This remains a major challenge for researchers.

Fourth, the selection of the protected attributes is also a matter of consideration. In the datasets surveyed in this paper, \textit{gender (sex), race, age} and \textit{marriage} are the prevalent protected attributes. The selection of one or more protected attributes for the experiment depends on many factors such as domain, problem and the purpose of the experiment. In our experiments, for each dataset, we only demonstrate the performance of the predictive model w.r.t one of the most popular protected attributes. In addition, the identification and handling of ``proxy'' attributes is also an issue that requires more research.

Fifth, collecting new datasets is always a requirement of data scientists. The surveyed datasets were all collected quite a long time in the past with an average \emph{age} of about 20 years. The oldest dataset was obtained 48 years ago, while the newest dataset was identified from 7 years ago. Of course, the newer the data, the more up-to-date with the trends of the modern society, so the analysis and application of fairness-aware algorithms on the new datasets will reflect the manifestations of the social behaviors more realistic. On the other hand, the old datasets are of reference value in comparing and contrasting the movement and variation of fairness in the same or different domains. The datasets are collected in the US and European countries where the data protection laws are in place. However, the general policies on data quality or collection still need to be studied and proposed~\cite{ntoutsi2020bias}.

To conclude, fairness-aware ML has attracted many recently in various domains from criminology, healthcare, finance to education. This paper reviews the most popular datasets used in fairness-aware ML researches. We explore the relationship of the variables as well as analyze their correlation concerning protected attributes and the class label. We believe our analysis will be the basis for developing frameworks or simulation environments to evaluate fairness-aware algorithms. In another aspect, an excellent understanding of well-known datasets can also inspire researchers to develop synthetic data generators because finding a suitable real-world dataset is never a simple task.

%% file: appendix.tex
\newpage
\appendix
\section{Citations}
\label{sec:citation}
\begin{enumerate}
    \item \textbf{Adult dataset}
    
    \cite{krasanakis2018adaptive,kamiran2012data,kamiran2013quantifying,calders2009building,vzliobaite2011handling,calders2010three,luong2011k,kamiran2010discrimination,iosifidisdealing,calmon2017optimized,feldman2015certifying,hajian2013methodology,zafar2015fairness,zliobaite2015relation,calders2010classification,feldman2015computational,fish2015fair,fish2016confidence,friedler2019comparative,ristanoski2013discrimination,chakraborty2020making,quadrianto2017recycling,xu2020algorithmic,zafar2019fairness,choi2020learning,oneto2019taking,grari2019fair,l2019framework,agarwal2018reductions,backurs2019scalable,hu2020fairnn,chierichetti2017fair,ziko2019variational,haeri2020crucial,berk2017convex,esmaeili2020probabilistic,deepak2020fair,mahabadi2020individual,huang2019coresets,kearns2019empirical,bechavod2017penalizing,Ruoss2020,zhang2018mitigating,iosifidis2019adafair,du2020fairness,zhang2019faht,galhotra2021learning,abbasi2021fair,diana2021minimax,galhotra2021learning,chakraborty2020fairway}.
    
    \item \textbf{KDD Census-Income dataset}
    
    \cite{iosifidis2019adafair,ristanoski2013discrimination,iosifidis2020fabboo,zhang2019faht}.
    
    \item \textbf{German credit dataset} 
    
    \cite{calders2009building,luong2011k,ruggieri2010data,pedreshi2008discrimination,pedreschi2009measuring,iosifidisdealing,feldman2015certifying,hajian2013methodology,kamiran2009classifying,feldman2015computational,fish2016confidence,zemel2013learning,friedler2019comparative,mancuhan2014combating,ristanoski2013discrimination,choi2020learning,Ruoss2020,ahn2019fairsight,chakraborty2020fairway}.
    
    \item \textbf{Dutch census dataset}
    
    \cite{kamiran2010discrimination,kamiran2012data,kamiran2013quantifying,vzliobaite2011handling,kamiran2010discrimination,xu2020algorithmic,l2019framework,agarwal2018reductions}.
    
    \item \textbf{Bank marketing dataset}
    
    \cite{grari2019fair,zafar2019fairness,krasanakis2018adaptive,zafar2015fairness,fish2016confidence,backurs2019scalable,hu2020fairnn,chierichetti2017fair,ziko2019variational,haeri2020crucial,berafair2019,mahabadi2020individual,huang2019coresets,galhotra2021learning,abbasi2021fair,galhotra2021learning}.
    
    \item \textbf{Credit card clients dataset}
    
    \cite{yeh2009comparisons,berk2017convex,esmaeili2020probabilistic,berafair2019,deepak2020fair,bechavod2017penalizing,chakraborty2020fairway}.
    
    \item \textbf{COMPAS dataset}
    
    \cite{krasanakis2018adaptive,calmon2017optimized,chouldechova2017fair,zafar2017fairness,corbett2017algorithmic,friedler2019comparative,tu2020fairlong,quadrianto2017recycling,xu2020algorithmic,zafar2019fairness,slack2020fairness,choi2020learning,grgic2018beyond,oneto2019taking,l2019framework,agarwal2018reductions,haeri2020crucial,lahoti2019operationalizing,heidari2018fairness,russell2017worlds,berk2017convex,Ruoss2020,du2020fairness,diana2021minimax,van2021effect,chakraborty2020fairway}.
    
    \item \textbf{Communites \& Crime dataset}
    
    \cite{kamiran2012data,kamiran2013quantifying,kamiran2010discrimination,lahoti2019operationalizing,kearns2018preventing,narasimhan2020pairwise,slack2020fairness,sharifi2019average,heidari2018fairness,calders2013controlling,chzhen2020fair,kearns2019empirical,berk2017convex,Ruoss2020,galhotra2021learning,diana2021minimax,galhotra2021learning}.
    \item \textbf{Diabetes dataset}
    
    \cite{backurs2019scalable,chierichetti2017fair,mahabadi2020individual,huang2019coresets}.
    \item \textbf{Ricci dataset}
    
    \cite{feldman2015certifying,feldman2015computational,friedler2019comparative,ignatiev2020towards,schelter2019fairprep,valdiviafair}.
   \item \textbf{Student performance dataset}
   
   \cite{deepak2020fair,chzhen2020fair,kearns2019empirical,quy2021faircapacitated}.
   
   \item \textbf{OULAD dataset} 
   
   \cite{riazy2019predictive,quy2021faircapacitated,riazy2020fairness}.
   
   \item \textbf{Law School dataset}
   
   \cite{chzhen2020fair,kearns2019empirical,kusner2017counterfactual,russell2017worlds,lahoti2020fairness,bechavod2017penalizing,berk2017convex,yang2020fairness,Ruoss2020}.
\end{enumerate}

\newpage
\section{Datasets' characteristics}
\label{sec:dataset_characteristics}

\begin{table*}[h!]
\caption{KDD Census-Income: attributes characteristics (continued)}
\label{tbl:kdd_attribute_extra}
\begin{adjustbox}{width=1\linewidth}
    \begin{tabular}{llccl}
         \hline
        \multicolumn{1}{c}{\textbf{Attributes}} & \multicolumn{1}{c}{\textbf{Type}} & 
        \multicolumn{1}{c}{\textbf{Values}} & 
        \multicolumn{1}{c}{\textbf{\#Missing values}} &
        \multicolumn{1}{c}{\textbf{Description}}
        \\ \hline

enroll-in-edu-inst-last-wk  & Categorical &  3  & 0 & An individual enrolled in an educational institute last week?\\
major-industry  & Categorical &  24  & 0 & The major industry code\\
major-occupation  & Categorical &  15  & 0 & The major occupation code\\
hispanic-origin  & Categorical &  9  & 1,279 & The Hispanic origin \\
member-union  & Categorical &  3  & 0 & Member of a labor union\\
reason-unemployment  & Categorical &  6  & 0 & The reason for unemployment \\
region-previous  & Categorical &  6  & 0 & The region of previous residence\\
state-previous  & Categorical &  50  & 1038 & The state of previous residence\\
migration-code-change-in-msa  & Categorical &  10 &  149,642 & Migration code-change in MSA\\
migration-code-change-in-reg  & Categorical &  9 &  149,642 & Migration code-change in region\\
migration-code-move-within-reg  & Categorical &  10  & 149,642 & Migration code-move within region\\
live-hour-1-year-ago  & Categorical &  3 &  0 & Live in this house 1 year ago\\
migration-prev-res-in-sunbelt  & Categorical &  4  & 149,642 & Migration from the previous residence in the sunbelt\\
country-father  & Categorical &  42  & 10,142 & The country of birth of the father\\
country-mother  & Categorical &  42  & 9,191 & The country of birth of the mother\\
country-birth  & Categorical &  42  & 5,157 & The country of birth\\
fill-questionnaire  & Categorical &  3  & 0 & Fill the questionnaire for veteran's admin\\
        \hline
    \end{tabular}
\end{adjustbox}
\end{table*}


\begin{table*}[!ht]
\caption{COMPAS recid: attributes characteristics (continued)}
\label{tbl:COMPAS_attribute_extra}
\begin{adjustbox}{width=1\linewidth}
    \begin{tabular}{llccl}
        \hline
        \multicolumn{1}{c}{\textbf{Attributes}} & \multicolumn{1}{c}{\textbf{Type}} & 
        \multicolumn{1}{c}{\textbf{Values}} & 
        \multicolumn{1}{c}{\textbf{\#Missing values}} &
        \multicolumn{1}{c}{\textbf{Description}}
        \\ \hline
name  & Categorical &  7,158 &  0 & First and last name of the defendant\\
first  & Categorical &  2,800  & 0 & First name\\
last  & Categorical &  3,950  & 0 & Last name\\
compas\_screening\_date  & Categorical &  690  & 0 & The date on which the decile score was given\\
dob  & Categorical &  5,452  & 0 & Date of birth\\
decile\_score & Numerical &   [1 - 10] & 0 & The COMPAS Risk of Recidivism score\\
days\_b\_screening\_arrest & Numerical &   [-414 - 1,057] & 307 & The number of days between COMPAS screening and arrest\\
c\_jail\_in  & Categorical &  6,907 &  307 &The jail entry date for original crime\\
c\_jail\_out  & Categorical &  6,880 &  307 &The jail exit date for original crime\\
c\_case\_number  & Categorical &  7,192  & 22 & The case number for original crime\\
c\_offense\_date  & Categorical &  927  & 1,159 & The offense date of original crime\\
c\_arrest\_date  & Categorical &  580  & 6,077 & The arrest date for original crime\\
c\_days\_from\_compas & Numerical &   [0 - 9,485] & 22 & Between the COMPAS screening and the original crime offense date\\
c\_charge\_desc  & Categorical &  437  & 29 & Description of charge for original crime\\
is\_recid & Binary   & \{0, 1\} & 0 & The binary indicator of recidivation\\
r\_case\_number  & Categorical &  3,471  & 3,743 & The case number of follow-up crime \\
r\_charge\_degree  & Categorical &  10  & 3,743 & Charge degree of follow-up crime\\
r\_days\_from\_arrest & Numerical &   [-1 - 993] & 4,898 & Between the follow-up crime and the arrest date (days)\\
r\_offense\_date  & Categorical &  1,075  & 3,743 & The date of follow-up crime\\
r\_charge\_desc  & Categorical &  340 &  3,801 & Description of charge for follow-up crime \\
r\_jail\_in  & Categorical &  972 &  4,898 & The jail entry date for follow-up crime\\
r\_jail\_out  & Categorical &  938  & 4,898 & The jail exit date for follow-up crime\\
violent\_recid &    & NULL & 7,214 & Values are all NA. This column is ignored\\
is\_violent\_recid & Binary   & \{0, 1\} & 0 & The binary indicator of violent follow-up crime\\
vr\_case\_number  & Categorical &  819  & 6,395 & The case number for violent follow-up crime\\
vr\_charge\_degree  & Categorical &  9  & 6,395 & Charge degree for violent follow-up crime\\
vr\_offense\_date  & Categorical &  570  & 6,395 & The date of offense for violent follow-up crime\\
vr\_charge\_desc  & Categorical &  83 &  6,395 & Description of charge for violent follow-up crime\\
type\_of\_assessment  & Categorical &  1  & 0 & The type of COMPAS score given for decile score\\
decile\_score.1 & Numerical &   [1 - 10] & 0 & Repeat column of decile score\\
screening\_date  & Categorical &  690  & 0 & Repeat column of compas\_screening\_date\\
v\_type\_of\_assessment  & Categorical &  1  & 0 & The type of COMPAS score given for v\_decile\_score\\
v\_decile\_score & Numerical &   [1 - 10] & 0 &  The COMPAS Risk of Violence score from 1 to 10\\
v\_screening\_date  & Categorical &  690  & 0 & The date on which v\_decile\_score was given\\
in\_custody  & Categorical &  1,156 &  236 & The date on which individual was brought into custody\\
out\_custody  & Categorical &  1,169 &  236 & The date on which individual was released from custody\\
priors\_count.1 & Numerical &   0 - 38 & 0 & Repeat column of priors\_count \\
start & Numerical &   [0 - 937] & 0 & No information\\
end & Numerical &   [0 - 1,186] & 0 & No information\\
event & Binary &   \{0, 1\} & 0 & No information\\
        \hline
    \end{tabular}
\end{adjustbox}
\end{table*}


\begin{table*}[bh!]
\caption{Communities and Crime: attributes characteristics (continued)}
\label{tbl:C_C_attribute_extra1}
\begin{adjustbox}{width=1\linewidth}
    \begin{tabular}{llccl}
        \hline
        \multicolumn{1}{c}{\textbf{Attributes}} & \multicolumn{1}{c}{\textbf{Type}} & 
        \multicolumn{1}{c}{\textbf{Values}} & 
        \multicolumn{1}{c}{\textbf{\#Missing values}} &
        \multicolumn{1}{c}{\textbf{Description}}

        \\ \hline
state & Categorical  & 46 & 0 & The US state (by number)\\
county  & Categorical  &  109 & 1174 & The numeric code for county\\
community  & Categorical  &  800 & 1,177 & The numeric code for community\\
communityname  & Categorical  &  1,828 & 0 & The community name\\
fold & Numerical & [1 - 10] & 0 & The fold number for non-random 10 fold cross validation\\
population & Numerical & [0.0 - 1.0] & 0 & The population for community\\
householdsize & Numerical & [0.0 - 1.0] & 0 & The mean people per household\\
racePctWhite & Numerical & [0.0 - 1.0] & 0 & The percentage of population that is Caucasian\\
racePctAsian & Numerical & [0.0 - 1.0] & 0 & The percentage of population that is of Asian heritage\\
racePctHisp & Numerical & [0.0 - 1.0] & 0 & The percentage of population that is of Hispanic heritage\\
agePct12t21 & Numerical & [0.0 - 1.0] & 0 & The percentage of population that is 12-21 in age\\
agePct12t29 & Numerical & [0.0 - 1.0] & 0 & The percentage of population that is 12-29 in age\\
agePct16t24 & Numerical & [0.0 - 1.0] & 0 & The percentage of population that is 16-24 in age\\
agePct65up & Numerical & [0.0 - 1.0] & 0 & The percentage of population that is 65 and over in age\\
numbUrban & Numerical & [0.0 - 1.0] & 0 & The number of people living in areas classified as urban\\
pctUrban & Numerical & [0.0 - 1.0] & 0 & The percentage of people living in areas classified as urban\\
medIncome & Numerical & [0.0 - 1.0] & 0 & The median household income\\
pctWWage & Numerical & [0.0 - 1.0] & 0 & The percentage of households with wage or salary income in 1989 \\
pctWFarmSelf & Numerical & [0.0 - 1.0] & 0 & The percentage of households with farm or self employment income in 1989\\
pctWSocSec & Numerical & [0.0 - 1.0] & 0 & The percentage of households with social security income in 1989\\
pctWRetire & Numerical & [0.0 - 1.0] & 0 & The percentage of households with retirement income in 1989 \\
medFamInc & Numerical & [0.0 - 1.0] & 0 & The median family income\\
perCapInc & Numerical & [0.0 - 1.0] & 0 & Per capita income (national income divided by population size)\\
whitePerCap & Numerical & [0.0 - 1.0] & 0 & Per capita income for Caucasians\\
blackPerCap & Numerical & [0.0 - 1.0] & 0 & Per capita income for African Americans \\
indianPerCap & Numerical & [0.0 - 1.0] & 0 & Per capita income for native Americans \\
AsianPerCap & Numerical & [0.0 - 1.0] & 0 & Per capita income for people with Asian heritage\\
OtherPerCap  & Numerical &   [0.0 - 1.0]  & 1 & Per capita income for people with 'other' heritage\\
HispPerCap & Numerical & [0.0 - 1.0] & 0 & Per capita income for people with Hispanic heritage\\
PctLess9thGrade & Numerical & [0.0 - 1.0] & 0 & The percentage of people 25 and over with less than a 9th grade education\\
PctNotHSGrad & Numerical & [0.0 - 1.0] & 0 & The percentage of people 25 and over that are not high school graduates\\
PctBSorMore & Numerical & [0.0 - 1.0] & 0 & The percentage of people 25 and over with a bachelors degree or higher education\\
PctUnemployed & Numerical & [0.0 - 1.0] & 0 & The percentage of people 16 and over, in the labor force, and unemployed\\
PctEmploy & Numerical & [0.0 - 1.0] & 0 & The percentage of people 16 and over who are employed\\
PctEmplManu & Numerical & [0.0 - 1.0] & 0 & The percentage of people 16 and over who are employed in manufacturing\\
PctEmplProfServ & Numerical & [0.0 - 1.0] & 0 & The percentage of people 16 and over who are employed in professional services\\
PctOccupManu & Numerical & [0.0 - 1.0] & 0 & The percentage of people 16 and over who are employed in manufacturing\\
PctOccupMgmtProf & Numerical & [0.0 - 1.0] & 0 & The percentage of people 16 and over who are employed in management\\
MalePctNevMarr & Numerical & [0.0 - 1.0] & 0 & The percentage of males who have never married\\
PersPerFam & Numerical & [0.0 - 1.0] & 0 & The mean number of people per family \\
PctWorkMomYoungKids & Numerical & [0.0 - 1.0] & 0 & The percentage of moms of kids 6 and under in labor force\\
PctWorkMom & Numerical & [0.0 - 1.0] & 0 & The percentage of moms of kids under 18 in labor force \\
NumImmig & Numerical & [0.0 - 1.0] & 0 & The total number of people known to be foreign born\\
PctImmigRecent & Numerical & [0.0 - 1.0] & 0 & The percentage of immigrants who immigated within the last 3 years \\
PctImmigRec5 & Numerical & [0.0 - 1.0] & 0 & The percentage of immigrants who immigated within the last 5 years\\
PctImmigRec8 & Numerical & [0.0 - 1.0] & 0 & The percentage of immigrants who immigated within the last 8 years\\        
PctImmigRec10 & Numerical & [0.0 - 1.0] & 0 & The percentage of immigrants who immigated within the last 10 years\\
PctRecentImmig & Numerical & [0.0 - 1.0] & 0 & The percentage of the population who have immigrated within the last 3 years\\
PctRecImmig5 & Numerical & [0.0 - 1.0] & 0 & The percentage of the population who have immigrated within the last 5 years\\
PctRecImmig8 & Numerical & [0.0 - 1.0] & 0 & The percentage of the population who have immigrated within the last 8 years\\
PctRecImmig10 & Numerical & [0.0 - 1.0] & 0 & The percentage of the population who have immigrated within the last 10 years\\
PctSpeakEnglOnly & Numerical & [0.0 - 1.0] & 0 & The percentage of the population who speak only English\\
PctNotSpeakEnglWell & Numerical & [0.0 - 1.0] & 0 & The percentage of population who do not speak English well\\
PctLargHouseFam & Numerical & [0.0 - 1.0] & 0 & The percentage of family households that are large (6 or more)\\
PctLargHouseOccup & Numerical & [0.0 - 1.0] & 0 & The percentage of all occupied households that are large (6 or more people) \\        
PersPerOccupHous & Numerical & [0.0 - 1.0] & 0 & The mean persons per household\\
PersPerOwnOccHous & Numerical & [0.0 - 1.0] & 0 & The mean persons per owner occupied household\\
PersPerRentOccHous & Numerical & [0.0 - 1.0] & 0 & The mean persons per rental household\\
PctPersDenseHous & Numerical & [0.0 - 1.0] & 0 & The percentage of persons in dense housing (more than 1 person per room)\\
PctHousLess3BR & Numerical & [0.0 - 1.0] & 0 & The percentage of housing units with less than 3 bedrooms\\
MedNumBR & Numerical & [0.0 - 1.0] & 0 & The median number of bedrooms\\
PctHousOccup & Numerical & [0.0 - 1.0] & 0 & The percentage of housing occupied\\
PctVacMore6Mos & Numerical & [0.0 - 1.0] & 0 & The percentage of vacant housing that has been vacant more than 6 months\\
MedYrHousBuilt & Numerical & [0.0 - 1.0] & 0 & The median year housing units built\\
PctHousNoPhone & Numerical & [0.0 - 1.0] & 0 & The percentage of occupied housing units without phone (in 1990)\\
PctWOFullPlumb & Numerical & [0.0 - 1.0] & 0 & The percentage of housing without complete plumbing facilities\\
OwnOccLowQuart & Numerical & [0.0 - 1.0] & 0 & Owner-occupied housing - lower quartile value \\
OwnOccMedVal & Numerical & [0.0 - 1.0] & 0 & Owner-occupied housing - median value\\
OwnOccHiQuart & Numerical & [0.0 - 1.0] & 0 & Owner-occupied housing - upper quartile value\\
RentLowQ & Numerical & [0.0 - 1.0] & 0 & Rental housing - lower quartile rent\\
RentMedian & Numerical & [0.0 - 1.0] & 0 & Rental housing - median rent\\
        \hline
    \end{tabular}
\end{adjustbox}
\end{table*}
\begin{table*}[htb!]
\caption{Communities and Crime: attributes characteristics (continued)}
\label{tbl:C_C_attribute_extra2}
\begin{adjustbox}{width=1\linewidth}
    \begin{tabular}{llccl}
        \hline
        \multicolumn{1}{c}{\textbf{Attributes}} & \multicolumn{1}{c}{\textbf{Type}} & 
        \multicolumn{1}{c}{\textbf{Values}} & 
        \multicolumn{1}{c}{\textbf{\#Missing values}} &
        \multicolumn{1}{c}{\textbf{Description}}
        \\ \hline
RentHighQ & Numerical & [0.0 - 1.0] & 0 & Rental housing - upper quartile rent \\
MedRent & Numerical & [0.0 - 1.0] & 0 & The median gross rent\\
MedRentPctHousInc & Numerical & [0.0 - 1.0] & 0 & The median gross rent as a percentage of household income \\
MedOwnCostPctInc & Numerical & [0.0 - 1.0] & 0 & The median owners cost (with a mortgage) as a percentage of household income\\
MedOwnCostPctIncNoMtg & Numerical & [0.0 - 1.0] & 0 & The median owners cost (without a mortgage) as a percentage of household income\\
PctForeignBorn & Numerical & [0.0 - 1.0] & 0 & The percentage of people foreign born\\
PctBornSameState & Numerical & [0.0 - 1.0] & 0 & The percentage of people born in the same state as currently living\\
PctSameHouse85 & Numerical & [0.0 - 1.0] & 0 & The percentage of people living in the same house as in 1985 (5 years before)\\
PctSameCity85 & Numerical & [0.0 - 1.0] & 0 & The percentage of people living in the same city as in 1985 (5 years before)\\
PctSameState85 & Numerical & [0.0 - 1.0] & 0 & The percentage of people living in the same state as in 1985 (5 years before) \\        
LemasSwornFT  & Numerical & [0.0 - 1.0]  & 1,675 & The number of sworn full-time police officers \\
LemasSwFTPerPop  & Numerical &[0.0 - 1.0]  & 1,675 & The number of sworn full-time police officers in field operations\\
LemasSwFTFieldOps  & Numerical &[0.0 - 1.0]  & 1,675 & The sworn full-time police officers in field operations per 100,000 population\\
LemasSwFTFieldPerPop  & Numerical &[0.0 - 1.0]  & 1,675 & The number of sworn full time police officers in field operations\\
LemasTotalReq  & Numerical &[0.0 - 1.0]  & 1,675 & The total requests for police \\
LemasTotReqPerPop  & Numerical &[0.0 - 1.0]  & 1,675 & The total requests for police per 100,000 popuation\\
PolicReqPerOffic  & Numerical &[0.0 - 1.0]  & 1,675 & The total requests for police per police officer\\
PolicPerPop  & Numerical &[0.0 - 1.0]  & 1,675 & The number of police officers per 100,000 population\\
RacialMatchCommPol  & Numerical &[0.0 - 1.0]  & 1,675 & A measure of the racial match between the community and the police force \\
PctPolicWhite  & Numerical &[0.0 - 1.0]  & 1,675 & The percentage of police that are Caucasian\\
PctPolicBlack  & Numerical &[0.0 - 1.0]  & 1,675 & The percentage of police that are African American\\
PctPolicHisp  & Numerical &[0.0 - 1.0]  & 1,675 & The percentage of police that are Hispanic\\
PctPolicAsian  & Numerical &[0.0 - 1.0]  & 1,675 & The percentage of police that are Asian\\
PctPolicMinor  & Numerical &[0.0 - 1.0]  & 1,675 & The percentage of police that are minority of any kind\\
OfficAssgnDrugUnits  & Numerical &[0.0 - 1.0]  & 1,675 & The number of officers assigned to special drug units\\
NumKindsDrugsSeiz  & Numerical &[0.0 - 1.0]  & 1,675 & The number of different kinds of drugs seized\\
PolicAveOTWorked  & Numerical &[0.0 - 1.0]  & 1,675 & Police average overtime worked\\
LandArea & Numerical & [0.0 - 1.0] & 0 & Land area in square miles\\
PopDens & Numerical & [0.0 - 1.0] & 0 & The population density in persons per square mile\\
PctUsePubTrans & Numerical & [0.0 - 1.0] & 0 & The percentage of people using public transit for commuting\\
PolicCars  & Numerical &[0.0 - 1.0]  & 1,675 & The number of police cars\\
PolicOperBudg  & Numerical &[0.0 - 1.0]  & 1,675 & Police operating budget\\
LemasPctPolicOnPatr  & Numerical &[0.0 - 1.0]  & 1,675 & The percentage of sworn full-time police officers on patrol\\
LemasGangUnitDeploy  & Numerical &[0.0 - 1.0]  & 1,675 & Gang unit deployed\\
LemasPctOfficDrugUn & Numerical & [0.0 - 1.0] & 0 & The percentage of officers assigned to drug units\\
PolicBudgPerPop  & Numerical &[0.0 - 1.0]  & 1,675 & Police operating budget per population\\

        \hline
    \end{tabular}
\end{adjustbox}
\end{table*}


\begin{table*}[!ht]
\caption{Diabetes: attributes characteristics (continued)}
\label{tbl:diabetes_attributes_extra}
\begin{adjustbox}{width=1\linewidth}
    \begin{tabular}{llccl}
        \hline
        \multicolumn{1}{c}{\textbf{Attributes}} & \multicolumn{1}{c}{\textbf{Type}} & 
        \multicolumn{1}{c}{\textbf{Values}} & 
        \multicolumn{1}{c}{\textbf{\#Missing values}} &
        \multicolumn{1}{c}{\textbf{Description}} 
        \\ \hline
encounter\_ID & Numerical & [12,522 - 443,867,222] & 0 & Encounter's unique identifier \\
patient\_nbr & Numerical & [135 - 189,502,619] & 0 & Patient's unique identifier\\
weight  & Categorical &  10 & 98,569 & Weight (pounds)\\
admission\_type\_id & Categorical & 8 & 0 & The admission type (emergency, urgent, etc. \\
discharge\_disposition\_id & Categorical & 26 & 0 & Discharge disposition (discharged to home, expired, etc.)\\
admission\_source\_id & Categorical & 17 & 0 & The admission source (physician referral, emergency room, etc.)\\
payer\_code  & Categorical &  18 & 40,256 & Payer code (Medicare, self-pay, etc. \\
medical\_specialty  & Categorical &  73 & 49,949 & The specialty of the admitting physician\\
num\_lab\_procedures & Numerical & [1 - 132] & 0 & The number of lab tests performed during the encounter\\
diag\_1  & Categorical &  717 & 21 & The primary diagnosis\\
diag\_2  & Categorical &  749 & 358 & Secondary diagnosis\\
diag\_3  & Categorical &  790 & 1,423 & Additional secondary diagnosis\\
number\_diagnoses & Numerical & [1 - 16] & 0 & The number of diagnoses entered to the system\\
max\_glu\_serum  & Categorical &  4 & 0 & The range of the results or if the test was not taken\\
repaglinide  & Categorical &  4 & 0 & Whether the drug was prescribed or there was a change in the dosage\\
nateglinide  & Categorical &  4 & 0 & Whether the drug was prescribed or there was a change in the dosage\\
glimepiride  & Categorical &  4 & 0&Whether the drug was prescribed or there was a change in the dosage \\
acetohexamide  & Categorical &  2 & 0&Whether the drug was prescribed or there was a change in the dosage \\
glyburide  & Categorical &  4 & 0&Whether the drug was prescribed or there was a change in the dosage \\
tolbutamide  & Categorical &  2 & 0&Whether the drug was prescribed or there was a change in the dosage \\
pioglitazone  & Categorical &  4 & 0&Whether the drug was prescribed or there was a change in the dosage \\
troglitazone  & Categorical &  2 & 0&Whether the drug was prescribed or there was a change in the dosage \\
tolazamide  & Categorical &  3 & 0&Whether the drug was prescribed or there was a change in the dosage \\
examide  & Categorical &  1 & 0& Whether the drug was prescribed or there was a change in the dosage\\
citoglipton  & Categorical &  1 & 0&Whether the drug was prescribed or there was a change in the dosage \\
insulin  & Categorical &  4 & 0& Whether the drug was prescribed or there was a change in the dosage\\
glyburide-metformin  & Categorical &  4 & 0 &Whether the drug was prescribed or there was a change in the dosage\\
glipizide-metformin  & Categorical &  2 & 0 &Whether the drug was prescribed or there was a change in the dosage\\
glimepiride-pioglitazone  & Categorical &  2 & 0& Whether the drug was prescribed or there was a change in the dosage\\
metformin-rosiglitazone  & Categorical &  1 & 0 &Whether the drug was prescribed or there was a change in the dosage\\
metformin-pioglitazone  & Categorical &  1 & 0 &Whether the drug was prescribed or there was a change in the dosage\\
change  & Binary &  \{No, Ch\} & 0 & Was there a change in diabetic medications?\\
        \hline
    \end{tabular}
\end{adjustbox}
\end{table*}